\documentclass{article}

\usepackage{iclr2024_conference,times}

\usepackage[utf8]{inputenc} %
\usepackage[T1]{fontenc}    %
\usepackage[pagebackref=true,breaklinks=true,colorlinks,bookmarks=true]{hyperref}
\usepackage{url}            %
\usepackage{booktabs}       %
\usepackage{amsfonts}       %
\usepackage{nicefrac}       %
\usepackage{microtype}      %
\usepackage{xcolor}         %
\usepackage{colortbl}
\usepackage{graphicx}
\usepackage{amsmath,amsfonts,amssymb,amsthm}
\usepackage{color}
\usepackage{caption}
\usepackage{subcaption}
\usepackage{xspace}
\usepackage{array}
\usepackage{tikzducks}
\usepackage{wrapfig}
\usepackage[titletoc,title]{appendix}
\usepackage{lscape}
\usepackage{svg}
\usepackage{pifont}

\title{\vspace{-5mm}WildFusion: Learning 3D-Aware Latent \\Diffusion Models in View Space}

\author{Katja Schwarz$^{1,\dagger}$, Seung Wook Kim$^{2,3,4}$, Jun Gao$^{2,3,4}$, Sanja Fidler$^{2,3,4}$, Andreas Geiger$^1$, Karsten Kreis$^2$ \\
$^1$University of T{\"u}bingen, $^2$NVIDIA, $^3$Vector Institute, $^4$University of Toronto
\vspace{-3em}
}

\iclrfinalcopy %

\begin{document}

\newcommand{\ba}{\mathbf{a}}\newcommand{\bA}{\mathbf{A}}
\newcommand{\bb}{\mathbf{b}}\newcommand{\bB}{\mathbf{B}}
\newcommand{\bc}{\mathbf{c}}\newcommand{\bC}{\mathbf{C}}
\newcommand{\bd}{\mathbf{d}}\newcommand{\bD}{\mathbf{D}}
\newcommand{\be}{\mathbf{e}}\newcommand{\bE}{\mathbf{E}}
\newcommand{\bff}{\mathbf{f}}\newcommand{\bF}{\mathbf{F}} %
\newcommand{\bg}{\mathbf{g}}\newcommand{\bG}{\mathbf{G}}
\newcommand{\bh}{\mathbf{h}}\newcommand{\bH}{\mathbf{H}}
\newcommand{\bi}{\mathbf{i}}\newcommand{\bI}{\mathbf{I}}
\newcommand{\bj}{\mathbf{j}}\newcommand{\bJ}{\mathbf{J}}
\newcommand{\bk}{\mathbf{k}}\newcommand{\bK}{\mathbf{K}}
\newcommand{\bl}{\mathbf{l}}\newcommand{\bL}{\mathbf{L}}
\newcommand{\bm}{\mathbf{m}}\newcommand{\bM}{\mathbf{M}}
\newcommand{\bn}{\mathbf{n}}\newcommand{\bN}{\mathbf{N}}
\newcommand{\bo}{\mathbf{o}}\newcommand{\bO}{\mathbf{O}}
\newcommand{\bp}{\mathbf{p}}\newcommand{\bP}{\mathbf{P}}
\newcommand{\bq}{\mathbf{q}}\newcommand{\bQ}{\mathbf{Q}}
\newcommand{\br}{\mathbf{r}}\newcommand{\bR}{\mathbf{R}}
\newcommand{\bs}{\mathbf{s}}\newcommand{\bS}{\mathbf{S}}
\newcommand{\bt}{\mathbf{t}}\newcommand{\bT}{\mathbf{T}}
\newcommand{\bu}{\mathbf{u}}\newcommand{\bU}{\mathbf{U}}
\newcommand{\bv}{\mathbf{v}}\newcommand{\bV}{\mathbf{V}}
\newcommand{\bw}{\mathbf{w}}\newcommand{\bW}{\mathbf{W}}
\newcommand{\bx}{\mathbf{x}}\newcommand{\bX}{\mathbf{X}}
\newcommand{\by}{\mathbf{y}}\newcommand{\bY}{\mathbf{Y}}
\newcommand{\bz}{\mathbf{z}}\newcommand{\bZ}{\mathbf{Z}}

\newcommand{\balpha}{\boldsymbol{\alpha}}\newcommand{\bAlpha}{\boldsymbol{\Alpha}}
\newcommand{\bbeta}{\boldsymbol{\beta}}\newcommand{\bBeta}{\boldsymbol{\Beta}}
\newcommand{\bgamma}{\boldsymbol{\gamma}}\newcommand{\bGamma}{\boldsymbol{\Gamma}}
\newcommand{\bdelta}{\boldsymbol{\delta}}\newcommand{\bDelta}{\boldsymbol{\Delta}}
\newcommand{\bepsilon}{\boldsymbol{\epsilon}}\newcommand{\bEpsilon}{\boldsymbol{\Epsilon}}
\newcommand{\bzeta}{\boldsymbol{\zeta}}\newcommand{\bZeta}{\boldsymbol{\Zeta}}
\newcommand{\beeta}{\boldsymbol{\eta}}\newcommand{\bEta}{\boldsymbol{\Eta}} %
\newcommand{\btheta}{\boldsymbol{\theta}}\newcommand{\bTheta}{\boldsymbol{\Theta}}
\newcommand{\biota}{\boldsymbol{\iota}}\newcommand{\bIota}{\boldsymbol{\Iota}}
\newcommand{\bkappa}{\boldsymbol{\kappa}}\newcommand{\bKappa}{\boldsymbol{\Kappa}}
\newcommand{\blambda}{\boldsymbol{\lambda}}\newcommand{\bLambda}{\boldsymbol{\Lambda}}
\newcommand{\bmu}{\boldsymbol{\mu}}\newcommand{\bMu}{\boldsymbol{\Mu}}
\newcommand{\bnu}{\boldsymbol{\nu}}\newcommand{\bNu}{\boldsymbol{\Nu}}
\newcommand{\bxi}{\boldsymbol{\xi}}\newcommand{\bXi}{\boldsymbol{\Xi}}
\newcommand{\bomikron}{\boldsymbol{\omikron}}\newcommand{\bOmikron}{\boldsymbol{\Omikron}}
\newcommand{\bpi}{\boldsymbol{\pi}}\newcommand{\bPi}{\boldsymbol{\Pi}}
\newcommand{\brho}{\boldsymbol{\rho}}\newcommand{\bRho}{\boldsymbol{\Rho}}
\newcommand{\bsigma}{\boldsymbol{\sigma}}\newcommand{\bSigma}{\boldsymbol{\Sigma}}
\newcommand{\btau}{\boldsymbol{\tau}}\newcommand{\bTau}{\boldsymbol{\Tau}}
\newcommand{\bypsilon}{\boldsymbol{\ypsilon}}\newcommand{\bYpsilon}{\boldsymbol{\Ypsilon}}
\newcommand{\bphi}{\boldsymbol{\phi}}\newcommand{\bPhi}{\boldsymbol{\Phi}}
\newcommand{\bchi}{\boldsymbol{\chi}}\newcommand{\bChi}{\boldsymbol{\Chi}}
\newcommand{\bpsi}{\boldsymbol{\psi}}\newcommand{\bPsi}{\boldsymbol{\Psi}}
\newcommand{\bomega}{\boldsymbol{\omega}}\newcommand{\bOmega}{\boldsymbol{\Omega}}

\newcommand{\nA}{\mathbb{A}}
\newcommand{\nB}{\mathbb{B}}
\newcommand{\nC}{\mathbb{C}}
\newcommand{\nD}{\mathbb{D}}
\newcommand{\nE}{\mathbb{E}}
\newcommand{\nF}{\mathbb{F}}
\newcommand{\nG}{\mathbb{G}}
\newcommand{\nH}{\mathbb{H}}
\newcommand{\nI}{\mathbb{I}}
\newcommand{\nJ}{\mathbb{J}}
\newcommand{\nK}{\mathbb{K}}
\newcommand{\nL}{\mathbb{L}}
\newcommand{\nM}{\mathbb{M}}
\newcommand{\nN}{\mathbb{N}}
\newcommand{\nO}{\mathbb{O}}
\newcommand{\nP}{\mathbb{P}}
\newcommand{\nQ}{\mathbb{Q}}
\newcommand{\nR}{\mathbb{R}}
\newcommand{\nS}{\mathbb{S}}
\newcommand{\nT}{\mathbb{T}}
\newcommand{\nU}{\mathbb{U}}
\newcommand{\nV}{\mathbb{V}}
\newcommand{\nW}{\mathbb{W}}
\newcommand{\nX}{\mathbb{X}}
\newcommand{\nY}{\mathbb{Y}}
\newcommand{\nZ}{\mathbb{Z}}

\newcommand{\cA}{\mathcal{A}}
\newcommand{\cB}{\mathcal{B}}
\newcommand{\cC}{\mathcal{C}}
\newcommand{\cD}{\mathcal{D}}
\newcommand{\cE}{\mathcal{E}}
\newcommand{\cF}{\mathcal{F}}
\newcommand{\cG}{\mathcal{G}}
\newcommand{\cH}{\mathcal{H}}
\newcommand{\cI}{\mathcal{I}}
\newcommand{\cJ}{\mathcal{J}}
\newcommand{\cK}{\mathcal{K}}
\newcommand{\cL}{\mathcal{L}}
\newcommand{\cM}{\mathcal{M}}
\newcommand{\cN}{\mathcal{N}}
\newcommand{\cO}{\mathcal{O}}
\newcommand{\cP}{\mathcal{P}}
\newcommand{\cQ}{\mathcal{Q}}
\newcommand{\cR}{\mathcal{R}}
\newcommand{\cS}{\mathcal{S}}
\newcommand{\cT}{\mathcal{T}}
\newcommand{\cU}{\mathcal{U}}
\newcommand{\cV}{\mathcal{V}}
\newcommand{\cW}{\mathcal{W}}
\newcommand{\cX}{\mathcal{X}}
\newcommand{\cY}{\mathcal{Y}}
\newcommand{\cZ}{\mathcal{Z}}

\newcommand{\figref}[1]{Fig.~\ref{#1}}
\newcommand{\secref}[1]{Sec.~\ref{#1}}
\newcommand{\algref}[1]{Algorithm~\ref{#1}}
\newcommand{\eqnref}[1]{Eq.~\eqref{#1}}
\newcommand{\tabref}[1]{Table~\ref{#1}}

\def\mc{\mathcal}
\def\mb{\mathbf}

\newcommand{\T}{^{\raisemath{-1pt}{\mathsf{T}}}}

\makeatletter
\DeclareRobustCommand\onedot{\futurelet\@let@token\@onedot}
\def\@onedot{\ifx\@let@token.\else.\null\fi\xspace}
\def\eg{e.g\onedot} \def\Eg{E.g\onedot}
\def\ie{i.e\onedot} \def\Ie{I.e\onedot}
\def\cf{cf\onedot} \def\Cf{Cf\onedot}
\def\etc{etc\onedot}
\def\vs{vs\onedot}
\def\wrt{wrt\onedot}
\def\dof{d.o.f\onedot}
\def\etal{et~al\onedot}
\def\iid{i.i.d\onedot}
\makeatother

\renewcommand\UrlFont{\color{blue}\rmfamily}

\newcommand*\rot{\rotatebox{90}}

\newcommand\blfootnote[1]{%
  \begingroup
  \renewcommand\thefootnote{}\footnote{#1}%
  \addtocounter{footnote}{-1}%
  \endgroup
}

\newcommand{\boldparagraph}[1]{\vspace{0.2cm}\noindent{\bf #1} }

\definecolor{darkgreen}{rgb}{0,0.7,0}
\definecolor{darkblue}{RGB}{31,119,180}
\definecolor{darkred}{RGB}{214,39,40}

\newcommand{\red}[1]{\noindent{\color{red}{#1}}}
\newcommand{\ks}[1]{\noindent{\color{cyan}{\textbf{Katja}: #1}}}
\newcommand{\jg}[1]{\noindent{\color{magenta}{\textbf{@JUN} #1}}}
\newcommand{\sk}[1]{\noindent{\color{orange}{\textbf{@Seung} #1}}}
\newcommand{\skc}[1]{\noindent{\color{orange}{\textbf{Seung} #1}}}
\newcommand{\kk}[1]{\noindent{\color{darkgreen}{\textbf{@Karsten} #1}}}

\newcommand{\kkc}[1]{\noindent{\color{darkgreen}{\textbf{Karsten:} #1}}}

\newcommand{\ourmodel}{WildFusion\xspace}

\newcommand{\reconstruction}{
\begin{table}[t]
\vspace{-2mm}
  \caption{\small Reconstruction and novel-view synthesis on SDIP Dogs, Elephants and Horses at resolution $256^2$. All evaluations on held-out test set. We report LPIPS, novel-view FID (nvFID) and non-flatness-score (NFS).}
  \label{tab:reconstruction}
  \vspace{-0.3cm}
  \resizebox{\linewidth }{!}{
\begin{tabular}{lccc|ccc|cccc}
\toprule
                         & \multicolumn{3}{c}{SDIP Dogs}                           & \multicolumn{3}{c}{SDIP Elephants}                      & \multicolumn{3}{c}{SDIP Horses}                         & Rec.Time \\ \cmidrule{2-4} \cmidrule{5-7} \cmidrule{8-10}
                         & $\downarrow$LPIPS & $\downarrow$ nvFID & $\uparrow$ NFS & $\downarrow$LPIPS & $\downarrow$ nvFID & $\uparrow$ NFS & $\downarrow$LPIPS & $\downarrow$ nvFID & $\uparrow$ NFS & $\downarrow$t[s]   \\ \midrule
EG3D*~\citep{Chan2022CVPR} &          0.44         &           71.23         &        12.01        &         0.43          &            27.99         &        12.89        &       0.40            &       68.25             &       12.90         &            $>$ 100         \\
EG3D* + $D^{depth}$       &          0.38          &            36.65        &         14.43       &        0.45           &      56.86              &       15.74         &         0.34          &        36.27            &        12.98        &             $>$ 100        \\ \midrule
\ourmodel (\textit{ours}) &           \textbf{0.21}        &         \textbf{17.4}           &     \textbf{31.8}           & \textbf{0.28}              & \textbf{9.0}                & \textbf{32.0}           & \textbf{0.22}              &      \textbf{13.4}              &          \textbf{28.7}      &   \textbf{0.04}                  \\ \bottomrule
\end{tabular}
}
\end{table}
}

\newcommand{\ablations}{
\begin{wraptable}{r}{0.46\textwidth}
\vspace{-1mm}
  \caption{\small Ablation study on SDIP Dogs.} 
  \vspace{-0.3cm}
  \label{tab:ablations}
  \resizebox{1.0\linewidth }{!}{
\begin{tabular}{lcc}
\toprule
      Model configuration                    & $\downarrow \text{nvFID}$   & $\uparrow$NFS             \\ \midrule
Base config     &            53.2    &                 10.3          \\
+ $D_{\phi}$                &            61.1               &             37.0               \\
+ ViT backbone                     &        48.3                   &         33.9                  \\
+ $D^{depth}_{\chi}$                &       40.5                    &           34.0                 \\
+ modeling unbounded scenes & 32.8 & 32.4 \\
+ $\mathcal{L}_{depth}^{2D}$           &             34.6              &            32.0                \\
+ encode depth     &             33.3              &            33.3                \\
+ $\mathcal{L}_{depth}^{3D}$           &             34.0              &             33.7              \\ \bottomrule
\end{tabular}
}
\vspace{-0.5cm}
\end{wraptable}
}

\newcommand{\baselinecomp}{
\begin{table}[t!]
  \vspace{-2mm}
  \caption{\small 3D-aware image synthesis results on unimodal datasets. Baselines above double line require camera pose estimation; methods below work in view space.}
  \vspace{-0.3cm}
  \label{tab:baselinecomp}
  \setlength{\tabcolsep}{3pt}
  \resizebox{\linewidth }{!}{
\begin{tabular}{lccccc|ccccc|ccccc}
\toprule
                                & \multicolumn{5}{c}{SDIP Dogs}                                                                                     & \multicolumn{5}{c}{SDIP Elephants}                                                                                & \multicolumn{5}{c}{SDIP Horses}                                                                                   \\\cmidrule{2-16}
                                & $\downarrow$$\text{FID}$ & $\downarrow$$\text{FID}_\text{CLIP}$ & $\uparrow$NFS & $\uparrow$Precision & $\uparrow$Recall & $\downarrow$$\text{FID}$ & $\downarrow$$\text{FID}_\text{CLIP}$ & $\uparrow$NFS & $\uparrow$Precision & $\uparrow$Recall & $\downarrow$$\text{FID}$ & $\downarrow$$\text{FID}_\text{CLIP}$ & $\uparrow$NFS & $\uparrow$Precision & $\uparrow$Recall \\ \midrule
POF3D~\citep{Shi2023CVPR}        &                17.4                     &                 5.4                         &        28.9       &  0.57  &     0.36         &      6.4                               &                            9.2              &        30.2       &  0.59     &        0.30      &                    16.4                 &   15.1                                       &       \textbf{32.6}        &     0.56  & 0.25           \\
3DGP~\citep{skorokhodov20233dgp} & \textbf{5.9}                                 &          6.2                                & \textbf{36.3}          &  \textbf{0.73}  & \textbf{0.38}             & 3.7                                 &                  \textbf{5.9}                        & 32.1          &  0.67 & 0.22             & 9.0                                 &                                13.0          & 29.2          &  0.60  &0.28             \\ \midrule\midrule
EG3D*~\citep{Chan2022CVPR}        & 16.3                                & 5.8                                      & 11.8          & 0.60   & 0.29             & 3.0                                 & 6.8                                      & 13.3          &  0.59  & 0.31             & 6.7                                 & 10.2                                     & 14.1          &  0.57  & 0.23             \\
EG3D* + $D_{\text{depth}}$       & 18.7                                & 8.8                                      & 13.9          &  0.71  & 0.15             & 4.5                                 & 8.5                                      & 18.3          &  0.56  & 0.24             & 6.5                                 &               8.7                           & 13.7          & 0.59  &  0.30             \\
StyleNeRF~\citep{Gu2021ARXIV}       & 12.3                                & 7.9                                      & 30.0          &  0.65  & 0.34             & 10.0                                 & 9.1                                      & 20.1         &  0.53  & 0.17             & 4.5                                 &               \textbf{8.1}                           & 27.2          & 0.65  &  0.35             \\ \midrule
\ourmodel (\textit{ours})       & 12.2                                &              \textbf{5.2}                            &         31.7      & 0.66  & \textbf{0.38}                & \textbf{2.9}                                 & 6.5                                      & \textbf{32.2}          &   \textbf{0.70}  &\textbf{0.34}             & \textbf{4.3}                                 & 8.8                                      & 28.8          &  \textbf{0.70}  &\textbf{0.37}             \\ \bottomrule
\end{tabular}
}
\vspace{-0.4cm}
\end{table}
}

\newcommand{\baselinecompnocite}{
\begin{table}[t]
  \caption{\small 3D-aware image synthesis results on unimodal datasets. Baselines above double line require camera pose estimation; methods below work in view space. IVID results taken from their work; other baselines run ourselves. Newly added results are marked in \red{red}.}
  \label{tab:baselinecomp}
  \setlength{\tabcolsep}{3pt}
  \resizebox{\linewidth }{!}{
\begin{tabular}{lccccc|ccccc|ccccc}
\toprule
                                & \multicolumn{5}{c}{SDIP Dogs}                                                                                     & \multicolumn{5}{c}{SDIP Elephants}                                                                                & \multicolumn{5}{c}{SDIP Horses}                                                                                   \\\cmidrule{2-16}
                                & $\downarrow$$\text{FID}$ & $\downarrow$$\text{FID}_\text{CLIP}$ & $\uparrow$NFS & $\uparrow$\red{Precision} & $\uparrow$Recall & $\downarrow$$\text{FID}$ & $\downarrow$$\text{FID}_\text{CLIP}$ & $\uparrow$NFS & $\uparrow$\red{Precision} & $\uparrow$Recall & $\downarrow$$\text{FID}$ & $\downarrow$$\text{FID}_\text{CLIP}$ & $\uparrow$NFS & $\uparrow$\red{Precision} & $\uparrow$Recall \\ \midrule
POF3D        &                17.4                     &                 5.4                         &        28.9       &  \red{0.57}  &     0.36         &      6.4                               &                            9.2              &        30.2       &  \red{0.59}     &        0.30      &                    16.4                 &   15.1                                       &       \textbf{32.6}        &     \red{0.56}  & 0.25           \\
3DGP & \textbf{5.9}                                 &          6.2                                & \textbf{36.3}          &  \red{\textbf{0.73}}  & \textbf{0.38}             & 3.7                                 &                  \textbf{5.9}                        & 32.1          &  \red{0.67} & 0.22             & 9.0                                 &                                13.0          & 29.2          &  \red{0.60}  &0.28             \\ \midrule\midrule
EG3D*        & 16.3                                & 5.8                                      & 11.8          & \red{0.60}   & 0.29             & 3.0                                 & 6.8                                      & 13.3          &  \red{0.59}  & 0.31             & 6.7                                 & 10.2                                     & 14.1          &  \red{0.57}  & 0.23             \\
EG3D* + $D_{\text{depth}}$       & 18.7                                & 8.8                                      & 13.9          &  \red{0.71}  & 0.15             & 4.5                                 & 8.5                                      & 18.3          &  \red{0.56}  & 0.24             & 6.5                                 &               8.7                           & 13.7          & \red{0.59}  &  0.30             \\
\red{StyleNeRF} $       & \red{12.3}                                & \red{7.9}                                      & \red{30.0}          &  \red{0.65}  & \red{0.34}             & \red{10.0}                                 & \red{9.1}                                      & \red{20.1}         &  \red{0.53}  & \red{0.17}             & \red{4.5}                                 &               \textbf{\red{8.1}}                           & \red{27.2}          & \red{0.65}  &  \red{0.35}             \\
IVID       & 14.7                                & --                                       & --            & \red{--}       & --           & 11.0                                & --                                       & --         & \red{--}      & --               & 10.2                                & --                                       & --           & \red{--}    & --               \\ \midrule
\ourmodel (\textit{ours})       & 12.2                                &              \textbf{5.2}                            &         31.7      & \red{0.66}  & \textbf{0.38}                & \textbf{2.9}                                 & 6.5                                      & \textbf{32.2}          &   \red{\textbf{0.70}}  &\textbf{0.34}             & \textbf{4.3}                                 & 8.8                                      & 28.8          &  \red{\textbf{0.70}}  &\textbf{0.37}             \\ \bottomrule
\end{tabular}
}
\vspace{-0.4cm}
\end{table}
}

\newcommand{\baselinecompimagenet}{\begin{table}[t!]
\vspace{-0.1cm}
    \centering
    \begin{minipage}[t]{.75\textwidth}
        \scalebox{0.8}{\begin{tabular}{llcccccc}
\toprule
                           & $\downarrow$$\text{FID}$ & $\downarrow$$\text{FID}_\text{CLIP}$ & $\uparrow$NFS & $\uparrow$Precision & $\uparrow$Recall \\  \midrule
3DGP~\citep{skorokhodov20233dgp}       &     \textbf{19.7}              &             \textbf{8.1}             &      18.5                                &        0.31       &                   0.02                 \\
EG3D* + $D_{\text{depth}}$             &         111.6                 &          27.2                            &        20.6       &         0.48            &      0.01            \\
StyleNeRF~\citep{Gu2021ARXIV}                   &     77.7                       &                   17.5                 &        28.0       &           0.38          &        0.03          \\ \midrule
\ourmodel (\textit{ours}), $s=1$       &    65.1           &      15.3            &     33.6    &  0.58        &  0.16 \\
\ourmodel (\textit{ours}), $s=2$      &    35.4           &      11.7            &     \textbf{33.8}    &   \textbf{0.59}        & \textbf{0.20}  \\
\ourmodel (\textit{ours}), $s=5$      &    25.5           &      11.6            &     33.6    &   0.53        & 0.13  \\ \bottomrule
\end{tabular}}
    \end{minipage}\hfill
    \begin{minipage}[]{.25\textwidth}\vspace{0.3cm}
        \caption{\small 3D-aware image synthesis results on class-conditional ImageNet. For \ourmodel, we report results for different classifier-free guidance scales $s$ (App.~\ref{supp:some_details}); extended results in~\tabref{tab:cfgmetrics}.}
        \label{tab:baselinecomp_imagenet}
    \end{minipage}
    \vspace{-0.5cm}
\end{table}}

\newcommand{\tblldmarchitecture}{
\begin{table*}
\setlength{\tabcolsep}{4pt}
\caption{\small Hyperparameters for our latent diffusion models.}
\label{tab:hyperparameters_ldm}
\centering

\begin{tabular}{ll|ll|ll}
\toprule
\multicolumn{2}{c}{\textit{Architecture}}              & \multicolumn{2}{c}{\textit{Training}}   & \multicolumn{2}{c}{\textit{Diffusion Setup}} \\ \midrule
Image shape           & $256\times256\times3$ & Parameterization   & $v$       & Diffusion steps        & 1000       \\
$z$-shape             & $32\times32\times4$   & Learning rate      & $10^{-4}$ & Noise schedule         & Cosine     \\
Channels              & 224                   & Batch size per GPU & 64        & Offset $s$             & 0.008      \\
Depth                 & 2                     & $\#$GPUs             & 4         & Scale factor  $z$      & 0.5        \\
Channel multiplier    & 1,2,4,4               & $p_\mathrm{drop}$    &     0.1      & Sampler                & DDIM       \\
Attention resolutions & 32,16,8              &                    &           & Steps                  & 200        \\
Head channels         & 32                    &                    &           & $\eta$                 & 1.0        \\ \bottomrule
\end{tabular}
\vspace{-0.4cm}
\end{table*}
}

\newcommand{\losses}{
\begin{table}[t]
  \caption{\small{Weights of all losses used in the training objective.}}
  \label{tab:losses}
\centering
\begin{tabular}{llllllll}
    \toprule
       & $\lambda_{px}$ & $\lambda_{VGG}$ & $\lambda_{depth}^{2D}$ & $\lambda_{depth}^{3D}$ & $\lambda_{KL}$ & $\lambda$ & $\lambda_d$ \\ \midrule
Weight & 10                 & 10                  & 1                          & 1                          & 1e-4               & 1    & 10    \\ \bottomrule
\end{tabular}
\end{table}
}

\newcommand{\rebuttalablations}{
\begin{SCtable}
  \resizebox{0.5\linewidth }{!}{
\begin{tabular}{lcc}
\toprule
      Model configuration                    & $\downarrow \text{nvFID}$   & $\uparrow$NFS             \\ \midrule
Base config + ViT backbone + $D_{\phi}$ + $D^{depth}_{\chi}$  &             40.5                    &           34.0           \\
\red{Base config + ViT backbone + $D_{fused}$}                      &        \red{53.1}                   &         \red{33.8}                  \\
Full        &             34.0              &             33.7              \\ 
\red{Full - $D^{depth}_{\chi}$}      &             \red{32.4}              &            \red{30.8}                \\\bottomrule
\end{tabular}
}
\caption{\small Ablation study on SDIP Dogs. Newly added ablations are marked in \red{red}, the other values are taken from Table 3 of the main paper.} 
  \label{tab:ablations}
\end{SCtable}
}

\newcommand{\cfgmetrics}{
\begin{table}[h!]
\renewcommand*{\arraystretch}{1.0}
\setlength{\tabcolsep}{18pt}
  \resizebox{\linewidth }{!}{
\begin{tabular}{lccccc}
\toprule
        & $\downarrow$$\text{FID}$ & $\downarrow$$\text{FID}_\text{CLIP}$ & $\uparrow$NFS & $\uparrow$Precision & $\uparrow$Recall \\ \midrule
$s=1$   & 65.1                     & 15.3                                 & 33.6          & 0.58                & 0.16             \\
$s=1.5$ & 45.2                     & 12.9                                 & 33.7          & \textbf{0.60}                & 0.19             \\
$s=2$   & 35.4                     & 11.7                                 & \textbf{33.8}          & 0.59                & \textbf{0.20}             \\
$s=2.5$ & 30.2                     & 11.0                                 & 32.7          & 0.59                & 0.19             \\
$s=3$   & 28.5                     & \textbf{10.9}                                 & 32.9          & 0.58                   & 0.18                \\ 
$s=5$   & \textbf{25.5}                     & 11.6                                 & 33.6          & 0.53                & 0.13             \\
$s=10$  & 29.7                     & 13.7                                 & 33.1          & 0.44                & 0.08         \\ \bottomrule 
\end{tabular}
}
\caption{\small Evaluation on ImageNet with different classifier-free guidance scales $s$.} 
  \label{tab:cfgmetrics}
\end{table}
}

\newcommand{\teaser}{
\begin{figure*}
\vspace{-7mm}
\centering
  \includegraphics[width=0.99\textwidth]{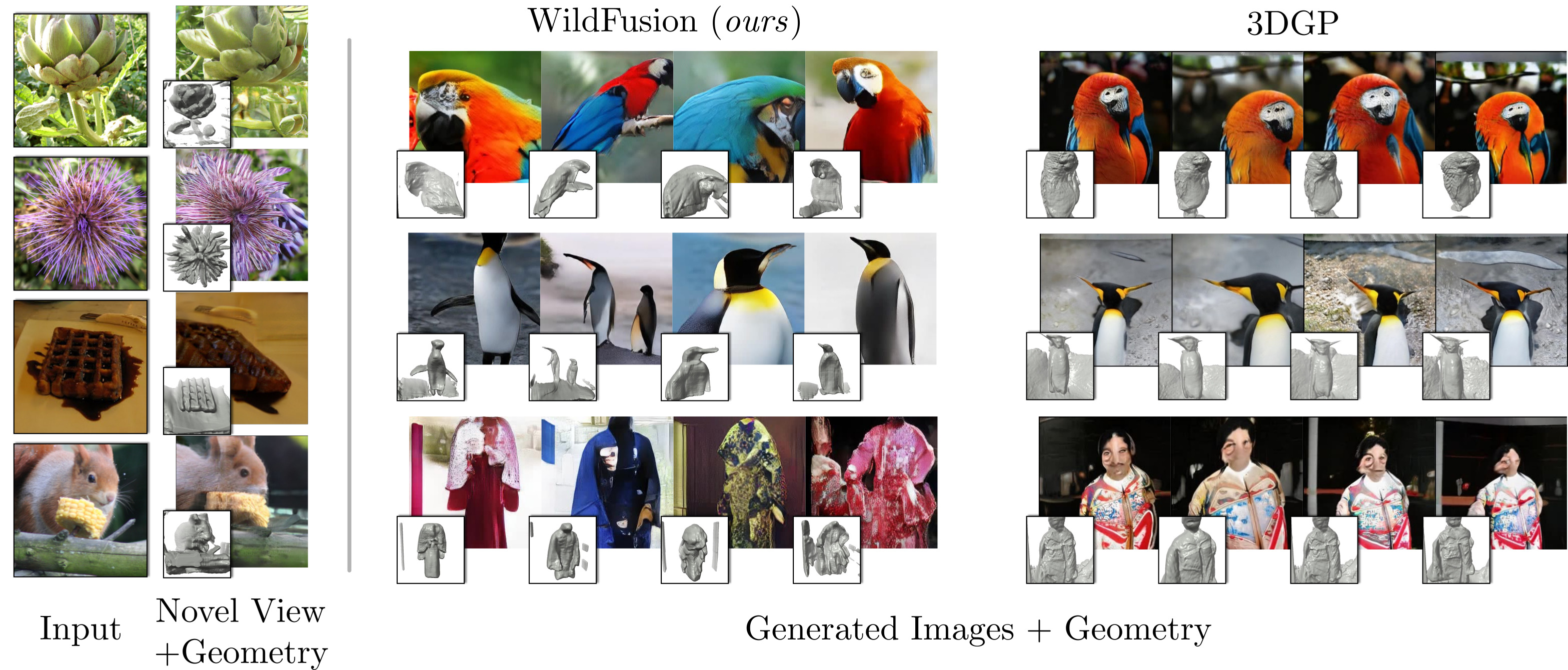}
  \vspace{-2mm}
  \caption{\small \textbf{WildFusion:} \textit{Left:} Input images, novel views and geometry from first-stage autoencoder. \textit{Right:} Novel samples and geometry from our second-stage latent diffusion model and 3DGP~\citep{skorokhodov20233dgp} for the ImageNet classes "macaw" (\textit{top}), "king penguin" (\textit{middle}), "kimono" (\textit{bottom}).
  Included videos for more results.\looseness=-1}
  \label{fig:teaser}
  \vspace{-7mm}
\end{figure*}
}

\newcommand{\system}{
\begin{figure*}[t]
\vspace{-8mm}
\centering
  \includegraphics[width=\linewidth]{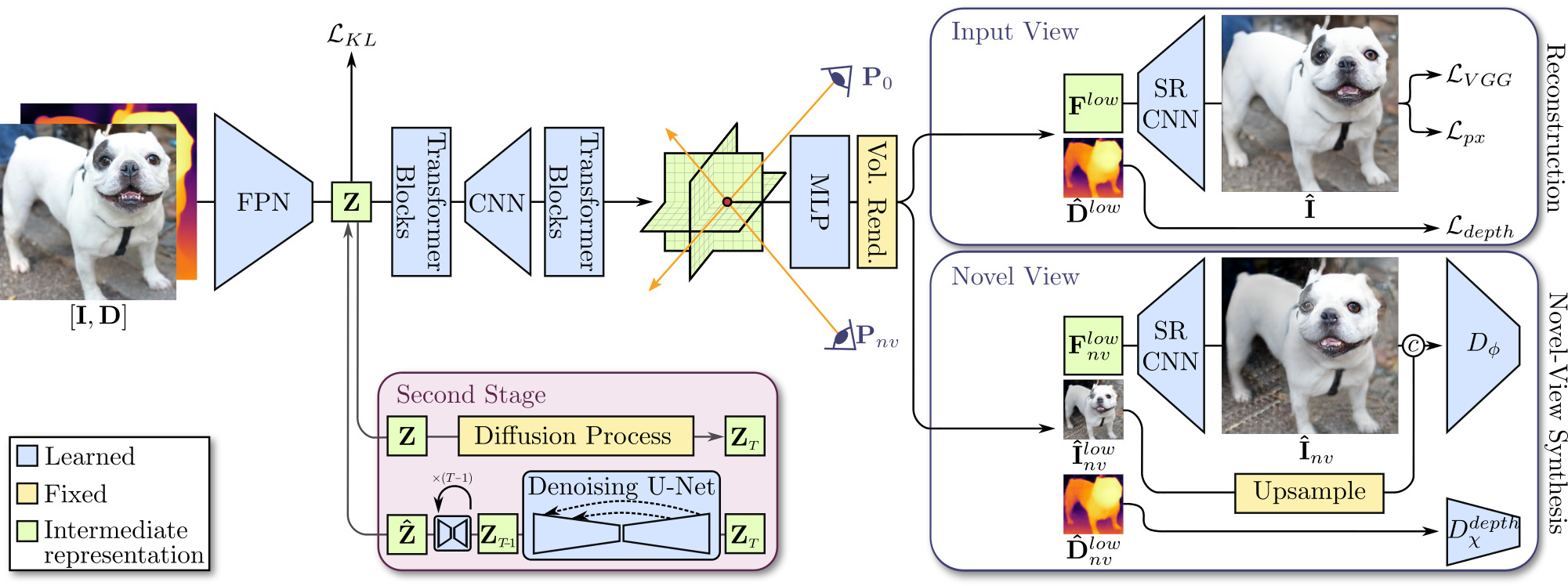}
  \vspace{-4mm}
  \caption{\small \textbf{\ourmodel Overview:} In the first stage, we train an autoencoder for both compression and novel-view synthesis. A Feature Pyramid Network (FPN)~\citep{Lin2017CVPRb} encodes a given unposed image $\mathbf{I}$ into an 3D-aware latent representation $\mathbf{Z}$, constructed as a 2D feature grid. A combination of transformer blocks and a CNN then decode $\mathbf{Z}$ into a triplane representation, which is rendered from both the input view $\mathbf{P}_0$ and a novel view $\mathbf{P}_{nv}$. As we model instances in view space, $\mathbf{P}_0$ is a fixed, pre-defined camera pose. The input view is supervised with reconstruction losses. Adversarial training provides supervision for novel views. In the second stage, a latent diffusion model is trained on the learned latent space to obtain a 3D-aware generative model.\looseness=-1} 
  \label{fig:system}
  \vspace{-5mm}
\end{figure*}
}

\newcommand{\camsystem}{
\begin{figure*}[t]
\vspace{-8mm}
\centering
  \includegraphics[width=0.99\linewidth]{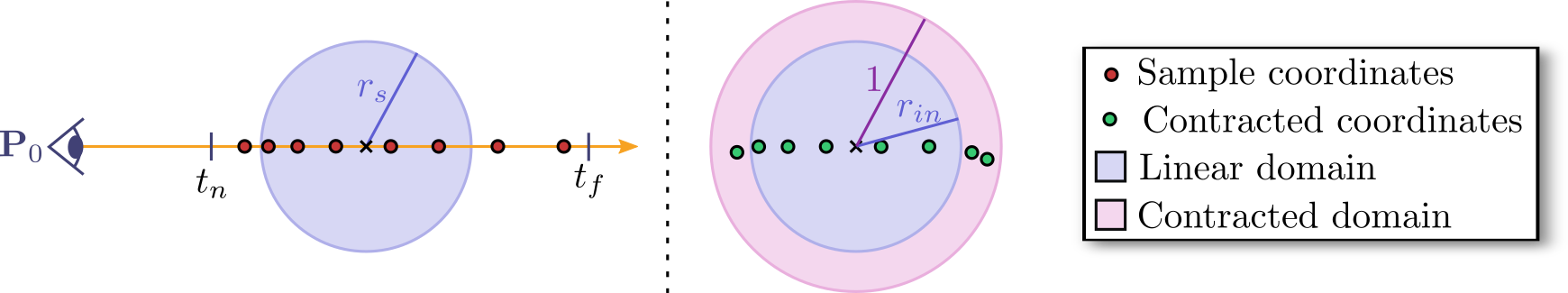}
  \caption{\small \textbf{Camera System:} Our autoencoder models objects in view space, \ie, it uses a fixed camera pose $P_0$ to render the reconstructed images. The orange line represents a camera ray and $t_n$ and $t_f$ denote the near and the far plane, respectively, between which the 3D object is located. We evaluate the camera ray at the red sample coordinates which are spaced linearly in disparity (inverse depth) to better model large depth ranges. However, the triplanes that carry the features for encoding the 3D object are only defined within a normalized range ($[-1,1]$ in all directions); hence, we need to normalize the samples on the camera ray accordingly to ensure that all coordinates are projected onto the triplanes. Specifically, we use the contraction function in Eq.~(\ref{eq:contraction_function}). Sample coordinates within a fixed radius $r$ are mapped linearly to a sphere of radius $r_{in} < 1$, \ie, into the domain where the triplanes are defined (linear domain). Sample coordinates with norm $>r$ are contracted such that they have norm $\leq 1$ in the domain of the triplanes (contracted domain).} 
  \label{fig:camsystem}
\vspace{-3mm}
\end{figure*}
}

\newcommand{\aeresults}{
\begin{figure*}[t]
\vspace{-8mm}
\centering
  \includegraphics[width=\linewidth]{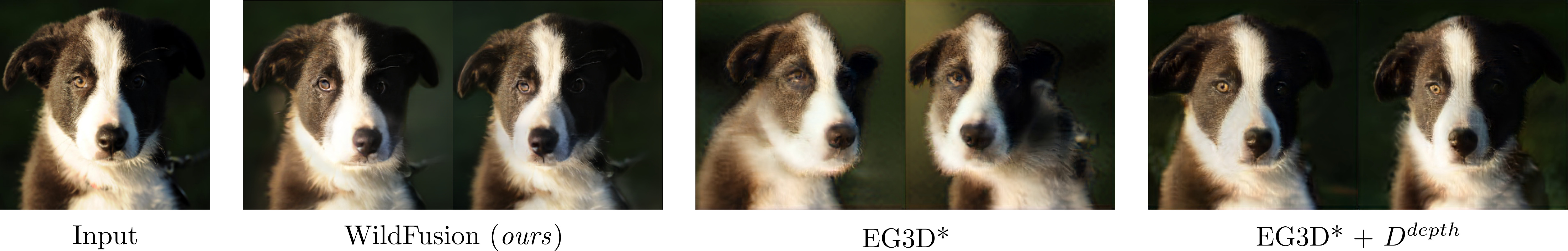}
  \vspace{-7mm}
  \caption{\small Baseline comparisons for novel view synthesis on images unseen during training. Shown are the input image and two novel views per method. Viewpoints across methods are the same. Included video for more results.\looseness=-1} 
  \label{fig:aeresults}
\end{figure*}
}

\newcommand{\ldmresults}{
\begin{figure*}[t]
\vspace{-8mm}
  \includegraphics[width=\textwidth]{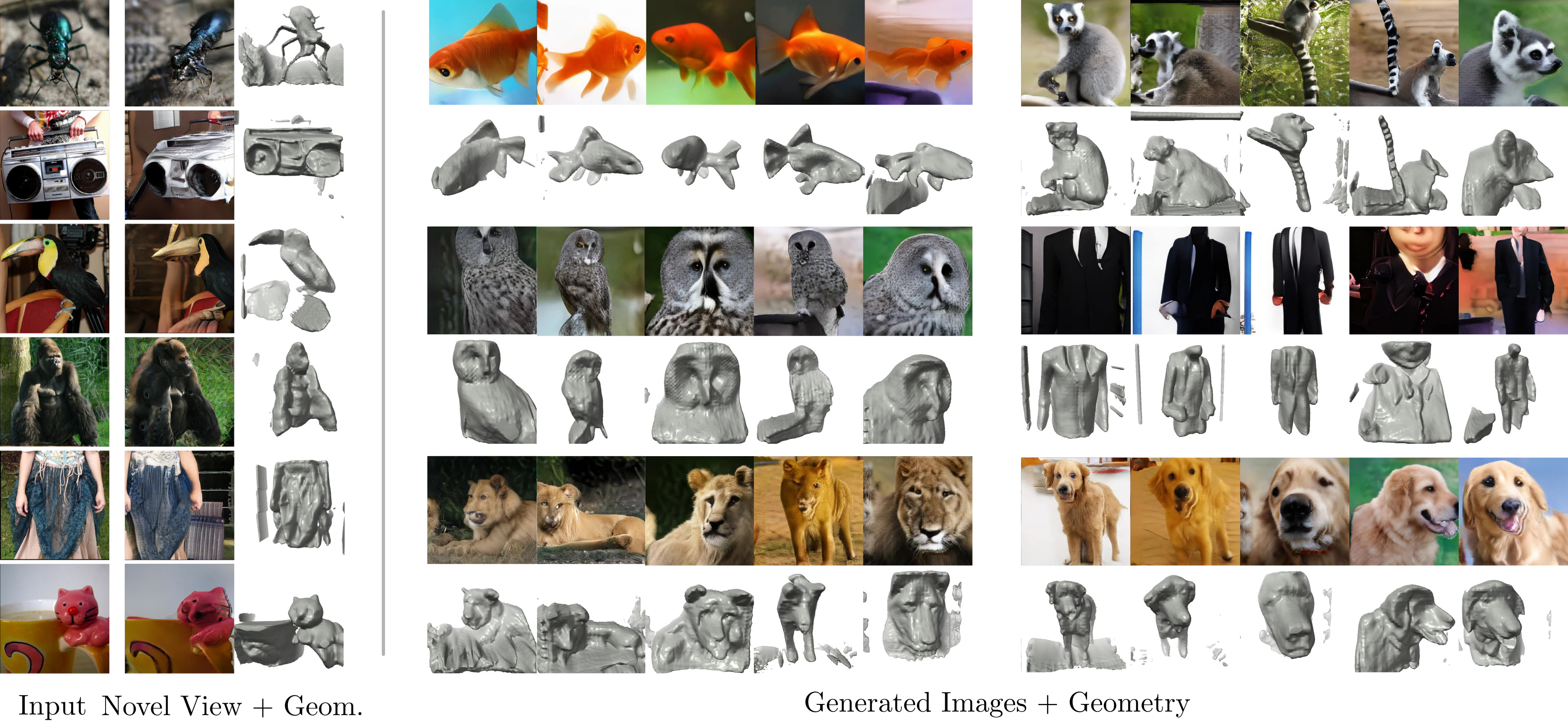}
  \vspace{-5mm}
  \caption{\small Generated 3D-aware image samples and geometry by WildFusion. Included videos for more results.} 
  \label{fig:ldmresults}
\end{figure*}
}

\newcommand{\ablation}{
\begin{figure*}[p]
\centering
  \includegraphics[width=\linewidth]{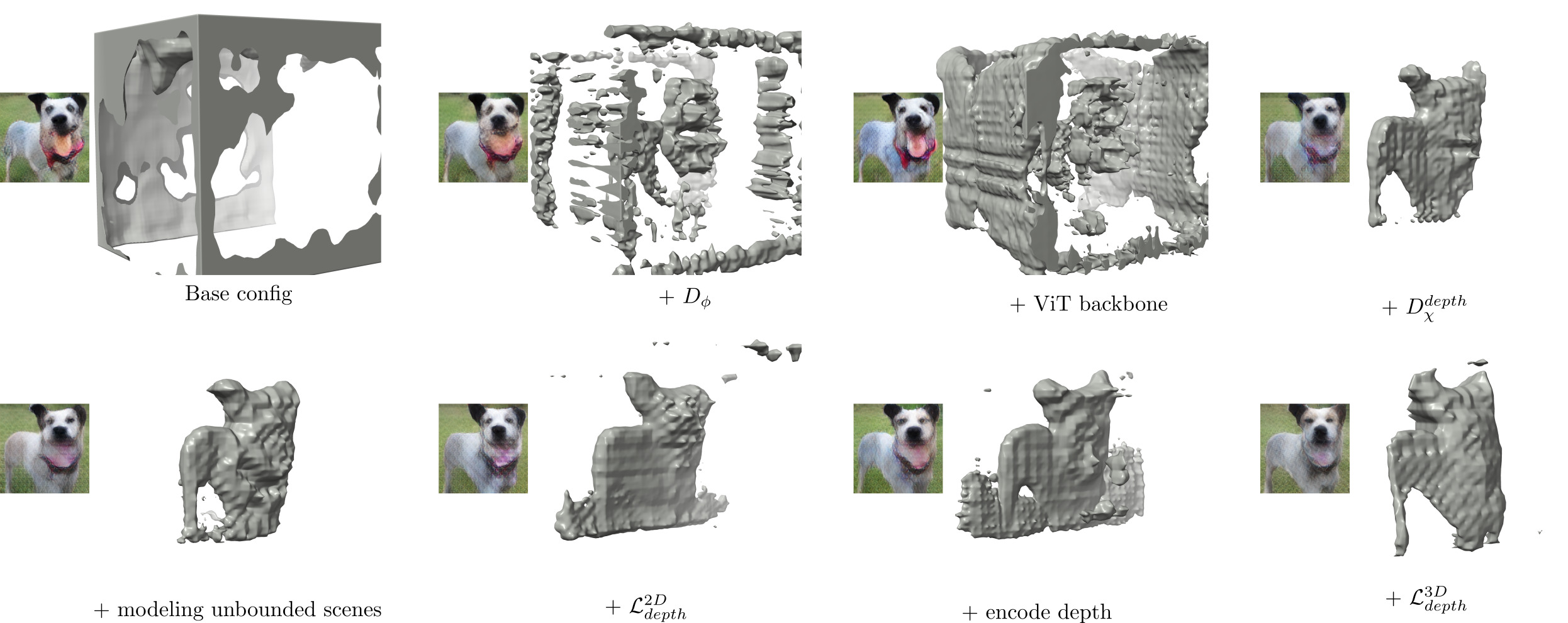}
  \caption{\small Reconstructions (small) and geometry for different settings in the ablation study. The geometry was extracted by applying marching cubes to the density values of the feature field. We can see an improvement in geometry, as more components are added to the model. Note that the underlying model for this experiment is very small and was used only for the ablation study. It has limited expressivity.} 
  \label{fig:ablation}
\end{figure*}
}

\newcommand{\aeresultssupp}{
\begin{figure*}[t]
\centering
  \includegraphics[width=\linewidth]{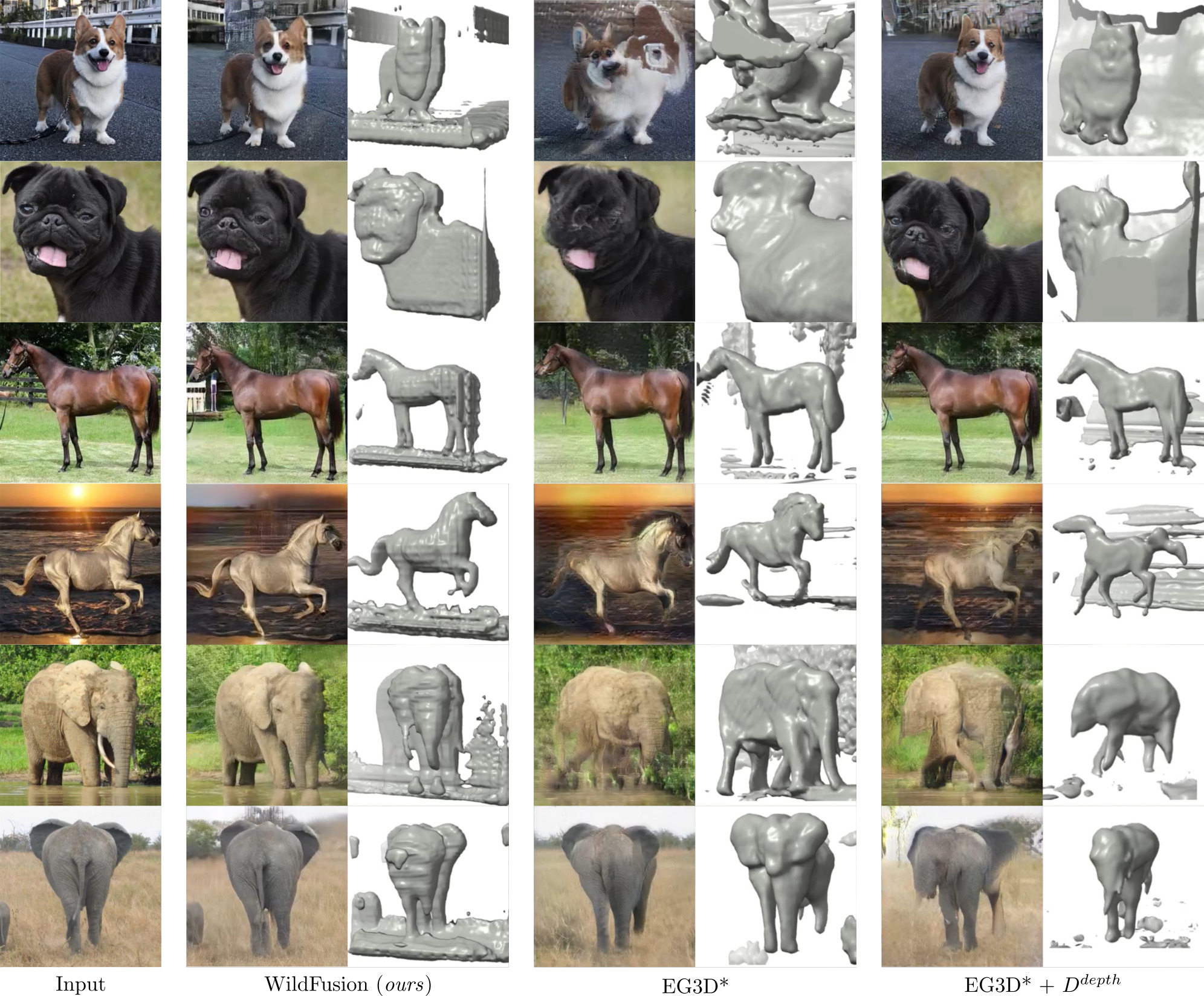}
  \vspace{-5mm}
  \caption{\small Comparison with baselines for novel view synthesis on images unseen during training. Shown are the input image, a novel view and the geometry extracted with marching cubes. The viewpoints across methods are the same. See the included video for more results.} 
  \label{fig:aeresultssupp}
\end{figure*}
}

\newcommand{\aeresultsoursdogs}{
\begin{figure*}[p]
\centering
  \includegraphics[width=\linewidth]{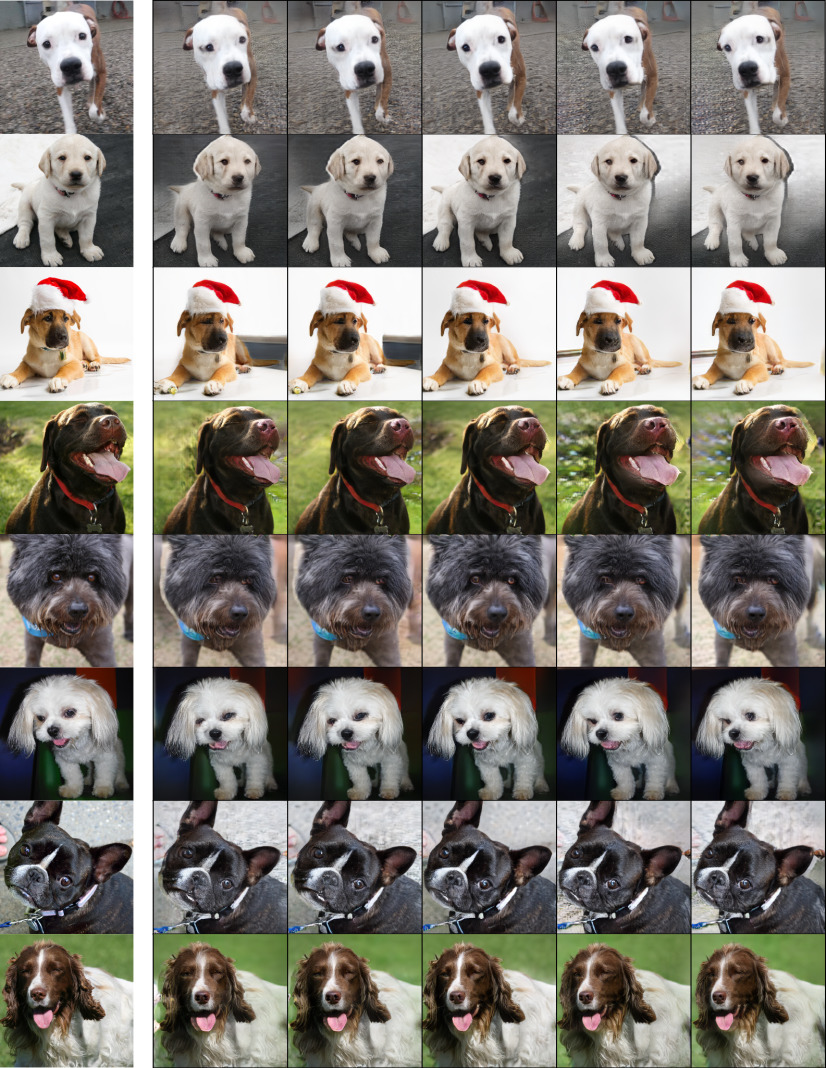}
  \vspace{-5mm}
  \caption{\small Input images (left column) and novel views from \ourmodel's 3D-aware autoencoder for SDIP Dogs. The results span a yaw angle of $40^\circ$.} 
  \label{fig:aeresultssupp_dog}
\end{figure*}
}

\newcommand{\aeresultsoursele}{
\begin{figure*}[p]
\centering
  \includegraphics[width=\linewidth]{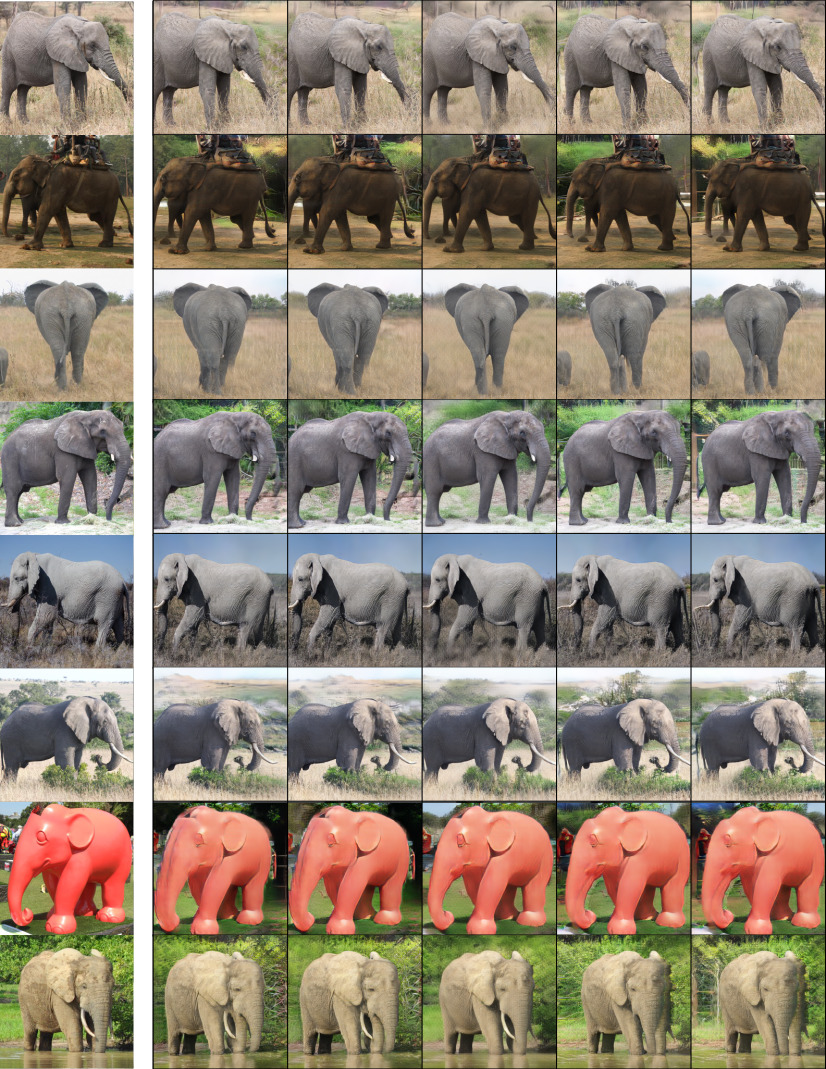}
  \vspace{-5mm}
  \caption{\small Input images (left column) and novel views from \ourmodel's 3D-aware autoencoder for SDIP Elephants. The results span a yaw angle of $40^\circ$.} 
  \label{fig:aeresultssupp_ele}
\end{figure*}
}

\newcommand{\aeresultsourshorse}{
\begin{figure*}[p]
\centering
  \includegraphics[width=\linewidth]{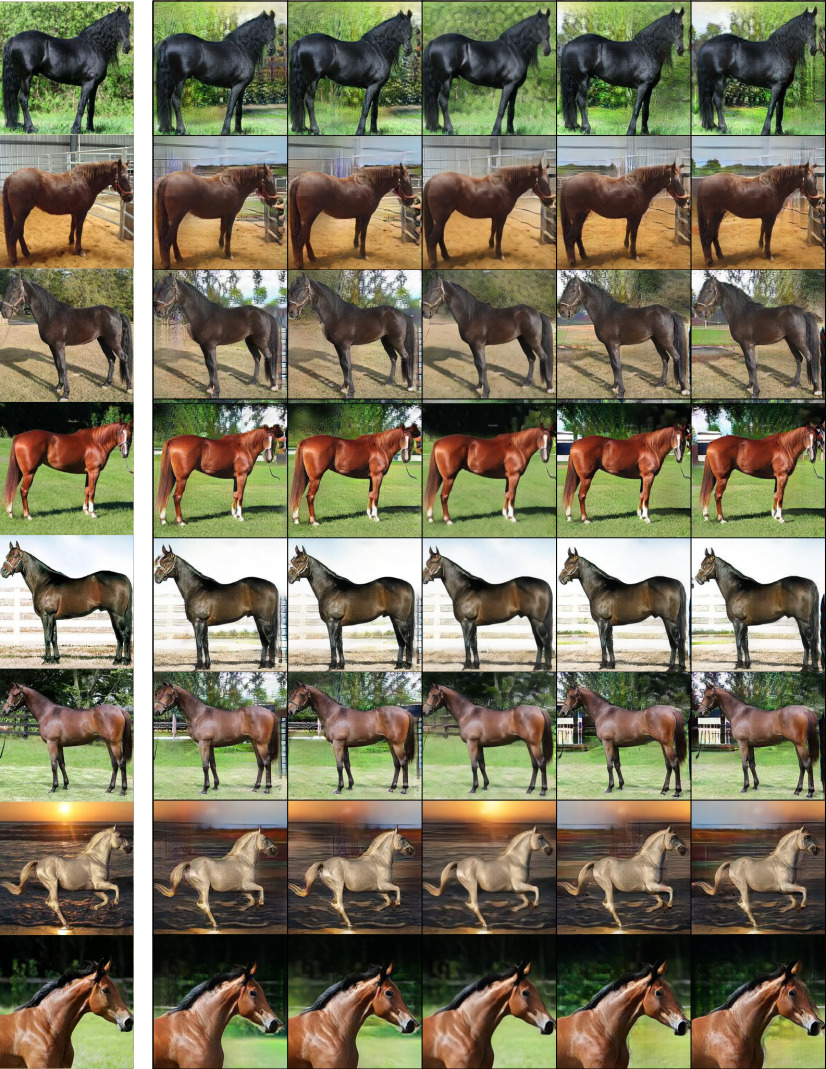}
  \vspace{-5mm}
  \caption{\small Input images (left column) and novel views from \ourmodel's 3D-aware autoencoder for SDIP Horses. The results span a yaw angle of $40^\circ$.} 
  \label{fig:aeresultssupp_horse}
\end{figure*}
}

\newcommand{\aeresultsoursimagenet}{
\begin{figure*}[p]
\centering
  \includegraphics[width=\linewidth]{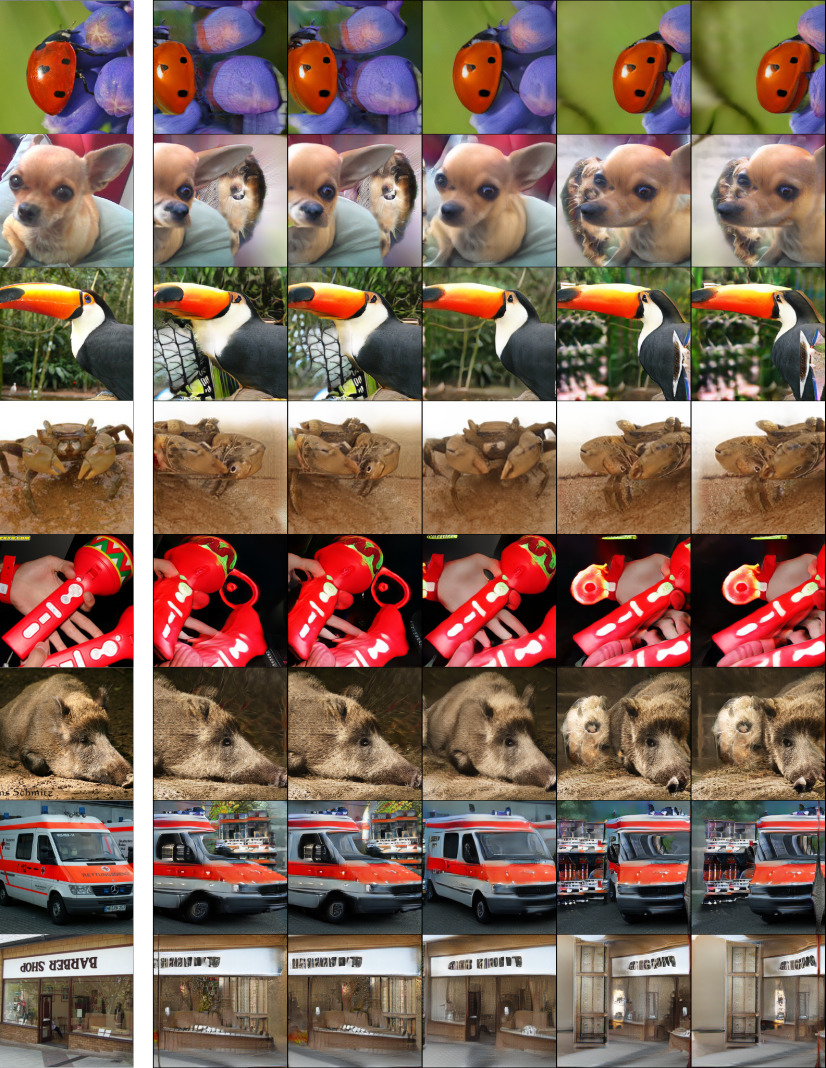}
  \vspace{-5mm}
  \caption{\small Input images (left column) and novel views from \ourmodel's 3D-aware autoencoder for ImageNet. The results span a yaw angle of $40^\circ$.} 
  \label{fig:aeresultssupp_imagenet}
\end{figure*}
}

\newcommand{\ldmresultsoursdog}{
\begin{figure*}[p]
\centering
  \includegraphics[width=\linewidth]{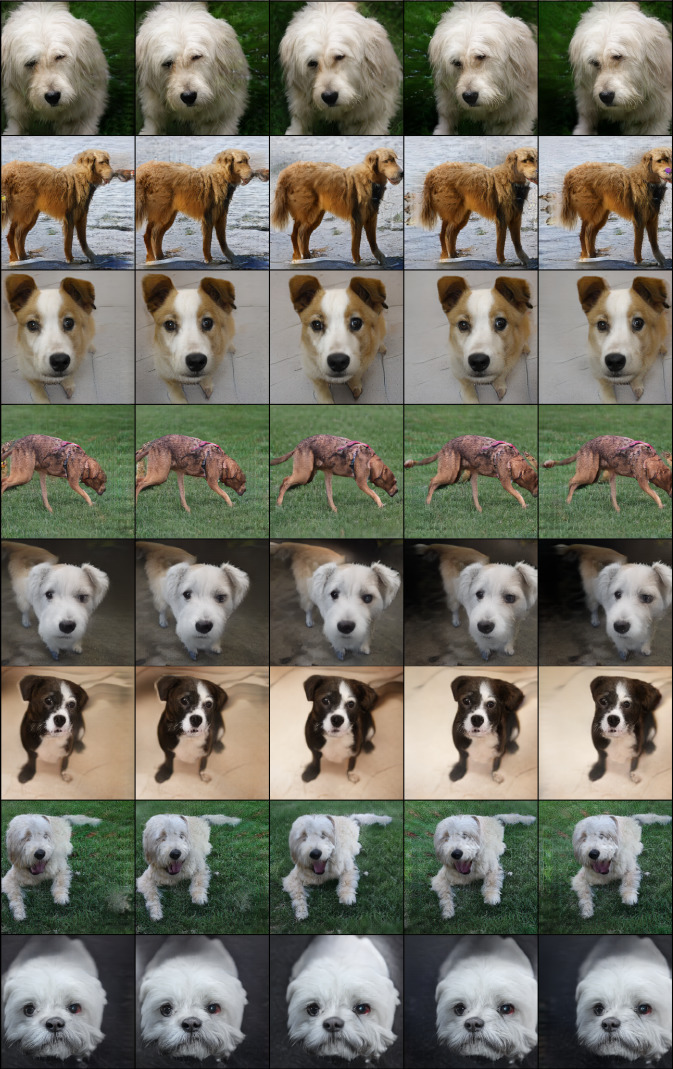}
  \vspace{-5mm}
  \caption{\small Generated images and novel views from \ourmodel for SDIP Dogs. The results span a yaw angle of $40^\circ$.} 
  \label{fig:ldmresultssupp_dog}
\end{figure*}
}

\newcommand{\ldmresultsourshorse}{
\begin{figure*}[p]
\centering
  \includegraphics[width=\linewidth]{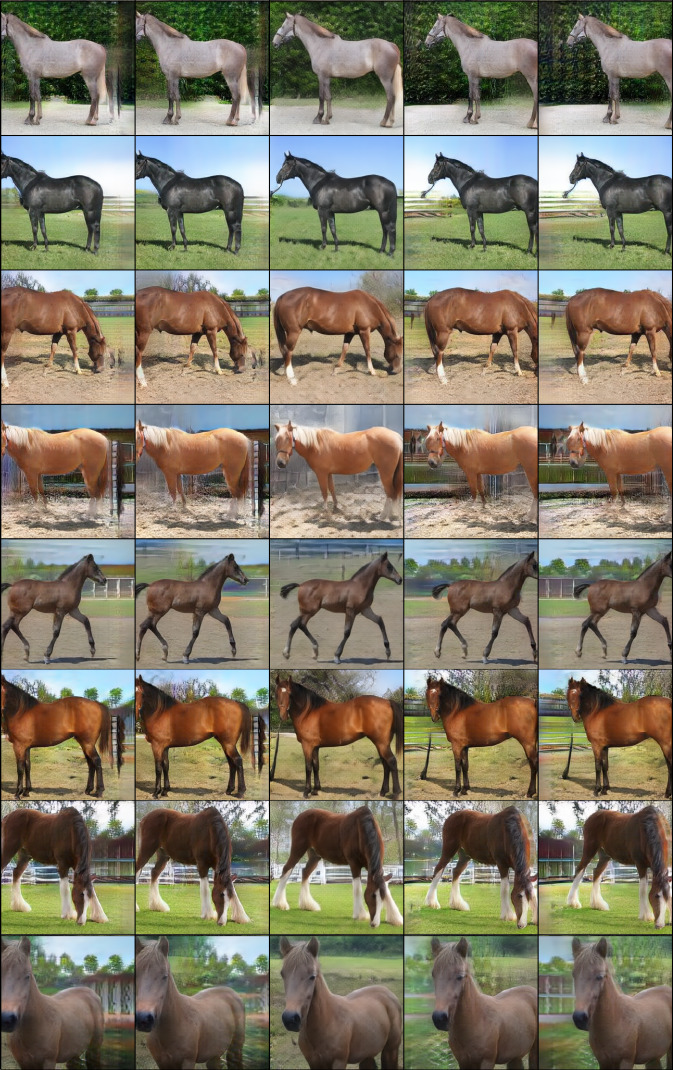}
  \vspace{-5mm}
  \caption{\small Generated images and novel views from \ourmodel for SDIP Horses. The results span a yaw angle of $40^\circ$.} 
  \label{fig:ldmresultssupp_horse}
\end{figure*}
}

\newcommand{\ldmresultsoursele}{
\begin{figure*}[p]
\centering
  \includegraphics[width=\linewidth]{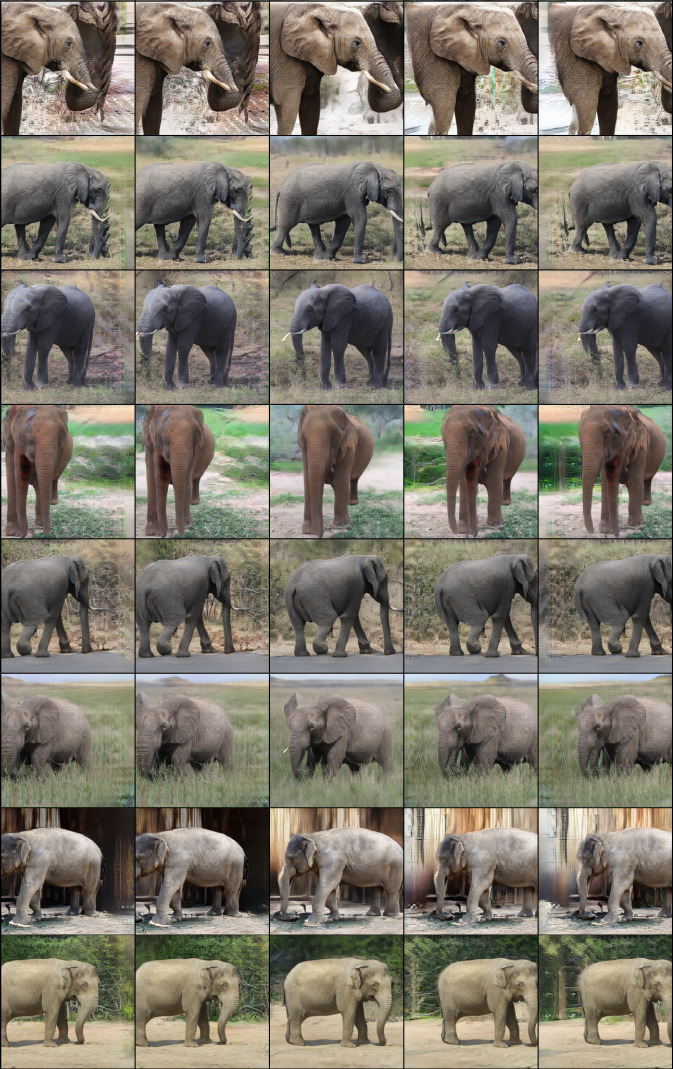}
  \vspace{-5mm}
  \caption{\small Generated images and novel views from \ourmodel for SDIP Elephants. The results span a yaw angle of $40^\circ$.} 
  \label{fig:ldmresultssupp_ele}
\end{figure*}
}

\newcommand{\ldmresultsoursimagenet}{
\begin{figure*}[p]
\centering
  \includegraphics[width=\linewidth]{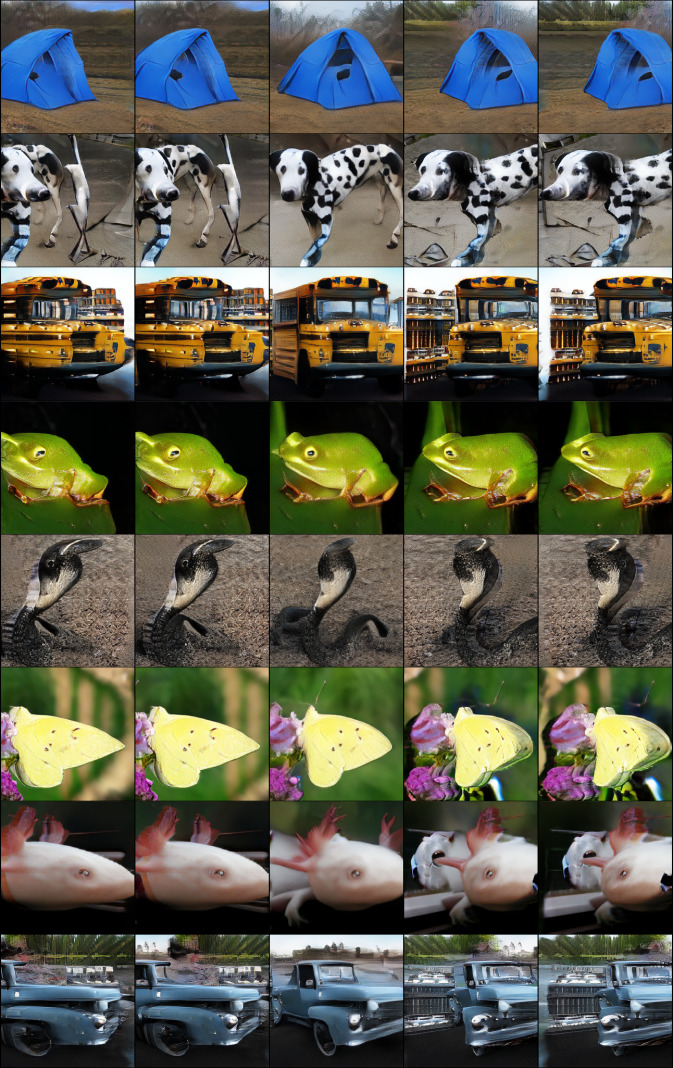}
  \vspace{-5mm}
  \caption{\small Generated images and novel views from \ourmodel for ImageNet. The results span a yaw angle of $40^\circ$.} 
  \label{fig:ldmresultssupp_imagenet}
\end{figure*}
}

\newcommand{\imagenetcomp}{
\begin{figure*}[t]
\vspace{-7mm}
\centering
  \includegraphics[width=\linewidth]{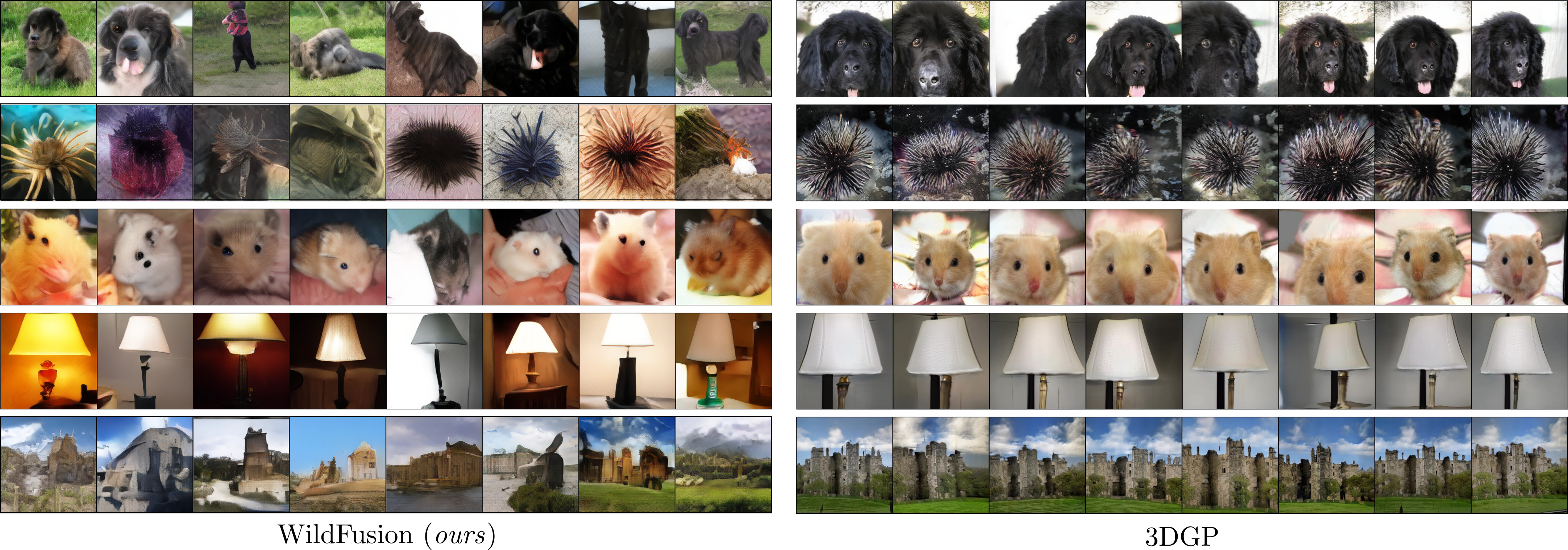}
  \vspace{-6mm}
  \caption{\small \textbf{Sample Diversity:} Generated samples on ImageNet. Rows indicate class; columns show uncurated random samples. While \ourmodel generates diverse samples due to its diffusion model-based framework (\textit{left}), the GAN-based 3DGP~\citep{skorokhodov20233dgp} has very low intra-class diversity (mode collapse, \textit{right}).} 
  \label{fig:imagenet_comp}
  \vspace{-4mm}
\end{figure*}
}

\newcommand{\imagenet}{
\begin{figure*}[t]
\centering
  \includegraphics[width=\linewidth]{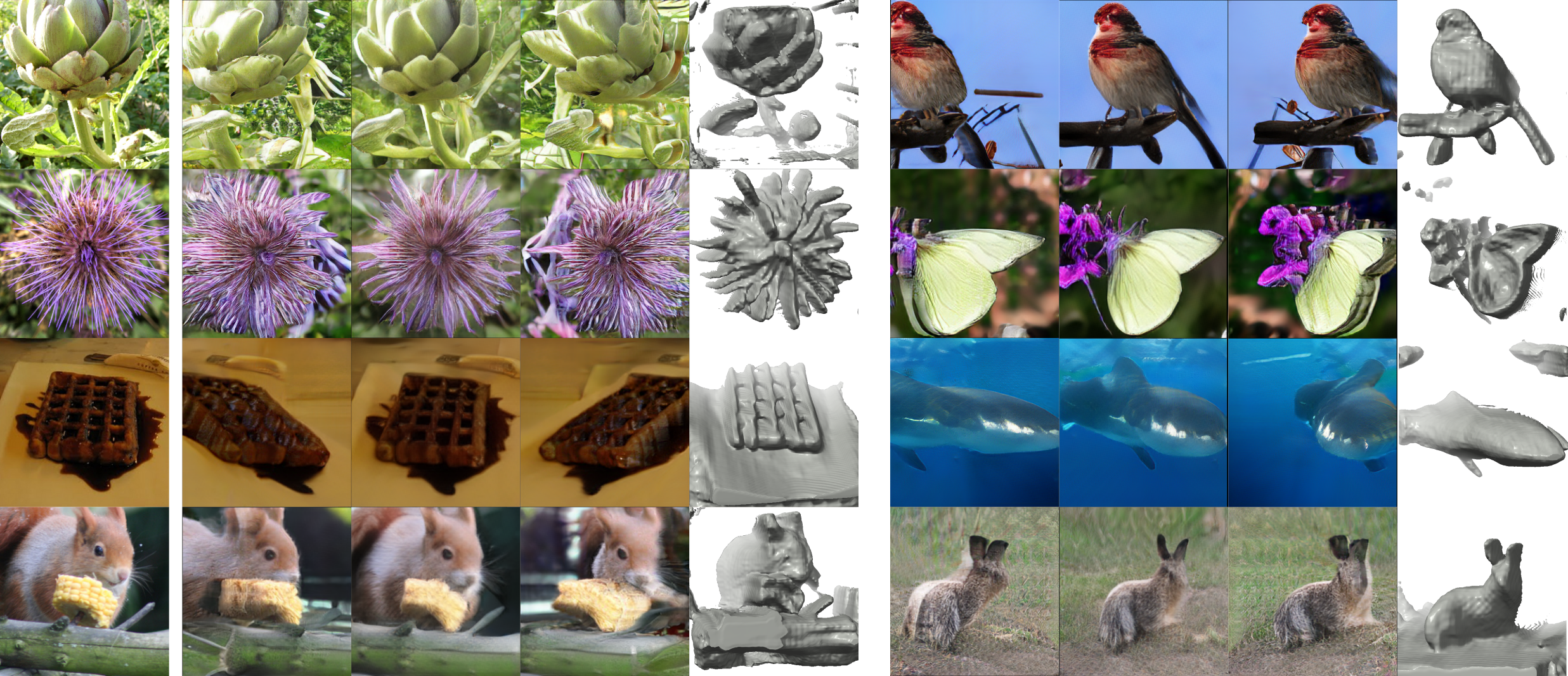}
  \vspace{-5mm}
  \caption{\small Preliminary Results on ImageNet. \textit{Left:} Input images, three novel views, and geometry from the first-stage autoencoder. \textit{Right:} Novel 3D-consistent samples synthesized from the second-stage latent diffusion model.} 
  \label{fig:imagenet}
\end{figure*}
}

\newcommand{\interpolation}{
\begin{figure}[t]
  \vspace{-0.9cm}
  \begin{minipage}[c]{0.67\textwidth}
    \includegraphics[width=1.0\textwidth]{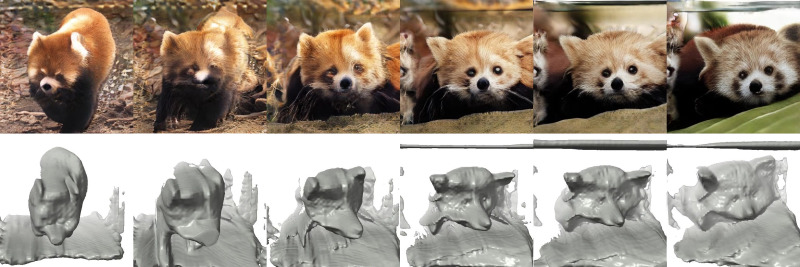}
  \end{minipage}\hfill
  \begin{minipage}[c]{0.3\textwidth}
  \vspace{0.3cm}
    \caption{\small \textbf{3D-Aware Image Interpolation.} We encode two images into latent space (far left and far right), further encode into the diffusion model's Gaussian prior space (inverse DDIM), interpolate the resulting encodings, and generate the corresponding 3D images along the interpolation path.\looseness=-1}  
    \label{fig:interpolation}
  \end{minipage}
  \vspace{-4.0mm}
\end{figure}
}

\newcommand{\denoise}{
\begin{figure}[t]
  \vspace{-0.5cm}
  \begin{minipage}[c]{0.67\textwidth}
    \includegraphics[width=1.0\textwidth]{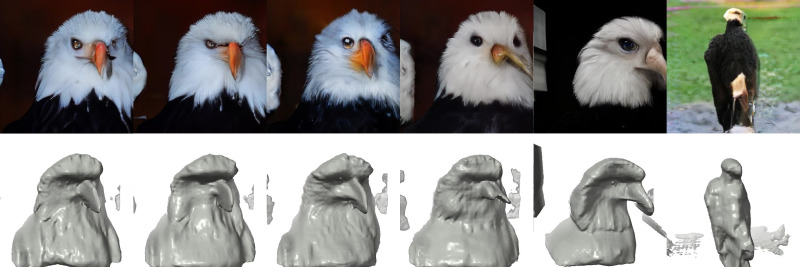}
  \end{minipage}\hfill
  \begin{minipage}[c]{0.3\textwidth}
  \vspace{0.3cm}
    \caption{\small \textbf{3D-Aware Generative Image Resampling.} Given an image (far left), we forward diffuse its latent encoding for varying numbers of steps and re-generate from the partially noised encodings. Depending on how far we diffuse, we obtain varying levels of generative image resampling.\looseness=-1}  
    \label{fig:denoise}
  \end{minipage}
\end{figure}
}

\newcommand{\inpainting}{
\begin{figure*}[t]
\centering
  \includegraphics[width=\linewidth]{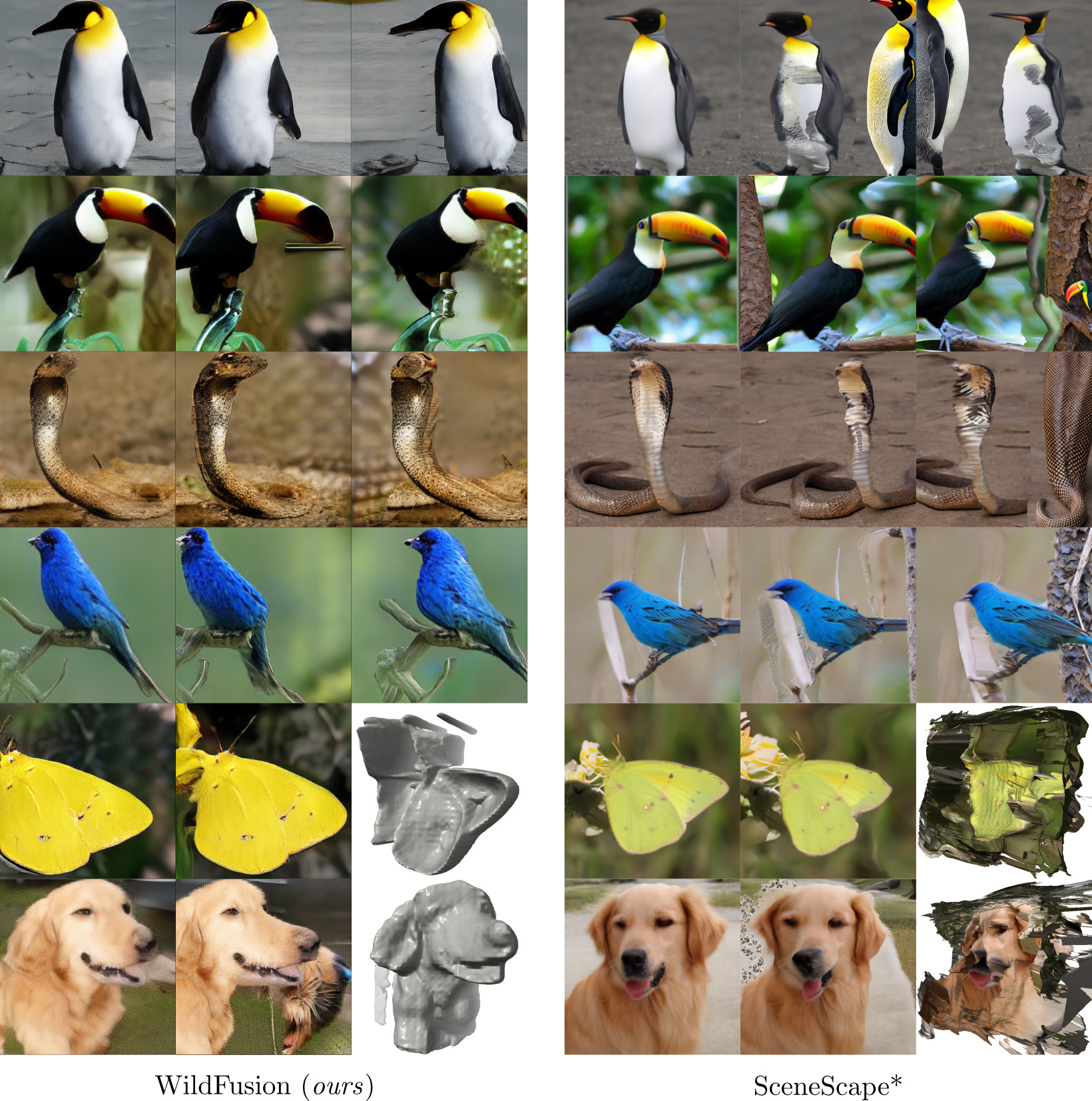}
  \caption{\small We compare WildFusion against a variant of SceneScape that combines a 2D generative model with a pre-trained inpainting model. \textit{First for rows}: Leftmost image is input view and next two images are novel views at $\pm17$ degree yaw angles for the two methods. We observe severe inpainting inconsistencies for the SceneScape baseline. \textit{Last two rows}: Leftmost image is again input image, next image is a novel view at $-17$ degree yaw angle, and the last image shows the extracted geometry/mesh for the two methods). We find that due to the inconsistencies of the inpainting model across views, the fused meshes for SceneScape have severe intersections and overall inferior geometry to WildFusion.
  } 
  \label{fig:inpainting}
    \vspace{-6mm}
\end{figure*}
}

\newcommand{\imagenetcompappendix}{
\begin{figure*}[p]
\vspace{-7mm}
\centering
  \includegraphics[width=\linewidth]{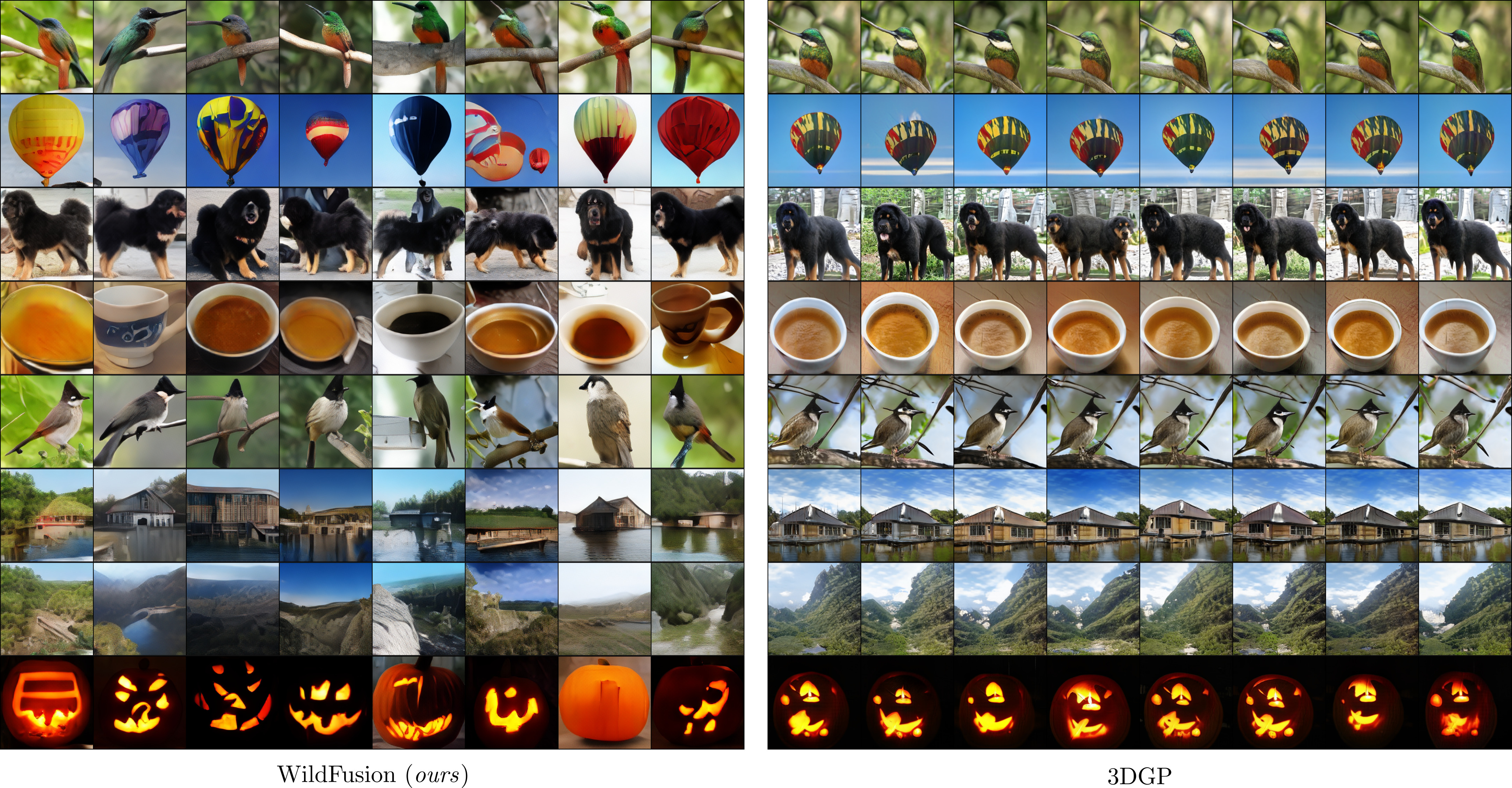}
  \vspace{-6mm}
  \caption{\small \textbf{Sample Diversity:} Generated samples on ImageNet. Rows indicate class; columns show uncurated random samples. While \ourmodel generates diverse samples due to its diffusion model-based framework (\textit{left}), the GAN-based 3DGP~\citep{skorokhodov20233dgp} has very low intra-class diversity (mode collapse, \textit{right}).} 
  \label{fig:diversity}
\end{figure*}
}

\maketitle

\blfootnote{$^\dagger$Part of the work was done during an internship at NVIDIA.}

\vspace{-4mm}
\begin{abstract}
Modern learning-based approaches to 3D-aware image synthesis achieve high photorealism and 3D-consistent viewpoint changes for the generated images. 
Existing approaches represent instances in a shared canonical space. 
However, for in-the-wild datasets a shared canonical system can be difficult to define or might not even exist.
In this work, we instead model instances in \textit{view space}, alleviating the need for posed images and learned camera distributions. 
We find that in this setting, existing GAN-based methods are prone to generating flat geometry and struggle with distribution coverage.
We hence propose \textit{\ourmodel}, a new approach to 3D-aware image synthesis based on latent diffusion models (LDMs). 
We first train an autoencoder that infers a compressed latent representation, which additionally captures the images' underlying 3D structure and enables not only reconstruction but also novel view synthesis. 
To learn a faithful 3D representation, we leverage cues from monocular depth prediction. 
Then, we train a diffusion model in the 3D-aware latent space, thereby enabling synthesis of high-quality 3D-consistent image samples, outperforming recent state-of-the-art GAN-based methods.
Importantly, our 3D-aware LDM is trained without any direct supervision from multiview images or 3D geometry and does not require posed images or learned pose or camera distributions. It directly learns a 3D representation without relying on canonical camera coordinates. This opens up promising research avenues for scalable 3D-aware image synthesis and 3D content creation from in-the-wild image data. See
\url{https://katjaschwarz.github.io/wildfusion/} for videos of our 3D results.\looseness=-1
\end{abstract}

\section{Introduction}
\vspace{-1mm}
High-quality 2D and 3D content creation is of great importance for virtual reality, animated movies, gaming and robotics simulation. 
In the past years, deep generative models have demonstrated immense potential, enabling photorealistic image synthesis at high resolution~\citep{Goodfellow2014NIPS,Karras2020CVPRa,Karras2021NEURIPS,rombach2021highresolution,Dhariwal2021NIPS,Sauer2022ARXIV}.
Recently, 3D-aware generative models advanced image synthesis to view-consistent, 3D-aware image generation~\citep{Schwarz2020NIPS,Chan2020CVPR,Gu2021ARXIV,Jo2021ARXIV,Xu2021ARXIV,Zhou2021ARXIV,Zhang2021ARXIV,OrEl2021ARXIV,Xu2021NEURIPS,Pan2021NEURIPS,DENG2021ARXIV,xiang2023gramhd,Schwarz2022NIPS,Chan2022CVPR}. They generate images with explicit control over the camera viewpoint. 
Importantly, these approaches do not require 3D training data or multiview supervision, which is costly or impossible to obtain for large-scale real-world data.\teaser

While existing 3D-aware generative models achieve high photorealism and 3D-consistent viewpoint control, the vast majority of approaches only consider single-class and aligned data like human faces~\citep{Karras2020CVPRa} or cat faces~\citep{Choi2020CVPR}.
The reason for this is that existing methods assume a shared canonical coordinate system to 
represent 3D objects.
As a consequence, they require either poses from an off-the-shelf pose estimator~\citep{Chan2022CVPR,Schwarz2022NIPS,Gu2021ARXIV,xiang2023gramhd} or assume, and sometimes learn to refine, a given pose distribution~\citep{Schwarz2020NIPS,Niemeyer2021THREEDV,Shi2023CVPR,skorokhodov20233dgp,Skorokhodov2022ARXIV}. 
In contrast, in-the-wild images typically have no clearly defined canonical camera system and camera poses or pose distributions are not available or very challenging to obtain.\looseness=-1 

We propose to instead model instances in \textit{view space}:
Our coordinate system is viewer-centric, \ie, it parameterizes the space as seen from the camera's point of view. This removes the need for camera poses and a priori camera pose distributions, unlocking 3D-aware image synthesis on unaligned, diverse datasets. 
We identify crucial challenges for training 
in view space: For complex datasets, without a shared canonical representation, existing techniques are prone to generating poor 3D representations and the GAN-based methods struggle with distribution coverage (see \figref{fig:imagenet_comp}, \tabref{tab:baselinecomp}).\looseness=-1

To prevent generating flat 3D representations, we leverage cues from monocular depth prediction. While monocular depth estimators are typically trained with multi-view data, we leverage an off-the-shelf pretrained model, such that our approach does not require any \textit{direct} multi-view supervision for training. Recent works \citep{Bhat2023ARXIV,Eftekhar2021ICCV,Ranftl2020PAMI,Miangoleh2021CVPR} demonstrate high prediction quality and generalization ability to in-the-wild data and have been successfully applied to improve 3D-reconstruction~\citep{Yu2022MonoSDF}.
To ensure distribution coverage on more diverse datasets, we build our approach upon denoising diffusion-based generative models (DDMs)~\citep{ho2020ddpm2,sohl2015deep,song2020score}. DDMs have shown state-of-the-art 2D image synthesis quality and offer a scalable and robust training objective~\citep{Nichol2021ICML,Nichol2022ICML,rombach2021highresolution,dhariwal2021diffusion,Ramesh2022ARXIV,saharia2022photorealistic,balaji2022eDiffi,Ho2022MLRes}.
More specifically, we develop a 3D-aware generative model based on latent diffusion models (LDMs)~\citep{rombach2021highresolution,Vahdat2021NIPS}. 
By training an autoencoder first and then modeling encoded data in a compressed latent space,
LDMs achieve an excellent trade-off between computational efficiency and quality. 
Further, their structured latent space can be learnt to capture a 3D representation of the modeled inputs, as we show in this work.\looseness=-1

Our 3D-aware LDM, called \textit{\ourmodel}, follows LDMs' two-stage approach: First, we train a powerful 3D-aware autoencoder from large collections of unposed images without multiview supervision that simultaneously performs both compression \textit{and} enables novel-view synthesis.
The autoencoder is trained with pixel-space reconstruction losses on the input views and uses adversarial training to supervise novel views. Note that by using adversarial supervision for the novel views, our autoencoder is trained for novel-view synthesis without the need for multiview supervision, in contrast to previous work~\citep{Watson2022ARXIV,chan2023genvs,liu2023zero1to3}. 
Adding monocular depth cues helps the model learn a faithful 3D representation and further improves novel-view synthesis. 
In the second stage, we train a diffusion model in the compressed and 3D-aware latent space, which enables us to synthesize novel samples and turns the novel-view synthesis system, \ie, our autoencoder, into a 3D-aware generative model. 
We validate 
\ourmodel
on multiple image generation benchmarks, including ImageNet, and find that it outperforms recent state-of-the-art 3D-aware GANs. Moreover, we show that our autoencoder is able to directly synthesize high-quality novel views for a given single image and performs superior compared to recent GAN-based methods, which usually require an inversion process to embed a given image into their latent space~\citep{abdal2019image2stylegan,richardson2021encoding,tov2021designing,zhu2020domain,RoichTOG2023}. Further, in contrast to inversion methods, our autoencoder is trained in a single stage and does not require a pretrained 3D-aware GAN as well as elaborate and often slow techniques for latent optimization.\looseness=-1

Main contributions: 
\textbf{(i)}~We remove the need for posed images and a priori camera pose distributions for 3D-aware image synthesis by modeling instances in \textit{view space} instead of canonical space.
\textbf{(ii)}~We learn a powerful 3D-aware autoencoder from unposed images without multiview supervision that simultaneously performs compression, while inferring a 3D representation suitable for novel-view synthesis.
\textbf{(iii)}~We show that our novel 3D-aware LDM, \ourmodel, enables high-quality 3D-aware image synthesis with reasonable geometry and strong distribution coverage, achieving state-of-the-art performance in the unposed image training setting, which corresponds to training on in-the-wild image data. Moreover, we can more efficiently perform novel view synthesis for given images than common GAN-based methods and explore promising 3d-aware image manipulation techniques. 
We hope that \ourmodel paves the way towards scalable and robust in-the-wild 3D-aware image synthesis.\looseness=-1

\imagenetcomp
\section{Related Work}
\vspace{-1mm}
We briefly provide the most relevant related works in this section, and present a comprehensive discussion of the theoretical fundamentals and the related literature, including concurrent works, in App.~\ref{app:related_work}.\looseness=-1

Most works on 3D-aware image synthesis rely on GANs~\citep{Goodfellow2014NIPS}
and focus on aligned datasets with well-defined pose distributions.
For instance, POF3D~\citep{Shi2023CVPR} 
infers camera poses and works in a canonical view space; it has been used only for datasets with simple pose distributions, such as cat and human faces. 
To enable training on more complex datasets, 3DGP~\citep{skorokhodov20233dgp} proposes an elaborate camera model and learns to refine an initial prior on the pose distribution. Specifically, 3DGP predicts the camera location in a canonical coordinate system per class and sample-specific camera rotation and intrinsics. This assumes that samples within a class share a canonical system, and we observe that learning this complex distribution can aggravate training instability. Further, the approach needs to be trained on heavily filtered training data. In contrast, WildFusion can generate high-quality and diverse samples even when trained on the entire ImageNet dataset without any filtering (see Sec. \ref{sec:expdiffusion}). 
Moreover, we use EG3D's triplanes and their dual discriminator to improve view consistency~\citep{Chan2022CVPR}. 
Note that in contrast to POF3D, 3DGP, EG3D, and the vast majority of 3D-aware image generative models, \ourmodel is not a GAN. GANs are notoriously hard to train~\citep{Mescheder2018ICML} and often do not cover the data distribution well (see mode collapse in 3DGP, Fig.~\ref{fig:imagenet_comp}). 
Instead, we explore 3D-aware image synthesis with latent diffusion models for the first time. \\
Concurrently with us, IVID~\citep{xiang2023ivid} trains a 2D diffusion model that first synthesizes an initial image and subsequently generates novel views conditioned on it. However, its iterative generation is extremely slow because it requires running the full reverse diffusion process for every novel view. Instead, \ourmodel only runs the reverse diffusion process once to generate a (latent) 3D representation.
Another concurrent work, VQ3D~\citep{sargent2023vq3d}, also proposes an autoencoder architecture, but uses sequence-like latent variables and trains an autoregressive transformer in the latent space whereas we train a diffusion model on latent feature maps.

\section{\ourmodel}
\vspace{-1mm}
\system
Our goal is to design a 3D-aware image synthesis framework that can be trained using unposed in-the-wild images. We base our framework, \ourmodel, on LDMs~\citep{rombach2021highresolution,Vahdat2021NIPS} for several reasons: \textbf{(i)} Compared to diffusion in the original data space, they offer excellent computational efficiency due to their compressed latent space. \textbf{(ii)} Diffusion models use a robust objective that offers sample diversity and does not suffer from problems that plague GANs such as mode collapse. \textbf{(iii)} Most importantly, 
one can construct a latent space that can be trained to not only perform compression but also to learn a powerful 3D representation for novel view synthesis, as we demonstrate in this work.
\figref{fig:system} shows an overview over \ourmodel. It consists of two training stages~\citep{rombach2021highresolution,Esser2021CVPR}. In the first stage, our new autoencoder learns a compressed and abstracted latent space suitable for reconstruction and novel view synthesis from single view training data. In the second stage, a latent diffusion model is trained on the latent representation from the first stage autoencoder to obtain a full generative model.

\vspace{-1mm}
\subsection{Autoencoder for Compression and Novel-View Synthesis}
\vspace{-1mm}
Following the LDM framework~\citep{rombach2021highresolution}, we first train an autoencoder that encodes training data into latent representations. 
Unlike LDMs, however, where the task of the autoencoder is simply to compress and reconstruct the inputs, our setting is more complex, as the autoencoding model must also learn a 3D representation of the data such that it can infer reasonable novel views from a single input image.
The capability for novel-view synthesis will be used later by the diffusion model to perform 3D-aware image synthesis with 3D-consistent viewpoint changes.
However, as no multiview or explicit 3D geometry supervision is available, this novel-view synthesis task is highly under-constrained and non-trivial to solve. To aid this process, we provide additional cues about the geometry in the form of monocular depth supervision from a pre-trained network~\citep{Bhat2023ARXIV}. 

Specifically, we concatenate a given image $\mathbf{I}\in\mathbb{R}^{3\times H\times W}$ with its estimated monocular depth $\mathbf{D}\in\mathbb{R}^{1\times H\times W}$ channel-wise and encode them into a compressed latent representation $\mathbf{Z}\in\mathbb{R}^{c\times h\times w}$ via an encoder. 
As the encoder must infer a latent representation that encodes the underlying 3D object or scene of the input image, we found it beneficial to provide both $\mathbf{I}$ and $\mathbf{D}$ as input (see~\tabref{tab:ablations}). 
We choose a Feature Pyramid Network (FPN)~\citep{Lin2017CVPRb} architecture due to its large receptive field. For LDMs, latent space compression is crucial to train a diffusion model efficiently in the second stage. At input resolution $256\times256$ pixels, we use $c{=}4$, $h{=}w{=}32$ as in~\cite{rombach2021highresolution}.\looseness=-1

The decoder predicts a feature field from the compressed latent code $\mathbf{Z}$, which can be rendered from arbitrary viewing directions. The feature field is represented with triplanes. In contrast to previous works~\citep{Chan2022CVPR,Skorokhodov2022ARXIV}, our triplane representation is constructed from the latent feature map $\mathbf{Z}$ instead of being generated from random noise such that it is possible to reconstruct the input image. 
Taking inspiration from~\citep{Lin2023WACV, chan2023genvs}, we process $\mathbf{Z}$ with a combination of transformer blocks to facilitate learning global features and a CNN to increase the resolution of the features. 
For the transformer blocks after the CNN we use efficient self-attention~\citep{XieNEURIPS2021} to keep the computational cost manageable for the larger resolution feature maps. We find that this combination of transformer blocks and convolutional layers achieves better novel view synthesis than using a fully convolutional architecture (see~\tabref{tab:ablations}).

Next, the feature field is projected to the input view $\mathbf{P_0}$ and a novel view $\mathbf{P}_{nv}$ via volume rendering~\citep{Kajiya1984SIGGRAPH,Mildenhall2020ECCV}, as described in Sec.~\ref{sec:preliminarieseg3d}.
For the input view, we use the same fixed pose $\mathbf{P_0}$ for all instances. 
This means that we are modeling instances in \textit{view space} where the coordinate system is defined from the input camera's point of view.
Therefore, novel views can be sampled uniformly from a predefined range of angles around $\mathbf{P_0}$. 
In this work, we assume fixed camera intrinsics that we choose according to our camera settings. We find that using the same intrinsics for all datasets works well in practice (for details, see Appendix).
To model unbounded scenes, we sample points along rays linearly in disparity (inverse depth) instead of depth. This effectively samples more points close to the camera and uses fewer samples at large depths. 
Recall that these points are projected onto triplanes for inferring the triplane features. To ensure that the full depth range is mapped onto the triplanes, we use a contraction function as in~\citep{barron2022mipnerf360}. The contraction function maps all coordinates to a bounded range, which ensures that sampling points are mapped to valid coordinates on the triplanes (Supp. Mat. for details). We find that representing unbounded scenes with a combination of disparity sampling and a contraction function improves novel view synthesis, see~\tabref{tab:ablations}.
We render low-resolution images $\mathbf{\hat{I}}^{low}$, $\mathbf{\hat{I}}^{low}_{nv}$, depth maps $\mathbf{\hat{D}}^{low}$, $\mathbf{\hat{D}}^{low}_{nv}$ and feature maps $\mathbf{F}^{low}$,  $\mathbf{F}^{low}_{nv}$ from the feature field using volume rendering at $64\times64$ resolution, see Eq.~\eqref{eq:volrend}. 
The rendered low-resolution feature maps are then processed with a superresolution CNN module (SR CNN) that increases the spatial dimensions by $4\times$ to yield the reconstructed image $\mathbf{\hat{I}}$ and a novel view image $\mathbf{\hat{I}}_{nv}$ (see \figref{fig:system}).\looseness=-1

\textbf{Training Objective.} 
\label{sec:objective}
We train the autoencoder with a reconstruction loss on the input view and use an adversarial objective to supervise novel views~\citep{Mi2022ARXIV, Cai2022CVPR}. 
Similar to~\cite{rombach2021highresolution}, we add a small Kullback-Leibler (KL) divergence regularization term $\mathcal{L}_{KL}$ on the latent space $\mathbf{Z}$.
The reconstruction loss $\mathcal{L}_{rec}$ consists of a pixel-wise loss $\mathcal{L}_{px}= |\mathbf{\hat{I}}-\mathbf{I}|$, a perceptual loss $\mathcal{L}_{VGG}$~\citep{Zhang2018CVPRb}, and depth losses $\mathcal{L}_{depth}$. 

As our monocular depth estimation $\mathbf{D}$ is defined only up to scale, we first compute a scale $s$ and shift $t$ for each image by solving a least-squares criterion for $s$ and $t$, which has a closed-form solution~\citep{Eigen2014NIPS}. %
Following~\cite{Ranftl2020PAMI}, we enforce consistency between rendered (2D) depth $\mathbf{\hat{D}}^{low}$ and the downsampled monocular depth $\mathbf{D}^{low}$ that was estimated on the input images:
$\mathcal{L}_{depth}^{2D} = || (s\mathbf{\hat{D}}^{low}+t) - \mathbf{D}^{low} ||^2$.
We further found it beneficial to directly supervise the (normalized) rendering weights $w_r^i$ of the 3D sampling points (see Eq.~\eqref{eq:volrend}) with the depth.  
Let $\mathcal{K}_r(s\mathbf{D}^{low}+t)$ denote the index set of the $k$ sampling points closest to the rescaled monocular depth along ray $r$. Then,\looseness=-1
\begin{equation}
    \mathcal{L}_{depth}^{3D} = \sum_r \biggl[(1-\sum_{i\in\mathcal{K}_r} w_r^i)^2 + (\sum_{i\notin\mathcal{K}_r} w_r^i)^2\biggr].
\end{equation}
Intuitively, the loss encourages large rendering weights for the points close to the re-scaled monocular depth and small rendering weights for points farther away.
Note that we regularize the sum of the weights in the neighborhood $\mathcal{K}_r$ instead of the individual weights to account for imperfections in the monocular depth. In our experiments, we use a neighborhood size of $k=5$.

In addition to the reconstruction losses on the input view, we supervise the novel views of the input image. As per-pixel supervision is not available for novel views in our setting, we follow~\citep{Mi2022ARXIV, Cai2022CVPR} and use adversarial training to supervise novel views. We use a dual discriminator~\citep{Chan2022CVPR}, \ie, we upsample $\mathbf{\hat{I}}^{low}_{nv}$ and concatenate it with $\mathbf{\hat{I}}_{nv}$ as input to the discriminator as a fake pair (see \figref{fig:system}). 
Similarly, $\mathbf{I}$ is first downsampled to simulate a lower resolution image, and then is upsampled back to the original resolution and concatenated with the original $\mathbf{I}$ to be used as the real pair to the discriminator.
Let $\mathcal{E}_\theta$, $\mathcal{G}_\psi$ and $D_\phi$ denote encoder, decoder and discriminator with parameters $\theta$, $\psi$ and $\phi$, respectively. 
For brevity, we omit the upsampling and concatenation of the discriminator inputs in the adversarial objective\looseness=-1
\begin{equation}
V(\bI, \bP_{nv}, \lambda; \theta, \psi, \phi) = 
f\left(-D_\phi\left(\mathcal{G}_\psi(\mathcal{E}_\theta(\mathbf{I}, \mathbf{D}),\bP_{nv})\right)\right) \,+\,
f(D_\phi(\bI))
\,-\, \lambda {\Vert \nabla D_\phi(\bI)\Vert}^2,
\end{equation}
where $f(x)=-\log(1+\exp(-x))$ and $\lambda$ controls the 
R1-regularizer~\citep{Mescheder2018ICML}.

We find that an additional discriminator $D^{depth}_{\chi}$ on the low-resolution depth maps further improves novel view synthesis.
$D^{depth}_{\chi}$ helps to ensure the volume-rendered $\mathbf{\hat{D}}^{low}$ is realistic (see~\tabref{tab:ablations}). 
The autoencoder and discriminators are trained with alternating gradient descent steps combining the adversarial objectives with the reconstruction and regularization terms.
Implementation details in Appendix.\looseness=-1

In conclusion, to learn a latent representation suitable not only for reconstruction but also novel view synthesis, we use both reconstruction and adversarial objectives. Note, however, that this is still fundamentally different from regular GAN-like training, which most existing works on 3D-aware image synthesis rely on. Our reconstruction losses on input views prevent mode collapse and ensure stable training. This makes our approach arguably more robust and scalable. Moreover, inputs to the decoder are not sampled from random noise, like in GANs, but correspond to image encodings.

\vspace{-1mm}
\subsection{Latent Diffusion Model}
\vspace{-1mm}
\label{sec:ldm_method}
The autoencoder trained as described above learns a latent representation that is \textbf{(i)} compressed, and therefore suitable for efficient training of a latent diffusion model, and simultaneously also \textbf{(ii)} 3D-aware in the sense that it enables the prediction of a triplane representation from which consistent novel views can be synthesized corresponding to different viewing directions onto the modeled scene.
Consequently, once the autoencoder is trained, we fit a latent diffusion model on its 3D-aware latent space in the second training stage (see \figref{fig:system}). To this end, we encode our training images into the latent space and train the diffusion model on the encoded data. 
Importantly, although the autoencoder produces a compact 3D-aware latent representation, it is structured as a spatial 2D latent feature grid, as in standard LDMs for 2D image synthesis.
Therefore, we can directly follow the training procedure of regular LDMs~\citep{rombach2021highresolution} when training the diffusion model.
Eventually, this allows us to train a powerful 3D-aware generative model that can be trained and sampled efficiently in 2D latent space.
Our training objective is the standard denoising score matching objective as given in Eq~\eqref{eq:diffusionobjective}, applied in latent space.\looseness=-1

We adopt the architecture of regular 2D image LDMs~\citep{rombach2021highresolution} to train our model.
The denoiser $\mathcal{F}_\omega$ is implemented as a 2D U-Net with residual blocks~\citep{He2016CVPR} and self-attention layers~\citep{Vaswani2017NIPS}. 
As discussed, the autoencoder's latent distribution is regularized with a KL divergence loss~\citep{Kingma2014ICLR,Rezende2014ICML,rombach2021highresolution} (see Appendix) to be roughly aligned with the standard normal distribution. However, as we enforce a very low-weighted KL loss, the distribution can have a larger variance. 
We estimate the standard deviation of the latent space using a batch of encoded training data and use the resulting value to normalize the latent distribution to yield a standard deviation close to $1$ before fitting the diffusion model.
We use the DDIM sampler~\citep{song2020denoising2} with 200 steps. More implementation details in the Appendix.\looseness=-1

\section{Experiments}
\vspace{-1mm}
The performance of LDMs is upper-bounded by the quality of the latent space they are trained on, \ie, we cannot expect to generate better novel images than what the autoencoder achieves in terms of reconstructions. Hence, a powerful autoencoder is key to training a good generative model in the second stage.
We first analyze the reconstruction quality of WildFusion's autoencoder as well as its ability to synthesize novel views (Sec.~\ref{sec:exp_autoencoder}). 
Next, we evaluate the full WildFusion model against the state-of-the-art approaches for 3D-aware image synthesis (Sec.~\ref{sec:expdiffusion}). We provide ablation studies in Sec.~\ref{sec:exp_ablation}. Our videos included on the project page (\url{https://katjaschwarz.github.io/wildfusion/}) show generated 3D-aware samples with camera motions. Further results are also shown in App.~\ref{sec:supp_results}.\looseness=-1

\textbf{Datasets.} While previous 3D-aware generative models mainly focus on aligned datasets like portrait images, we study a general setting in which a canonical camera system cannot be clearly defined. Hence, we use non-aligned datasets with complex geometry: SDIP Dogs, Elephants, Horses~\citep{mokady2022selfdistilled,Yu2015CVPR} as well as class-conditional ImageNet~\citep{Deng2009CVPR}.\looseness=-1
\\
\textbf{Baselines.} 
We compare against the state-of-the-art generative models for 3D-aware image synthesis, EG3D~\citep{Chan2022CVPR}, 3DGP~\citep{skorokhodov20233dgp} and POF3D~\citep{Shi2023CVPR} as well as StyleNeRF~\citep{Gu2021ARXIV}.
3DGP and POF3D learn a
camera distribution in canonical space and can be trained on unposed images. Since we also aim to compare to other models working in the same setting as WildFusion, \ie, in view space, we adapt EG3D so that it can be trained in view space and without camera poses (indicated as EG3D* below); see Appendix for details. We also train another variant of EG3D* where we add a depth discriminator to incorporate monocular depth information. Note that the regular version of EG3D that relies on object poses is clearly outperformed by 3DGP (as shown in their paper~\citep{skorokhodov20233dgp}); hence, we do not explicitly compare to it.\looseness=-1
\\
\textbf{Evaluation Metrics.}
For the autoencoder, we measure reconstruction via learned perceptual image patch similarity (LPIPS)~\citep{Zhan2018CVPR} and quantify novel view quality with Fr\'echet Inception Distance (nvFID)~\citep{Heusel2017NIPS} on $1000$ held-out dataset images. Following prior art, we sample camera poses around the input view $\mathbf{P}_0$ from Gaussian distributions with $\sigma=$ 0.3 and 0.15 radians for the yaw and pitch angles~\citep{Chan2020CVPR}.
We also report non-flatness-score (NFS)~\citep{skorokhodov20233dgp}. It measures average entropy of the normalized depth maps' histograms and quantifies surface flatness, indicating geometry quality.
For the full generative models, we measure NFS and evaluate FID between $10$k generated images and the full dataset, sampling camera poses as for nvFID. As FID can be prone to distortions~\citep{clipFID}, we also show $\text{FID}_\text{CLIP}$, which uses CLIP features.
To quantify diversity, we report Recall, and we also show Precision \citep{sajjadi2018assessing,kynkaanniemi2019improved}. 
Qualitative results display a view range of 30$^\circ$ and 15$^\circ$ for azimuth and polar angles
around the input view, similar to \cite{skorokhodov20233dgp}.
\vspace{-1mm}
\subsection{Autoencoder for Reconstruction and Novel-view Synthesis}
\label{sec:exp_autoencoder}
\vspace{-1mm}
\aeresults
\reconstruction
As EG3D~\citep{Chan2022CVPR} is a GAN-based approach, we need to perform GAN-inversion~\citep{Chan2022CVPR,RoichTOG2023} to reconstruct input images (we use the scripts provided by the authors).
Quantitative results are in~\tabref{tab:reconstruction}, qualitative comparisons
in~\figref{fig:aeresults} (more in App.~\ref{sec:supp_results}). 
Compared to EG3D using GAN-inversion, 
WildFusion's autoencoder achieves superior performance on all metrics and is also more efficient, since we do not require a lengthy optimization process (which occasionally diverges) to embed input images into latent space. Despite its low latent space dimension of $32\times32\times4$, our autoencoder achieves both good compression and novel view synthesis. In \figref{fig:teaser}, \figref{fig:ldmresults} and in App.~\ref{sec:supp_results}, we also show novel view synthesis results on ImageNet, which performs equally well.\looseness=-1
\vspace{-1mm}
\subsection{3D-Aware Image Synthesis with Latent Diffusion Models}
\label{sec:expdiffusion}
\vspace{-1mm}
\baselinecomp
\textbf{SDIP Datasets.} We compare \ourmodel against state-of-the-art 3D-aware generative models (\tabref{tab:baselinecomp}) and provide model samples in the App.~\ref{sec:supp_results}. Our videos in the Supp. Mat. show more generated 3D-aware samples, including smooth viewpoint changes.
Compared to EG3D*~\citep{Chan2022CVPR},
which effectively also models instances in view-space, \ourmodel achieves higher performance on all metrics, \ie, in terms of image quality (FID/$\text{FID}_{\text{CLIP}}$), 3D geometry (NFS), and diversity (Recall). This validates the effectiveness of our LDM framework in this setting. When also adding a depth discriminator to EG3D*, \ourmodel still achieves better FID and NFS and Recall. Note that in particular on NFS and Recall, the performance gap is generally large. We also outperform StyleNeRF. We conclude that previous works that operate in view space
can struggle with flat geometry and distribution coverage when training on reasonably complex, non-aligned images. In contrast, WildFusion achieves state-of-the-art performance in this setting.\looseness=-1

The baselines 3DGP~\citep{skorokhodov20233dgp} and POF3D~\citep{Shi2023CVPR} both rely on sophisticated camera pose estimation procedures and do not operate in view space, which makes a direct comparison with \ourmodel difficult. Nevertheless, WildFusion performs on-par or even better on the different metrics, despite not relying on posed images or learned pose or camera distributions at all. These previous works' reliance on complex canonical camera systems represents a major limitation with regards to their scalability, which our 3D-aware LDM framework avoids. Note that in particular on Recall, WildFusion always performs superior compared to all other methods, demonstrating that diffusion-based frameworks are usually better than GAN-based ones in terms of sample diversity.

\ldmresults
\baselinecompimagenet
\textbf{ImageNet.} We find that WildFusion outperforms all baselines on NFS, Precision and Recall by large margins, for varying classifier-free guidance scales (\tabref{tab:baselinecomp_imagenet}). The extremely low Recall scores of the GAN-based baselines indicate very low sample diversity (mode collapse). We visually validate this for 3DGP, the strongest of the three baselines, in \figref{fig:imagenet_comp}: 3DGP collapses and produces almost identical samples within classes, showing virtually no diversity. In contrast, our model produces diverse, high-quality samples. Note that this failure of 3DGP is pointed out by the authors (see \textit{``Limitations and failure cases''} at \url{https://snap-research.github.io/3dgp/}). The FID metric does not accurately capture that mode collapse. While we outperform EG3D and StyleNeRF also on FID, \ourmodel is slightly behind 3DGP. However, it is known that FID is a questionable metric~\citep{clipFID} and the qualitative results in \figref{fig:ldmresults} and \figref{fig:imagenet_comp} show that \ourmodel generates diverse and high-fidelity images and generally learns a reasonable geometry in this challenging setting. We believe that a mode-collapsed generative model like 3DGP, despite good FID scores, is not useful in practice. Finally, note that due to limited compute resources our ImageNet model was trained only with a total batch size of 256. However, non-3D-aware ImageNet diffusion models are typically trained on batch sizes $>$1000 to achieve strong performance~\citep{rombach2021highresolution,karras2022edm,kingma2023understanding}. Hence, scaling our model and the compute resources would likely boost the results significantly.

\textbf{Interpolation and Generative Resampling.} In \figref{fig:interpolation}, we use \ourmodel to perform semantically meaningful 3D-aware image interpolation between two given (or generated) images. Moreover, in \figref{fig:denoise} we demonstrate how we can use our 3D-aware latent diffusion model to refine images and geometry by only partially diffusing their encodings and regenerating from those intermediate diffusion levels. These two applications highlight the versatility of \ourmodel and have potential use cases in 3D-aware image editing. To the best of our knowledge, this is the first time such applications have been demonstrated for such 3D-aware image generative models. See the project page (\url{https://katjaschwarz.github.io/wildfusion/}) for animations with viewpoint changes for these 3D-aware image interpolation and generative image resampling results.
\interpolation
\denoise

\vspace{-1mm}
\subsection{Ablation Studies}
\label{sec:exp_ablation}
\vspace{-1mm}
We provide a detailed ablation study in~\tabref{tab:ablations}, 
starting from a base model,
where the decoder architecture follows EG3D's generator.
We gradually introduce changes and track their impact on novel view synthesis ($\text{nvFID}$) and geometry (NFS) (corresponding samples in App.~\ref{sec:supp_results} in \figref{fig:ablation}). 
For computational efficiency, we perform the study at image resolution $128^2$ and reduce the number of network parameters compared to our main models. 
Like LDMs~\citep{rombach2021highresolution}, the initial setting is trained only with reconstruction losses. Unsurprisingly, this results in planar geometry, indicated by low NFS.
This changes when a discriminator on novel views is added. However, geometry becomes noisy and incomplete, indicated by high NFS and worse nvFID.
We hypothesize the purely convolutional architecture is suboptimal. Hence, in the next step, we instead use a combination of convolutions and transformer blocks (ViT~\citep{Dosovitskiy2020ARXIV}) in the decoder, which improves novel view quality and results in less noisy geometry. \ablations
Adding the monocular depth discriminator $D^{depth}_{\chi}$ significantly improves nvFID, and we observe an even bigger improvement when tailoring the model to represent unbounded scenes with disparity sampling and a coordinate contraction function.
Further supervision on the rendered depth ($\mathcal{L}_{depth}^{2D}$) does not improve results but as it does not hurt performance either, we kept it in our pipeline. 
Lastly, we find that both adding depth as an input to the encoder and directly supervising the rendering weights with $\mathcal{L}_{depth}^{3D}$ result in slight improvements in NFS and qualitatively improve geometry.

\section{Conclusions}
\vspace{-1mm}
We introduce \ourmodel, a 3D-aware LDM for 3D-aware image synthesis. \ourmodel is trained without multiview or 3D geometry supervision and relies neither on posed images nor on learned pose or camera distributions.
Key to our framework is an image autoencoder with a 3D-aware latent space that simultaneously enables not only novel view synthesis but also compression. This allows us to efficiently train a diffusion model in the autoencoder's latent space. \ourmodel outperforms recent state-of-the-art GAN-based methods when training on diverse data without camera poses.
Future work could scale up 3D-aware LDMs to the text-conditional setting, similar to how 2D diffusion models have been applied on extremely diverse datasets~\citep{Ramesh2022ARXIV,rombach2021highresolution,saharia2022photorealistic,balaji2022eDiffi}.

\clearpage
\begin{ack}
Katja Schwarz and Andreas Geiger were supported by the ERC Starting Grant LEGO-3D (850533) and the DFG EXC number 2064/1 - project number 390727645. The authors thank the International Max Planck Research School for Intelligent Systems (IMPRS-IS) for supporting Katja Schwarz. Lastly, we would like to thank Nicolas Guenther for his general support.
\end{ack}
 
\bibliographystyle{iclr2024_conference}
\bibliography{bibliography,further_references}

\begin{thebibliography}{120}
\providecommand{\natexlab}[1]{#1}
\providecommand{\url}[1]{\texttt{#1}}
\expandafter\ifx\csname urlstyle\endcsname\relax
  \providecommand{\doi}[1]{doi: #1}\else
  \providecommand{\doi}{doi: \begingroup \urlstyle{rm}\Url}\fi

\bibitem[Abdal et~al.(2019)Abdal, Qin, and Wonka]{abdal2019image2stylegan}
Rameen Abdal, Yipeng Qin, and Peter Wonka.
\newblock Image2stylegan: How to embed images into the stylegan latent space?
\newblock In \emph{Proceedings of the IEEE/CVF International Conference on Computer Vision}, pp.\  4432--4441, 2019.

\bibitem[Anciukevicius et~al.(2023)Anciukevicius, Xu, Fisher, Henderson, Bilen, Mitra, and Guerrero]{anciukevicius2022renderdiffusion}
Titas Anciukevicius, Zexiang Xu, Matthew Fisher, Paul Henderson, Hakan Bilen, Niloy~J. Mitra, and Paul Guerrero.
\newblock {RenderDiffusion}: Image diffusion for {3D} reconstruction, inpainting and generation.
\newblock In \emph{Proceedings of the IEEE/CVF Conference on Computer Vision and Pattern Recognition (CVPR)}, 2023.

\bibitem[Balaji et~al.(2022)Balaji, Nah, Huang, Vahdat, Song, Kreis, Aittala, Aila, Laine, Catanzaro, Karras, and Liu]{balaji2022eDiffi}
Yogesh Balaji, Seungjun Nah, Xun Huang, Arash Vahdat, Jiaming Song, Karsten Kreis, Miika Aittala, Timo Aila, Samuli Laine, Bryan Catanzaro, Tero Karras, and Ming-Yu Liu.
\newblock ediff-i: Text-to-image diffusion models with ensemble of expert denoisers.
\newblock \emph{arXiv preprint arXiv:2211.01324}, 2022.

\bibitem[Barron et~al.(2022)Barron, Mildenhall, Verbin, Srinivasan, and Hedman]{barron2022mipnerf360}
Jonathan~T. Barron, Ben Mildenhall, Dor Verbin, Pratul~P. Srinivasan, and Peter Hedman.
\newblock Mip-nerf 360: Unbounded anti-aliased neural radiance fields.
\newblock \emph{CVPR}, 2022.

\bibitem[Bautista et~al.(2022)Bautista, Guo, Abnar, Talbott, Toshev, Chen, Dinh, Zhai, Goh, Ulbricht, Dehghan, and Susskind]{Bautista2022ARXIV}
Miguel~{\'{A}}ngel Bautista, Pengsheng Guo, Samira Abnar, Walter Talbott, Alexander Toshev, Zhuoyuan Chen, Laurent Dinh, Shuangfei Zhai, Hanlin Goh, Daniel Ulbricht, Afshin Dehghan, and Josh~M. Susskind.
\newblock {GAUDI:} {A} neural architect for immersive 3d scene generation.
\newblock \emph{arXiv}, 2022.

\bibitem[Bhat et~al.(2023)Bhat, Birkl, Wofk, Wonka, and M{\"{u}}ller]{Bhat2023ARXIV}
Shariq~Farooq Bhat, Reiner Birkl, Diana Wofk, Peter Wonka, and Matthias M{\"{u}}ller.
\newblock Zoedepth: Zero-shot transfer by combining relative and metric depth.
\newblock \emph{arXiv}, 2023.

\bibitem[Cai et~al.(2022)Cai, Obukhov, Dai, and Van~Gool]{Cai2022CVPR}
Shengqu Cai, Anton Obukhov, Dengxin Dai, and Luc Van~Gool.
\newblock Pix2nerf: Unsupervised conditional p-gan for single image to neural radiance fields translation.
\newblock In \emph{CVPR}, 2022.

\bibitem[Chan et~al.(2021)Chan, Monteiro, Kellnhofer, Wu, and Wetzstein]{Chan2020CVPR}
Eric~R. Chan, Marco Monteiro, Petr Kellnhofer, Jiajun Wu, and Gordon Wetzstein.
\newblock Pi-gan: Periodic implicit generative adversarial networks for 3d-aware image synthesis.
\newblock In \emph{CVPR}, 2021.

\bibitem[Chan et~al.(2022)Chan, Lin, Chan, Nagano, Pan, Mello, Gallo, Guibas, Tremblay, Khamis, Karras, and Wetzstein]{Chan2022CVPR}
Eric~R. Chan, Connor~Z. Lin, Matthew~A. Chan, Koki Nagano, Boxiao Pan, Shalini~De Mello, Orazio Gallo, Leonidas Guibas, Jonathan Tremblay, Sameh Khamis, Tero Karras, and Gordon Wetzstein.
\newblock Efficient geometry-aware {3D} generative adversarial networks.
\newblock In \emph{CVPR}, 2022.

\bibitem[Chan et~al.(2023)Chan, Nagano, Chan, Bergman, Park, Levy, Aittala, Mello, Karras, and Wetzstein]{chan2023genvs}
Eric~R. Chan, Koki Nagano, Matthew~A. Chan, Alexander~W. Bergman, Jeong~Joon Park, Axel Levy, Miika Aittala, Shalini~De Mello, Tero Karras, and Gordon Wetzstein.
\newblock {GeNVS}: Generative novel view synthesis with {3D}-aware diffusion models.
\newblock In \emph{arXiv}, 2023.

\bibitem[Choi et~al.(2020)Choi, Uh, Yoo, and Ha]{Choi2020CVPR}
Yunjey Choi, Youngjung Uh, Jaejun Yoo, and Jung-Woo Ha.
\newblock Stargan v2: Diverse image synthesis for multiple domains.
\newblock In \emph{CVPR}, 2020.

\bibitem[Deng et~al.(2022{\natexlab{a}})Deng, Jiang, Qi, Yan, Zhou, Guibas, and Anguelov]{deng2022nerdi}
Congyue Deng, Chiyu~"Max'' Jiang, Charles~R. Qi, Xinchen Yan, Yin Zhou, Leonidas Guibas, and Dragomir Anguelov.
\newblock Nerdi: Single-view nerf synthesis with language-guided diffusion as general image priors.
\newblock \emph{arXiv preprint arXiv:2212.03267}, 2022{\natexlab{a}}.

\bibitem[Deng et~al.(2009)Deng, Dong, Socher, jia Li, Li, and Fei-fei]{Deng2009CVPR}
Jia Deng, Wei Dong, Richard Socher, Li~jia Li, Kai Li, and Li~Fei-fei.
\newblock Imagenet: A large-scale hierarchical image database.
\newblock In \emph{CVPR}, 2009.

\bibitem[Deng et~al.(2022{\natexlab{b}})Deng, Yang, Xiang, and Tong]{DENG2021ARXIV}
Yu~Deng, Jiaolong Yang, Jianfeng Xiang, and Xin Tong.
\newblock Gram: Generative radiance manifolds for 3d-aware image generation.
\newblock In \emph{CVPR}, 2022{\natexlab{b}}.

\bibitem[Dhariwal \& Nichol(2021{\natexlab{a}})Dhariwal and Nichol]{dhariwal2021diffusion}
Prafulla Dhariwal and Alexander Nichol.
\newblock Diffusion models beat gans on image synthesis.
\newblock \emph{Advances in Neural Information Processing Systems}, 34:\penalty0 8780--8794, 2021{\natexlab{a}}.

\bibitem[Dhariwal \& Nichol(2021{\natexlab{b}})Dhariwal and Nichol]{Dhariwal2021NIPS}
Prafulla Dhariwal and Alexander~Quinn Nichol.
\newblock Diffusion models beat gans on image synthesis.
\newblock In Marc'Aurelio Ranzato, Alina Beygelzimer, Yann~N. Dauphin, Percy Liang, and Jennifer~Wortman Vaughan (eds.), \emph{NeurIPS}, 2021{\natexlab{b}}.

\bibitem[Dockhorn et~al.(2022{\natexlab{a}})Dockhorn, Vahdat, and Kreis]{Dockhorn2022ICLR}
Tim Dockhorn, Arash Vahdat, and Karsten Kreis.
\newblock Score-based generative modeling with critically-damped langevin diffusion.
\newblock In \emph{ICLR}, 2022{\natexlab{a}}.

\bibitem[Dockhorn et~al.(2022{\natexlab{b}})Dockhorn, Vahdat, and Kreis]{dockhorn2022genie}
Tim Dockhorn, Arash Vahdat, and Karsten Kreis.
\newblock {GENIE: Higher-Order Denoising Diffusion Solvers}.
\newblock In \emph{Advances in Neural Information Processing Systems}, 2022{\natexlab{b}}.

\bibitem[Dosovitskiy et~al.(2020)Dosovitskiy, Beyer, Kolesnikov, Weissenborn, Zhai, Unterthiner, Dehghani, Minderer, Heigold, Gelly, Uszkoreit, and Houlsby]{Dosovitskiy2020ARXIV}
Alexey Dosovitskiy, Lucas Beyer, Alexander Kolesnikov, Dirk Weissenborn, Xiaohua Zhai, Thomas Unterthiner, Mostafa Dehghani, Matthias Minderer, Georg Heigold, Sylvain Gelly, Jakob Uszkoreit, and Neil Houlsby.
\newblock An image is worth 16x16 words: Transformers for image recognition at scale.
\newblock \emph{arXiv}, 2010.11929, 2020.

\bibitem[Du et~al.(2021)Du, Collins, Tenenbaum, and Sitzmann]{du2021gem}
Yilun Du, M.~Katherine Collins, B.~Joshua Tenenbaum, and Vincent Sitzmann.
\newblock Learning signal-agnostic manifolds of neural fields.
\newblock In \emph{Advances in Neural Information Processing Systems}, 2021.

\bibitem[Dupont et~al.(2022)Dupont, Kim, Eslami, Rezende, and Rosenbaum]{dupont2022functa}
Emilien Dupont, Hyunjik Kim, S.~M.~Ali Eslami, Danilo~Jimenez Rezende, and Dan Rosenbaum.
\newblock From data to functa: Your data point is a function and you can treat it like one.
\newblock In \emph{39th International Conference on Machine Learning (ICML)}, 2022.

\bibitem[Eftekhar et~al.(2021)Eftekhar, Sax, Malik, and Zamir]{Eftekhar2021ICCV}
Ainaz Eftekhar, Alexander Sax, Jitendra Malik, and Amir Zamir.
\newblock Omnidata: A scalable pipeline for making multi-task mid-level vision datasets from 3d scans.
\newblock In \emph{ICCV}, 2021.

\bibitem[Eigen et~al.(2014)Eigen, Puhrsch, and Fergus]{Eigen2014NIPS}
David Eigen, Christian Puhrsch, and Rob Fergus.
\newblock Depth map prediction from a single image using a multi-scale deep network.
\newblock In \emph{NeurIPS}, 2014.

\bibitem[Esser et~al.(2021)Esser, Rombach, and Ommer]{Esser2021CVPR}
Patrick Esser, Robin Rombach, and Bj{\"{o}}rn Ommer.
\newblock Taming transformers for high-resolution image synthesis.
\newblock In \emph{CVPR}, 2021.

\bibitem[Fridman et~al.(2023)Fridman, Abecasis, Kasten, and Dekel]{SceneScape}
Rafail Fridman, Amit Abecasis, Yoni Kasten, and Tali Dekel.
\newblock Scenescape: Text-driven consistent scene generation.
\newblock \emph{arXiv preprint arXiv:2302.01133}, 2023.

\bibitem[Goodfellow et~al.(2014)Goodfellow, Pouget{-}Abadie, Mirza, Xu, Warde{-}Farley, Ozair, Courville, and Bengio]{Goodfellow2014NIPS}
Ian~J. Goodfellow, Jean Pouget{-}Abadie, Mehdi Mirza, Bing Xu, David Warde{-}Farley, Sherjil Ozair, Aaron~C. Courville, and Yoshua Bengio.
\newblock Generative adversarial nets.
\newblock In \emph{NeurIPS}, 2014.

\bibitem[Gu et~al.(2022)Gu, Liu, Wang, and Theobalt]{Gu2021ARXIV}
Jiatao Gu, Lingjie Liu, Peng Wang, and Christian Theobalt.
\newblock Stylenerf: {A} style-based 3d-aware generator for high-resolution image synthesis.
\newblock \emph{ICLR}, 2022.

\bibitem[He et~al.(2016)He, Zhang, Ren, and Sun]{He2016CVPR}
Kaiming He, Xiangyu Zhang, Shaoqing Ren, and Jian Sun.
\newblock Deep residual learning for image recognition.
\newblock In \emph{CVPR}, 2016.

\bibitem[Henzler et~al.(2019)Henzler, Mitra, , and Ritschel]{Henzler2019ICCV}
Philipp Henzler, Niloy~J Mitra, , and Tobias Ritschel.
\newblock Escaping plato's cave: 3d shape from adversarial rendering.
\newblock In \emph{ICCV}, 2019.

\bibitem[Heusel et~al.(2017)Heusel, Ramsauer, Unterthiner, Nessler, and Hochreiter]{Heusel2017NIPS}
Martin Heusel, Hubert Ramsauer, Thomas Unterthiner, Bernhard Nessler, and Sepp Hochreiter.
\newblock Gans trained by a two time-scale update rule converge to a local nash equilibrium.
\newblock In \emph{NeurIPS}, 2017.

\bibitem[Ho \& Salimans(2021)Ho and Salimans]{ho2021classifierfree}
Jonathan Ho and Tim Salimans.
\newblock Classifier-free diffusion guidance.
\newblock In \emph{NeurIPS 2021 Workshop on Deep Generative Models and Downstream Applications}, 2021.

\bibitem[Ho et~al.(2020)Ho, Jain, and Abbeel]{ho2020ddpm2}
Jonathan Ho, Ajay Jain, and Pieter Abbeel.
\newblock Denoising diffusion probabilistic models.
\newblock In \emph{Advances in Neural Information Processing Systems}, 2020.

\bibitem[Ho et~al.(2022)Ho, Saharia, Chan, Fleet, Norouzi, and Salimans]{Ho2022MLRes}
Jonathan Ho, Chitwan Saharia, William Chan, David~J. Fleet, Mohammad Norouzi, and Tim Salimans.
\newblock Cascaded diffusion models for high fidelity image generation.
\newblock \emph{J. Mach. Learn. Res.}, 2022.

\bibitem[Hyv\"{a}rinen(2005)]{hyvarinen2005scorematching}
Aapo Hyv\"{a}rinen.
\newblock Estimation of non-normalized statistical models by score matching.
\newblock \emph{Journal of Machine Learning Research}, 6:\penalty0 695–709, 2005.
\newblock ISSN 1532-4435.

\bibitem[Jo et~al.(2021)Jo, Shim, Jung, Yang, and Choo]{Jo2021ARXIV}
Kyungmin Jo, Gyumin Shim, Sanghun Jung, Soyoung Yang, and Jaegul Choo.
\newblock Cg-nerf: Conditional generative neural radiance fields.
\newblock \emph{arXiv}, 2021.

\bibitem[Kajiya \& Herzen(1984)Kajiya and Herzen]{Kajiya1984SIGGRAPH}
James~T. Kajiya and Brian~Von Herzen.
\newblock Ray tracing volume densities.
\newblock In \emph{ACM Trans. on Graphics}, 1984.

\bibitem[Kalischek et~al.(2022)Kalischek, Peters, Wegner, and Schindler]{kalischek2022tetrahedral}
Nikolai Kalischek, Torben Peters, Jan~D Wegner, and Konrad Schindler.
\newblock Tetrahedral diffusion models for 3d shape generation.
\newblock \emph{arXiv preprint arXiv:2211.13220}, 2022.

\bibitem[Karras et~al.(2018)Karras, Aila, Laine, and Lehtinen]{Karras2018ICLR}
Tero Karras, Timo Aila, Samuli Laine, and Jaakko Lehtinen.
\newblock Progressive growing of {GAN}s for improved quality, stability, and variation.
\newblock In \emph{ICLR}, 2018.

\bibitem[Karras et~al.(2019)Karras, Laine, and Aila]{Karras2019CVPR}
Tero Karras, Samuli Laine, and Timo Aila.
\newblock A style-based generator architecture for generative adversarial networks.
\newblock In \emph{CVPR}, 2019.

\bibitem[Karras et~al.(2020)Karras, Laine, Aittala, Hellsten, Lehtinen, and Aila]{Karras2020CVPRa}
Tero Karras, Samuli Laine, Miika Aittala, Janne Hellsten, Jaakko Lehtinen, and Timo Aila.
\newblock Analyzing and improving the image quality of {StyleGAN}.
\newblock In \emph{CVPR}, 2020.

\bibitem[Karras et~al.(2021)Karras, Aittala, Laine, H\"ark\"onen, Hellsten, Lehtinen, and Aila]{Karras2021NEURIPS}
Tero Karras, Miika Aittala, Samuli Laine, Erik H\"ark\"onen, Janne Hellsten, Jaakko Lehtinen, and Timo Aila.
\newblock Alias-free generative adversarial networks.
\newblock In \emph{NeurIPS}, 2021.

\bibitem[Karras et~al.(2022)Karras, Aittala, Aila, and Laine]{karras2022edm}
Tero Karras, Miika Aittala, Timo Aila, and Samuli Laine.
\newblock Elucidating the design space of diffusion-based generative models.
\newblock In \emph{Advances in Neural Information Processing Systems}, 2022.

\bibitem[Kim et~al.(2023)Kim, Brown, Yin, Kreis, Schwarz, Li, Rombach, Torralba, and Fidler]{kim2023nfldm}
Seung~Wook Kim, Bradley Brown, Kangxue Yin, Karsten Kreis, Katja Schwarz, Daiqing Li, Robin Rombach, Antonio Torralba, and Sanja Fidler.
\newblock Neuralfield-ldm: Scene generation with hierarchical latent diffusion models.
\newblock In \emph{CVPR}, 2023.

\bibitem[Kingma \& Ba(2015)Kingma and Ba]{Kingma2015ICLR}
Diederik~P. Kingma and Jimmy Ba.
\newblock Adam: {A} method for stochastic optimization.
\newblock In \emph{ICLR}, 2015.

\bibitem[Kingma \& Gao(2023)Kingma and Gao]{kingma2023understanding}
Diederik~P. Kingma and Ruiqi Gao.
\newblock Understanding diffusion objectives as the elbo with simple data augmentation, 2023.

\bibitem[Kingma \& Welling(2014)Kingma and Welling]{Kingma2014ICLR}
Diederik~P. Kingma and Max Welling.
\newblock Auto-encoding variational bayes.
\newblock \emph{ICLR}, 2014.

\bibitem[Kynk{\"a}{\"a}nniemi et~al.(2019)Kynk{\"a}{\"a}nniemi, Karras, Laine, Lehtinen, and Aila]{kynkaanniemi2019improved}
Tuomas Kynk{\"a}{\"a}nniemi, Tero Karras, Samuli Laine, Jaakko Lehtinen, and Timo Aila.
\newblock Improved precision and recall metric for assessing generative models.
\newblock \emph{Advances in Neural Information Processing Systems}, 32, 2019.

\bibitem[Kynk{\"{a}}{\"{a}}nniemi et~al.(2023)Kynk{\"{a}}{\"{a}}nniemi, Karras, Aittala, Aila, and Lehtinen]{clipFID}
Tuomas Kynk{\"{a}}{\"{a}}nniemi, Tero Karras, Miika Aittala, Timo Aila, and Jaakko Lehtinen.
\newblock The role of imagenet classes in fr{\'{e}}chet inception distance.
\newblock In \emph{ICLR}, 2023.

\bibitem[Lan et~al.(2022)Lan, Meng, Yang, Loy, and Dai]{Lan2022ARXIV}
Yushi Lan, Xuyi Meng, Shuai Yang, Chen~Change Loy, and Bo~Dai.
\newblock Self-supervised geometry-aware encoder for style-based 3d {GAN} inversion.
\newblock \emph{arXiv}, 2022.

\bibitem[Li et~al.(2022)Li, Xu, Wu, Zheng, Dai, Pumarola, Zhang, Vajda, and Kitani]{Li2022ARXIV}
Yu{-}Jhe Li, Tao Xu, Bichen Wu, Ningyuan Zheng, Xiaoliang Dai, Albert Pumarola, Peizhao Zhang, Peter Vajda, and Kris Kitani.
\newblock 3d-aware encoding for style-based neural radiance fields.
\newblock \emph{arXiv}, 2022.

\bibitem[Liao et~al.(2020)Liao, Schwarz, Mescheder, and Geiger]{Liao2020CVPRa}
Yiyi Liao, Katja Schwarz, Lars~M. Mescheder, and Andreas Geiger.
\newblock Towards unsupervised learning of generative models for 3d controllable image synthesis.
\newblock \emph{CVPR}, 2020.

\bibitem[Lin et~al.(2022)Lin, Gao, Tang, Takikawa, Zeng, Huang, Kreis, Fidler, Liu, and Lin]{lin2022magic3d}
Chen-Hsuan Lin, Jun Gao, Luming Tang, Towaki Takikawa, Xiaohui Zeng, Xun Huang, Karsten Kreis, Sanja Fidler, Ming-Yu Liu, and Tsung-Yi Lin.
\newblock Magic3d: High-resolution text-to-3d content creation.
\newblock \emph{arXiv preprint arXiv:2211.10440}, 2022.

\bibitem[Lin et~al.(2023)Lin, Yen-Chen, Lai, Lin, Shih, and Ramamoorthi]{Lin2023WACV}
Kai-En Lin, Lin Yen-Chen, Wei-Sheng Lai, Tsung-Yi Lin, Yi-Chang Shih, and Ravi Ramamoorthi.
\newblock Vision transformer for nerf-based view synthesis from a single input image.
\newblock In \emph{IEEE/CVF Winter Conference on Applications of Computer Vision}, 2023.

\bibitem[Lin et~al.(2017)Lin, Doll{\'{a}}r, Girshick, He, Hariharan, and Belongie]{Lin2017CVPRb}
Tsung{-}Yi Lin, Piotr Doll{\'{a}}r, Ross~B. Girshick, Kaiming He, Bharath Hariharan, and Serge~J. Belongie.
\newblock Feature pyramid networks for object detection.
\newblock In \emph{CVPR}, 2017.

\bibitem[Liu et~al.(2023)Liu, Wu, Hoorick, Tokmakov, Zakharov, and Vondrick]{liu2023zero1to3}
Ruoshi Liu, Rundi Wu, Basile~Van Hoorick, Pavel Tokmakov, Sergey Zakharov, and Carl Vondrick.
\newblock Zero-1-to-3: Zero-shot one image to 3d object.
\newblock \emph{arXiv preprint arXiv:2303.11328}, 2023.

\bibitem[Liu et~al.(2015)Liu, Luo, Wang, and Tang]{liu2015faceattributes}
Ziwei Liu, Ping Luo, Xiaogang Wang, and Xiaoou Tang.
\newblock Deep learning face attributes in the wild.
\newblock In \emph{Proceedings of International Conference on Computer Vision (ICCV)}, December 2015.

\bibitem[Luo \& Hu(2021)Luo and Hu]{Luo2021CVPR}
Shitong Luo and Wei Hu.
\newblock Diffusion probabilistic models for 3d point cloud generation.
\newblock In \emph{CVPR}, 2021.

\bibitem[Mescheder et~al.(2018)Mescheder, Geiger, and Nowozin]{Mescheder2018ICML}
Lars Mescheder, Andreas Geiger, and Sebastian Nowozin.
\newblock Which training methods for gans do actually converge?
\newblock In \emph{Proc. of the International Conf. on Machine learning (ICML)}, 2018.

\bibitem[Metzer et~al.(2022)Metzer, Richardson, Patashnik, Giryes, and Cohen-Or]{metzer2022latentnerf}
Gal Metzer, Elad Richardson, Or~Patashnik, Raja Giryes, and Daniel Cohen-Or.
\newblock Latent-nerf for shape-guided generation of 3d shapes and textures.
\newblock \emph{arXiv preprint arXiv:2211.07600}, 2022.

\bibitem[Mi et~al.(2022)Mi, Kundu, Ross, Dellaert, Snavely, and Fathi]{Mi2022ARXIV}
Lu~Mi, Abhijit Kundu, David~A. Ross, Frank Dellaert, Noah Snavely, and Alireza Fathi.
\newblock im2nerf: Image to neural radiance field in the wild.
\newblock \emph{arXiv}, 2022.

\bibitem[Miangoleh et~al.(2021)Miangoleh, Dille, Mai, Paris, and Aksoy]{Miangoleh2021CVPR}
S.~Mahdi~H. Miangoleh, Sebastian Dille, Long Mai, Sylvain Paris, and Yagiz Aksoy.
\newblock Boosting monocular depth estimation models to high-resolution via content-adaptive multi-resolution merging.
\newblock In \emph{CVPR}, 2021.

\bibitem[Mildenhall et~al.(2020)Mildenhall, Srinivasan, Tancik, Barron, Ramamoorthi, and Ng]{Mildenhall2020ECCV}
Ben Mildenhall, Pratul~P Srinivasan, Matthew Tancik, Jonathan~T Barron, Ravi Ramamoorthi, and Ren Ng.
\newblock {NeRF}: Representing scenes as neural radiance fields for view synthesis.
\newblock In \emph{ECCV}, 2020.

\bibitem[Mokady et~al.(2022)Mokady, Yarom, Tov, Lang, Cohen-Or, Dekel, Irani, and Mosseri]{mokady2022selfdistilled}
Ron Mokady, Michal Yarom, Omer Tov, Oran Lang, Daniel Cohen-Or, Tali Dekel, Michal Irani, and Inbar Mosseri.
\newblock Self-distilled stylegan: Towards generation from internet photos, 2022.

\bibitem[Nam et~al.(2022)Nam, Khlifi, Rodriguez, Tono, Zhou, and Guerrero]{nam20223dldm}
Gimin Nam, Mariem Khlifi, Andrew Rodriguez, Alberto Tono, Linqi Zhou, and Paul Guerrero.
\newblock 3d-ldm: Neural implicit 3d shape generation with latent diffusion models.
\newblock \emph{arXiv preprint arXiv:2212.00842}, 2022.

\bibitem[Nguyen-Phuoc et~al.(2019)Nguyen-Phuoc, Li, Theis, Richardt, and Yang]{Nguyen2019ICCV}
Thu Nguyen-Phuoc, Chuan Li, Lucas Theis, Christian Richardt, and Yong-Liang Yang.
\newblock Hologan: Unsupervised learning of 3d representations from natural images.
\newblock In \emph{ICCV}, 2019.

\bibitem[Nichol et~al.(2022{\natexlab{a}})Nichol, Jun, Dhariwal, Mishkin, and Chen]{nichol2022pointe}
Alex Nichol, Heewoo Jun, Prafulla Dhariwal, Pamela Mishkin, and Mark Chen.
\newblock Point-e: A system for generating 3d point clouds from complex prompts.
\newblock \emph{arXiv preprint arXiv:2212.08751}, 2022{\natexlab{a}}.

\bibitem[Nichol \& Dhariwal(2021)Nichol and Dhariwal]{Nichol2021ICML}
Alexander~Quinn Nichol and Prafulla Dhariwal.
\newblock Improved denoising diffusion probabilistic models.
\newblock In Marina Meila and Tong Zhang (eds.), \emph{Proceedings of the 38th International Conference on Machine Learning, {ICML} 2021, 18-24 July 2021, Virtual Event}, 2021.

\bibitem[Nichol et~al.(2022{\natexlab{b}})Nichol, Dhariwal, Ramesh, Shyam, Mishkin, McGrew, Sutskever, and Chen]{Nichol2022ICML}
Alexander~Quinn Nichol, Prafulla Dhariwal, Aditya Ramesh, Pranav Shyam, Pamela Mishkin, Bob McGrew, Ilya Sutskever, and Mark Chen.
\newblock {GLIDE:} towards photorealistic image generation and editing with text-guided diffusion models.
\newblock In Kamalika Chaudhuri, Stefanie Jegelka, Le~Song, Csaba Szepesv{\'{a}}ri, Gang Niu, and Sivan Sabato (eds.), \emph{Proc. of the International Conf. on Machine learning (ICML)}, 2022{\natexlab{b}}.

\bibitem[Niemeyer \& Geiger(2021{\natexlab{a}})Niemeyer and Geiger]{Niemeyer2021CVPR}
Michael Niemeyer and Andreas Geiger.
\newblock Giraffe: Representing scenes as compositional generative neural feature fields.
\newblock In \emph{CVPR}, 2021{\natexlab{a}}.

\bibitem[Niemeyer \& Geiger(2021{\natexlab{b}})Niemeyer and Geiger]{Niemeyer2021THREEDV}
Michael Niemeyer and Andreas Geiger.
\newblock Campari: Camera-aware decomposed generative neural radiance fields.
\newblock In \emph{Proc. of the International Conf. on 3D Vision (3DV)}, 2021{\natexlab{b}}.

\bibitem[Or-El et~al.(2022)Or-El, Luo, Shan, Shechtman, Park, and Kemelmacher]{OrEl2021ARXIV}
Roy Or-El, Xuan Luo, Mengyi Shan, Eli Shechtman, Jeong Park, and Ira Kemelmacher.
\newblock Stylesdf: High-resolution 3d-consistent image and geometry generation.
\newblock In \emph{CVPR}, 2022.

\bibitem[Pan et~al.(2021)Pan, Xu, Loy, Theobalt, and Dai]{Pan2021NEURIPS}
Xingang Pan, Xudong Xu, Chen~Change Loy, Christian Theobalt, and Bo~Dai.
\newblock A shading-guided generative implicit model for shape-accurate 3d-aware image synthesis.
\newblock In \emph{NeurIPS}, 2021.

\bibitem[Pavllo et~al.(2022)Pavllo, Tan, Rakotosaona, and Tombari]{Pavollo2022ARXIV}
Dario Pavllo, David~Joseph Tan, Marie{-}Julie Rakotosaona, and Federico Tombari.
\newblock Shape, pose, and appearance from a single image via bootstrapped radiance field inversion.
\newblock \emph{arXiv}, 2022.

\bibitem[Peng et~al.(2020)Peng, Niemeyer, Mescheder, Pollefeys, and Geiger]{Peng2020ECCV}
Songyou Peng, Michael Niemeyer, Lars Mescheder, Marc Pollefeys, and Andreas Geiger.
\newblock Convolutional occupancy networks.
\newblock In \emph{ECCV}, 2020.

\bibitem[Poole et~al.(2022)Poole, Jain, Barron, and Mildenhall]{poole2022dreamfusion}
Ben Poole, Ajay Jain, Jonathan~T. Barron, and Ben Mildenhall.
\newblock Dreamfusion: Text-to-3d using 2d diffusion.
\newblock \emph{arXiv}, 2022.

\bibitem[Ramesh et~al.(2022)Ramesh, Dhariwal, Nichol, Chu, and Chen]{Ramesh2022ARXIV}
Aditya Ramesh, Prafulla Dhariwal, Alex Nichol, Casey Chu, and Mark Chen.
\newblock Hierarchical text-conditional image generation with {CLIP} latents.
\newblock \emph{arXiv}, abs/2204.06125, 2022.

\bibitem[Ranftl et~al.(2020)Ranftl, Lasinger, Hafner, Schindler, and Koltun]{Ranftl2020PAMI}
Ren{\'e} Ranftl, Katrin Lasinger, David Hafner, Konrad Schindler, and Vladlen Koltun.
\newblock Towards robust monocular depth estimation: Mixing datasets for zero-shot cross-dataset transfer.
\newblock \emph{IEEE TPAMI}, 2020.

\bibitem[Rebain et~al.(2021)Rebain, Matthews, Yi, Lagun, and Tagliasacchi]{Rebain2021ARXIV}
Daniel Rebain, Mark Matthews, Kwang~Moo Yi, Dmitry Lagun, and Andrea Tagliasacchi.
\newblock Lolnerf: Learn from one look.
\newblock \emph{arXiv}, 2021.

\bibitem[Rezende et~al.(2014)Rezende, Mohamed, and Wierstra]{Rezende2014ICML}
Danilo~Jimenez Rezende, Shakir Mohamed, and Daan Wierstra.
\newblock Stochastic backpropagation and approximate inference in deep generative models.
\newblock In \emph{Proc. of the International Conf. on Machine learning (ICML)}, 2014.

\bibitem[Richardson et~al.(2021)Richardson, Alaluf, Patashnik, Nitzan, Azar, Shapiro, and Cohen-Or]{richardson2021encoding}
Elad Richardson, Yuval Alaluf, Or~Patashnik, Yotam Nitzan, Yaniv Azar, Stav Shapiro, and Daniel Cohen-Or.
\newblock Encoding in style: a stylegan encoder for image-to-image translation.
\newblock In \emph{Proceedings of the IEEE/CVF Conference on Computer Vision and Pattern Recognition}, pp.\  2287--2296, 2021.

\bibitem[Roich et~al.(2023)Roich, Mokady, Bermano, and Cohen{-}Or]{RoichTOG2023}
Daniel Roich, Ron Mokady, Amit~H. Bermano, and Daniel Cohen{-}Or.
\newblock Pivotal tuning for latent-based editing of real images.
\newblock \emph{ACM TOG}, 2023.

\bibitem[Rombach et~al.(2021)Rombach, Blattmann, Lorenz, Esser, and Ommer]{rombach2021highresolution}
Robin Rombach, Andreas Blattmann, Dominik Lorenz, Patrick Esser, and Björn Ommer.
\newblock High-resolution image synthesis with latent diffusion models, 2021.

\bibitem[Saharia et~al.(2022)Saharia, Chan, Saxena, Li, Whang, Denton, Ghasemipour, Ayan, Mahdavi, Lopes, et~al.]{saharia2022photorealistic}
Chitwan Saharia, William Chan, Saurabh Saxena, Lala Li, Jay Whang, Emily Denton, Seyed Kamyar~Seyed Ghasemipour, Burcu~Karagol Ayan, S~Sara Mahdavi, Rapha~Gontijo Lopes, et~al.
\newblock Photorealistic text-to-image diffusion models with deep language understanding.
\newblock \emph{arXiv preprint arXiv:2205.11487}, 2022.

\bibitem[Sajjadi et~al.(2018)Sajjadi, Bachem, Lucic, Bousquet, and Gelly]{sajjadi2018assessing}
Mehdi~SM Sajjadi, Olivier Bachem, Mario Lucic, Olivier Bousquet, and Sylvain Gelly.
\newblock Assessing generative models via precision and recall.
\newblock \emph{Advances in Neural Information Processing Systems}, 31, 2018.

\bibitem[Salimans \& Ho(2022)Salimans and Ho]{salimans2022progressive}
Tim Salimans and Jonathan Ho.
\newblock Progressive distillation for fast sampling of diffusion models.
\newblock \emph{arXiv preprint arXiv:2202.00512}, 2022.

\bibitem[Sargent et~al.(2023)Sargent, Koh, Zhang, Chang, Herrmann, Srinivasan, Wu, and Sun]{sargent2023vq3d}
Kyle Sargent, Jing~Yu Koh, Han Zhang, Huiwen Chang, Charles Herrmann, Pratul Srinivasan, Jiajun Wu, and Deqing Sun.
\newblock Vq3d: Learning a 3d-aware generative model on imagenet.
\newblock In \emph{Proceedings of the IEEE/CVF International Conference on Computer Vision (ICCV)}, pp.\  4240--4250, October 2023.

\bibitem[Sauer et~al.(2022)Sauer, Schwarz, and Geiger]{Sauer2022ARXIV}
Axel Sauer, Katja Schwarz, and Andreas Geiger.
\newblock Stylegan-xl: Scaling stylegan to large diverse datasets.
\newblock \emph{ACM Trans. on Graphics}, 2022.

\bibitem[Schwarz et~al.(2020)Schwarz, Liao, Niemeyer, and Geiger]{Schwarz2020NIPS}
Katja Schwarz, Yiyi Liao, Michael Niemeyer, and Andreas Geiger.
\newblock {GRAF:} generative radiance fields for 3d-aware image synthesis.
\newblock \emph{NeurIPS}, 2020.

\bibitem[Schwarz et~al.(2022)Schwarz, Sauer, Niemeyer, Liao, and Geiger]{Schwarz2022NIPS}
Katja Schwarz, Axel Sauer, Michael Niemeyer, Yiyi Liao, and Andreas Geiger.
\newblock Voxgraf: Fast 3d-aware image synthesis with sparse voxel grids.
\newblock In \emph{NeurIPS}, 2022.

\bibitem[Shi et~al.(2023)Shi, Shen, Xu, Peng, Liao, Guo, Chen, and Yeung]{Shi2023CVPR}
Zifan Shi, Yujun Shen, Yinghao Xu, Sida Peng, Yiyi Liao, Sheng Guo, Qifeng Chen, and Dit{-}Yan Yeung.
\newblock Learning 3d-aware image synthesis with unknown pose distribution.
\newblock In \emph{CVPR}, 2023.

\bibitem[Shue et~al.(2022)Shue, Chan, Po, Ankner, Wu, and Wetzstein]{shue2022triplanediffusion}
J.~Ryan Shue, Eric~Ryan Chan, Ryan Po, Zachary Ankner, Jiajun Wu, and Gordon Wetzstein.
\newblock 3d neural field generation using triplane diffusion.
\newblock \emph{arXiv preprint arXiv:2211.16677}, 2022.

\bibitem[Skorokhodov et~al.(2022)Skorokhodov, Tulyakov, Wang, and Wonka]{Skorokhodov2022ARXIV}
Ivan Skorokhodov, Sergey Tulyakov, Yiqun Wang, and Peter Wonka.
\newblock Epigraf: Rethinking training of 3d gans.
\newblock \emph{arXiv}, 2022.

\bibitem[Skorokhodov et~al.(2023)Skorokhodov, Siarohin, Xu, Ren, Lee, Wonka, and Tulyakov]{skorokhodov20233dgp}
Ivan Skorokhodov, Aliaksandr Siarohin, Yinghao Xu, Jian Ren, Hsin-Ying Lee, Peter Wonka, and Sergey Tulyakov.
\newblock 3d generation on imagenet.
\newblock In \emph{The Eleventh International Conference on Learning Representations}, 2023.

\bibitem[Sohl-Dickstein et~al.(2015)Sohl-Dickstein, Weiss, Maheswaranathan, and Ganguli]{sohl2015deep}
Jascha Sohl-Dickstein, Eric Weiss, Niru Maheswaranathan, and Surya Ganguli.
\newblock Deep unsupervised learning using nonequilibrium thermodynamics.
\newblock In \emph{International Conference on Machine Learning}, pp.\  2256--2265. PMLR, 2015.

\bibitem[Song et~al.(2021)Song, Meng, and Ermon]{song2020denoising2}
Jiaming Song, Chenlin Meng, and Stefano Ermon.
\newblock Denoising diffusion implicit models.
\newblock In \emph{International Conference on Learning Representations}, 2021.

\bibitem[Song \& Ermon(2019)Song and Ermon]{song2019generative}
Yang Song and Stefano Ermon.
\newblock Generative modeling by estimating gradients of the data distribution.
\newblock In \emph{Proceedings of the 33rd Annual Conference on Neural Information Processing Systems}, 2019.

\bibitem[Song et~al.(2020)Song, Sohl-Dickstein, Kingma, Kumar, Ermon, and Poole]{song2020score}
Yang Song, Jascha Sohl-Dickstein, Diederik~P Kingma, Abhishek Kumar, Stefano Ermon, and Ben Poole.
\newblock Score-based generative modeling through stochastic differential equations.
\newblock \emph{arXiv preprint arXiv:2011.13456}, 2020.

\bibitem[Tov et~al.(2021)Tov, Alaluf, Nitzan, Patashnik, and Cohen-Or]{tov2021designing}
Omer Tov, Yuval Alaluf, Yotam Nitzan, Or~Patashnik, and Daniel Cohen-Or.
\newblock Designing an encoder for stylegan image manipulation.
\newblock \emph{ACM Transactions on Graphics (TOG)}, 40\penalty0 (4):\penalty0 1--14, 2021.

\bibitem[Vahdat et~al.(2021)Vahdat, Kreis, and Kautz]{Vahdat2021NIPS}
Arash Vahdat, Karsten Kreis, and Jan Kautz.
\newblock Score-based generative modeling in latent space.
\newblock In \emph{NeurIPS}, 2021.

\bibitem[Vaswani et~al.(2017)Vaswani, Shazeer, Parmar, Uszkoreit, Jones, Gomez, Kaiser, and Polosukhin]{Vaswani2017NIPS}
Ashish Vaswani, Noam Shazeer, Niki Parmar, Jakob Uszkoreit, Llion Jones, Aidan~N. Gomez, Lukasz Kaiser, and Illia Polosukhin.
\newblock Attention is all you need.
\newblock In \emph{NeurIPS}, pp.\  5998--6008, 2017.

\bibitem[Vincent(2011)]{vincent2011}
Pascal Vincent.
\newblock A connection between score matching and denoising autoencoders.
\newblock \emph{Neural Computation}, 23\penalty0 (7):\penalty0 1661--1674, 2011.

\bibitem[Wang et~al.(2022{\natexlab{a}})Wang, Du, Li, Yeh, and Shakhnarovich]{wang2022scorejacobian}
Haochen Wang, Xiaodan Du, Jiahao Li, Raymond~A. Yeh, and Greg Shakhnarovich.
\newblock Score jacobian chaining: Lifting pretrained 2d diffusion models for 3d generation.
\newblock \emph{arXiv preprint arXiv:2212.00774}, 2022{\natexlab{a}}.

\bibitem[Wang et~al.(2022{\natexlab{b}})Wang, Zhang, Zhang, Gu, Bao, Baltrusaitis, Shen, Chen, Wen, Chen, and Guo]{wang2022rodin}
Tengfei Wang, Bo~Zhang, Ting Zhang, Shuyang Gu, Jianmin Bao, Tadas Baltrusaitis, Jingjing Shen, Dong Chen, Fang Wen, Qifeng Chen, and Baining Guo.
\newblock Rodin: A generative model for sculpting 3d digital avatars using diffusion.
\newblock \emph{arXiv preprint arXiv:2212.06135}, 2022{\natexlab{b}}.

\bibitem[Watson et~al.(2022)Watson, Chan, Martin{-}Brualla, Ho, Tagliasacchi, and Norouzi]{Watson2022ARXIV}
Daniel Watson, William Chan, Ricardo Martin{-}Brualla, Jonathan Ho, Andrea Tagliasacchi, and Mohammad Norouzi.
\newblock Novel view synthesis with diffusion models.
\newblock \emph{arXiv}, 2022.

\bibitem[Xiang et~al.(2023{\natexlab{a}})Xiang, Yang, Deng, and Tong]{xiang2023gramhd}
Jianfeng Xiang, Jiaolong Yang, Yu~Deng, and Xin Tong.
\newblock Gram-hd: 3d-consistent image generation at high resolution with generative radiance manifolds.
\newblock In \emph{Proceedings of the IEEE/CVF International Conference on Computer Vision (ICCV)}, pp.\  2195--2205, October 2023{\natexlab{a}}.

\bibitem[Xiang et~al.(2023{\natexlab{b}})Xiang, Yang, Huang, and Tong]{xiang2023ivid}
Jianfeng Xiang, Jiaolong Yang, Binbin Huang, and Xin Tong.
\newblock 3d-aware image generation using 2d diffusion models.
\newblock In \emph{Proceedings of the IEEE/CVF International Conference on Computer Vision (ICCV)}, pp.\  2383--2393, October 2023{\natexlab{b}}.

\bibitem[Xie et~al.(2021)Xie, Wang, Yu, Anandkumar, Alvarez, and Luo]{XieNEURIPS2021}
Enze Xie, Wenhai Wang, Zhiding Yu, Anima Anandkumar, Jose~M. Alvarez, and Ping Luo.
\newblock Segformer: Simple and efficient design for semantic segmentation with transformers.
\newblock In Marc'Aurelio Ranzato, Alina Beygelzimer, Yann~N. Dauphin, Percy Liang, and Jennifer~Wortman Vaughan (eds.), \emph{NeurIPS}, 2021.

\bibitem[Xie et~al.(2022)Xie, Ouyang, Piao, Lei, and Chen]{Xie2022ARXIV}
Jiaxin Xie, Hao Ouyang, Jingtan Piao, Chenyang Lei, and Qifeng Chen.
\newblock High-fidelity 3d {GAN} inversion by pseudo-multi-view optimization.
\newblock \emph{arXiv}, 2022.

\bibitem[Xu et~al.(2021)Xu, Pan, Lin, and Dai]{Xu2021NEURIPS}
Xudong Xu, Xingang Pan, Dahua Lin, and Bo~Dai.
\newblock Generative occupancy fields for 3d surface-aware image synthesis.
\newblock In \emph{NeurIPS}, 2021.

\bibitem[Xu et~al.(2022)Xu, Peng, Yang, Shen, and Zhou]{Xu2021ARXIV}
Yinghao Xu, Sida Peng, Ceyuan Yang, Yujun Shen, and Bolei Zhou.
\newblock 3d-aware image synthesis via learning structural and textural representations.
\newblock \emph{CVPR}, 2022.

\bibitem[Yin et~al.(2022)Yin, Zhang, Wang, Wang, Li, Gong, Fan, Cun, Shan, {\"{O}}ztireli, and Yang]{Yin2022ARXIV}
Fei Yin, Yong Zhang, Xuan Wang, Tengfei Wang, Xiaoyu Li, Yuan Gong, Yanbo Fan, Xiaodong Cun, Ying Shan, Cengiz {\"{O}}ztireli, and Yujiu Yang.
\newblock 3d {GAN} inversion with facial symmetry prior.
\newblock \emph{arXiv}, 2022.

\bibitem[Yu et~al.(2015)Yu, Xiao, and Funkhouser]{Yu2015CVPR}
Fisher Yu, Jianxiong Xiao, and Thomas~A. Funkhouser.
\newblock Semantic alignment of lidar data at city scale.
\newblock In \emph{CVPR}, pp.\  1722--1731, 2015.

\bibitem[Yu et~al.(2022)Yu, Peng, Niemeyer, Sattler, and Geiger]{Yu2022MonoSDF}
Zehao Yu, Songyou Peng, Michael Niemeyer, Torsten Sattler, and Andreas Geiger.
\newblock Monosdf: Exploring monocular geometric cues for neural implicit surface reconstruction.
\newblock \emph{NeurIPS}, 2022.

\bibitem[Zeng et~al.(2022)Zeng, Vahdat, Williams, Gojcic, Litany, Fidler, and Kreis]{Zeng2022ARXIV}
Xiaohui Zeng, Arash Vahdat, Francis Williams, Zan Gojcic, Or~Litany, Sanja Fidler, and Karsten Kreis.
\newblock {LION:} latent point diffusion models for 3d shape generation.
\newblock In \emph{NeurIPS}, 2022.

\bibitem[Zhan et~al.(2018)Zhan, Garg, Saroj~Weerasekera, Li, Agarwal, and Reid]{Zhan2018CVPR}
Huangying Zhan, Ravi Garg, Chamara Saroj~Weerasekera, Kejie Li, Harsh Agarwal, and Ian Reid.
\newblock Unsupervised learning of monocular depth estimation and visual odometry with deep feature reconstruction.
\newblock In \emph{CVPR}, 2018.

\bibitem[Zhang et~al.(2021)Zhang, Sangineto, Tang, Siarohin, Zhong, Sebe, and Wang]{Zhang2021ARXIV}
Jichao Zhang, Enver Sangineto, Hao Tang, Aliaksandr Siarohin, Zhun Zhong, Nicu Sebe, and Wei Wang.
\newblock 3d-aware semantic-guided generative model for human synthesis.
\newblock \emph{arXiv}, 2021.

\bibitem[Zhang et~al.(2018)Zhang, Isola, Efros, Shechtman, and Wang]{Zhang2018CVPRb}
Richard Zhang, Phillip Isola, Alexei~A. Efros, Eli Shechtman, and Oliver Wang.
\newblock The unreasonable effectiveness of deep features as a perceptual metric.
\newblock In \emph{CVPR}, 2018.

\bibitem[Zhou et~al.(2021{\natexlab{a}})Zhou, Du, and Wu]{Zhou2021ICCV}
Linqi Zhou, Yilun Du, and Jiajun Wu.
\newblock 3d shape generation and completion through point-voxel diffusion.
\newblock In \emph{ICCV}, 2021{\natexlab{a}}.

\bibitem[Zhou et~al.(2021{\natexlab{b}})Zhou, Xie, Ni, and Tian]{Zhou2021ARXIV}
Peng Zhou, Lingxi Xie, Bingbing Ni, and Qi~Tian.
\newblock {{CIPS}}-{{3D}}: A {{3D}}-{{Aware Generator}} of {{GANs Based}} on {{Conditionally}}-{{Independent Pixel Synthesis}}.
\newblock \emph{arXiv}, 2021{\natexlab{b}}.

\bibitem[Zhu et~al.(2020)Zhu, Shen, Zhao, and Zhou]{zhu2020domain}
Jiapeng Zhu, Yujun Shen, Deli Zhao, and Bolei Zhou.
\newblock In-domain gan inversion for real image editing.
\newblock In \emph{European conference on computer vision}, pp.\  592--608. Springer, 2020.

\end{thebibliography}

\clearpage
\appendix
\section{Theoretical background}

\textit{We summarize the most relevant theoretical concepts to our work in the following.}

\label{sec:preliminaries_dm}
\textbf{Diffusion Models}~\citep{sohl2015deep,ho2020ddpm2,song2020score} create diffused inputs $\mathbf{x}_\tau{=}\alpha_\tau \mathbf{x}{+}\sigma_\tau \boldsymbol{\epsilon}, \; \boldsymbol{\epsilon}{\sim}\mathcal{N}(\mathbf{0}, \mathbf{I})$ from data $\mathbf{x}{\sim}p_{\text{data}}$, where $\alpha_{\tau}$ and $\sigma_\tau$ define a noise schedule, parameterized by a diffusion-time $\tau$.
A denoiser model $\mathcal{F}_\omega$ with parameters $\omega$ is trained to denoise the perturbed data via denoising score matching~\citep{hyvarinen2005scorematching,vincent2011,song2019generative},\looseness=-1
\begin{align}
\arg \min_\omega\; \mathbb{E}_{\mathbf{x} \sim p_{\text{data}}, \tau \sim p_{\tau}, \boldsymbol{\epsilon} \sim \mathcal{N}(\mathbf{0}, \mathbf{I})} \left[\Vert \mathbf{v} - \mathcal{F}_\omega(\mathbf{x}_\tau, \tau) \Vert_2^2 \right],
\label{eq:diffusionobjective}
\end{align}
with the target $\mathbf{v} = \alpha_\tau \boldsymbol{\epsilon} - \sigma_\tau \mathbf{x}$ (this is known as \textit{$\mathbf{v}$-prediction}~\citep{salimans2022progressive}). Further, $p_\tau$ is a uniform distribution over the diffusion time $\tau$, such that the model is trained to denoise for all different times $\tau$. The noise schedule is designed such that input data is entirely perturbed into Gaussian random noise after the maximum diffusion time. An iterative generative denoising process that employs the learned denoiser $\mathcal{F}_\omega$ can then be initialized from such Gaussian noise to synthesize novel data. Classifier-free guidance can be used to amplify conditioning strength when conditioning the diffusion model on data such as classes; see ~\citet{ho2021classifierfree} and App.~\ref{supp:some_details}.

Diffusion models have also been applied to 3D data~\citep{Zeng2022ARXIV,wang2022rodin,Bautista2022ARXIV,shue2022triplanediffusion,nam20223dldm} but usually require explicit 3D or multiview supervision. 
In contrast,\ourmodel learns from an unstructured image set without multiview supervision.\looseness=-1

\textbf{Latent Diffusion Models (LDMs)}~\citep{rombach2021highresolution,Vahdat2021NIPS} first train a regularized autoencoder with encoder $\mathcal{E}$ and decoder $\mathcal{D}$ to transform input images $\mathbf{I} \sim p_{\text{data}}$ into a spatially lower-dimensional latent space $\mathbf{Z}$ of reduced complexity, from which the original data can be reconstructed,
this is, $\hat{\mathbf{I}} = \mathcal{D}(\mathcal{E}(\mathbf{I})) \approx \mathbf{I}$.
A diffusion model is then trained in the compressed latent space,
with $\mathbf{x}$ in Eq.~(\ref{eq:diffusionobjective}) replaced by an image's latent representation $\mathbf{Z} = \mathcal{E}(\mathbf{I})$. This latent space diffusion model can be typically smaller in terms of parameter count and memory consumption compared to corresponding pixel-space diffusion models of similar performance. More diffusion model details in Appendix.\looseness=-1

\textbf{3D-Representations for 3D-Aware Image Synthesis.}
\label{sec:preliminarieseg3d}
3D-aware generative models typically generate neural radiance fields or feature fields, \ie, they represent a scene by generating a color or a feature value $\mathbf{f}$ and a density $\sigma$ for each 3D point $\mathbf{p}\in\mathbb{R}^3$~\citep{Mildenhall2020ECCV,Schwarz2020NIPS, Niemeyer2021CVPR}. 
Features and densities can be efficiently computed from a triplane representation $[\mathbf{T}_{xy}, \mathbf{T}_{xz}, \mathbf{T}_{yz}]$~\citep{Chan2022CVPR,Peng2020ECCV}. The triplane feature $\mathbf{t}$ is obtained by projecting $\mathbf{p}$ onto each of the three feature planes and averaging their feature vectors $(\mathbf{t}_{xy}, \mathbf{t}_{xz}, \mathbf{t}_{yz})$. An MLP then converts the triplane feature $\mathbf{t}$ to a feature and density value $ [\mathbf{f}, \sigma] = MLP(\mathbf{t})$.

Given a camera pose, the feature field is rendered via volume rendering~\citep{Kajiya1984SIGGRAPH,Mildenhall2020ECCV}. 
For that, the feature field is evaluated at discrete points $\mathbf{p}^i_r$ along each camera ray $r$  yielding features and densities $\{(\mathbf{f}_r^i, \sigma_r^i)\}_{i=1}^N$.
For each ray $r$, these features are aggregated to a feature $\mathbf{f}_r$ using alpha composition
\begin{gather}
{
\mathbf{f}_r = \sum_{i=1}^N \, w_r^i \, \mathbf{f}_r^i, \hspace{0.6cm} w_r^i = T_r^i \, \alpha_r^i,\hspace{0.6cm}
T_r^i = \prod_{j=1}^{i-1}\left(1-\alpha_r^j\right),\hspace{0.6cm}
\alpha_r^i = 1-\exp\left(-\sigma_r^i\delta_r^i\right),
}
\label{eq:volrend}
\end{gather}
where $T_r^i$ and $\alpha_r^i$ denote the transmittance and alpha value of sample point $\mathbf{p}^i_r$ along ray $r$
and $\delta_r^i=\left\Vert\mathbf{p}_r^{i+1}-\mathbf{p}_r^{i}\right\Vert_2$ is the distance between neighboring sample points. Similarly, depth can be rendered, see Appendix.
For efficiency, a low-resolution feature map, and optionally a low-resolution image $\mathbf{\hat{I}}^{low}$, can be rendered instead of an image at full resolution~\citep{Niemeyer2021CVPR,Chan2022CVPR}. The feature map is then subsequently upsampled and decoded into a higher-resolution image $\mathbf{\hat{I}}$.\looseness=-1

\section{Related Work} \label{app:related_work}
\textit{Here, we present an extended discussion about related work.}

\boldparagraph{Diffusion Models.}
Diffusion models (DMs)~\citep{sohl2015deep,ho2020ddpm2,song2020score} have proven to be powerful image generators, yielding state-of-the art results in unconditional as well as class- and text-guided synthesis~\citep{Nichol2021ICML,rombach2021highresolution,dhariwal2021diffusion,Ho2022MLRes,Dockhorn2022ICLR,dockhorn2022genie,Vahdat2021NIPS,Nichol2022ICML,Ramesh2022ARXIV,saharia2022photorealistic,balaji2022eDiffi}. However, none of these works tackles 3D-aware image synthesis.

\boldparagraph{3D Diffusion Models.} There is also much literature on applying diffusion models to 3D data, \eg 3D point clouds ~\citep{Zhou2021ICCV,Luo2021CVPR,Zeng2022ARXIV} or tetrahedral meshes~\citep{kalischek2022tetrahedral}. 
\cite{shue2022triplanediffusion} learn a diffusion model on a triplane representation parameterizing a neural occupancy field. GAUDI~\citep{Bautista2022ARXIV} and 3D-LDM~\citep{nam20223dldm} train diffusion models in latent spaces learnt using an autodecoder framework and generate 3D scenes and 3D shapes, respectively.
RODIN~\citep{wang2022rodin} proposes a hierarchical latent diffusion model framework to learn 3D human avatars and NF-LDM~\citep{kim2023nfldm} trains a hierarchical diffusion model for outdoor scene generation. 
\cite{dupont2022functa} and \cite{du2021gem} treat data as functions and also explore encoding 3D signals into latent spaces, but using more inefficient meta-learning~\citep{dupont2022functa} or auto-decoder~\citep{du2021gem} methods. \cite{dupont2022functa} also trains a diffusion model on the encoded 3D data. However, all aforementioned works rely on explicit 3D or multiview supervision. In contrast, our approach learns from an unstructured image collection without multiview
supervision.

RenderDiffusion~\citep{anciukevicius2022renderdiffusion} trains a diffusion model directly on images, using a triplanar 3D feature representation inside the denoiser network architecture, thereby enabling 3D-aware image generation during synthesis. However, scaling RenderDiffusion to high-resolution outputs is challenging, as it operates directly on images. In fact, it considers only small image resolutions of 32x32 or 64x64, likely due to computational limitations. When trained on single-view real-world data, the paper only considers data with little pose variation (human and cat faces) and it is unclear whether the approach is scalable to diverse or high-resolution image data (moreover, no perceptual quality evaluations on metrics such as FID are presented and there is no code available). In contrast, our diffusion model is trained efficiently in a low-resolution, compressed and 3D-aware latent space, while simultaneously predicting high-resolution triplanes and enabling high-resolution rendering. Hence, WildFusion generates significantly higher quality 3D-aware images than RenderDiffusion and it is scalable to diverse datasets such as ImageNet, as we demonstrate.

Concurrently with us, IVID~\citep{xiang2023ivid} trains a 2D diffusion model that first synthesizes an initial image and then iteratively generates novel views conditioned on it. However, the iterative generation is extremely slow because it requires running the full reverse diffusion process for every novel view. Further, an explicit 3D representation can only be constructed indirectly from a large collection of generated multi-view images, afterwards. Instead, \ourmodel uses a fundamentally different approach and only runs the reverse diffusion process once to generate a (latent) 3D representation from which multi-view images can be rendered directly and geometry can be extracted easily. At time of submission code for experimental comparisons to IVID was not available.

\boldparagraph{Optimization from Text-to-Image Diffusion Models.} Another line of work distills 3D objects from large-scale 2D text-to-image diffusion models~\citep{poole2022dreamfusion,lin2022magic3d,nichol2022pointe,metzer2022latentnerf,wang2022scorejacobian,deng2022nerdi}. However, these methods follow an entirely different approach compared to 3D- and 3D-aware diffusion models and require a slow optimization process that needs to be run per instance.

\boldparagraph{3D-Aware Image Synthesis.}
3D-aware generative models consider image synthesis with control over the camera viewpoint~\citep{Liao2020CVPRa,Schwarz2020NIPS,Chan2020CVPR}. 
Most existing works rely on generative adversarial networks (GANs)~\citep{Goodfellow2014NIPS} and use coordinate-based MLPs as 3D-generator~\citep{Schwarz2020NIPS,Chan2020CVPR,Gu2021ARXIV,Jo2021ARXIV,Xu2021ARXIV,Zhou2021ARXIV,Zhang2021ARXIV,OrEl2021ARXIV,Xu2021NEURIPS,Pan2021NEURIPS,DENG2021ARXIV,xiang2023gramhd,Gu2021ARXIV}, building on Neural Radiance Fields~\citep{Mildenhall2020ECCV} as 3D representation.
VoxGRAF~\citep{Schwarz2022NIPS} and EG3D~\citep{Chan2022CVPR} proposed efficient convolutional 3D generators that require only a single forward pass for generating a 3D scene. 
Our autoencoder uses EG3D's triplane representation and their dual discriminator to improve view consistency. 
Early 3D-aware generative models that do not require camera poses during training and operate in view space include HoloGAN~\cite{Nguyen2019ICCV} and PlatonicGAN~\citep{Henzler2019ICCV}. They are outperformed, for instance, by the more recent
StyleNeRF~\citep{Gu2021ARXIV}, which uses a style-based architecture~\citep{Karras2019CVPR,Karras2020CVPRa} and proposes a novel path regularization loss to achieve 3D consistency. 
In contrast to the aforementioned approaches, however, our generative model is not a GAN. GANs are notoriously hard to train~\citep{Mescheder2018ICML} and often do not cover the data distribution well. 
Instead, we explore 3D-aware image synthesis with latent diffusion models for the first time.

Until recently, 3D-aware image synthesis focused on aligned datasets with well-defined pose distributions, such as portrait images~\citep{liu2015faceattributes,Karras2019CVPR}. For instance, POF3D~\citep{Shi2023CVPR} is a recent 3D-aware GAN that infers camera poses and works in a canonical view space; it has been used only for datasets with simple pose distributions, such as cat and human faces. 
Advancing to more complex datasets, GINA-3D learns to generate assets from driving data, assuming known camera and LIDAR sensors. The two-stage approach first trains a vision transformer encoder yielding a latent triplane representation. Next, a discrete token generative model is trained in the latent space. We consider a setting where camera information and ground truth depth are not available and train a latent diffusion model on 2D feature maps. \\
To scale 3D-aware image synthesis to more complex datasets, \ie ImageNet, 3DGP~\citep{skorokhodov20233dgp} proposes an elaborate camera model and learns to refine an initial prior on the pose distribution. Specifically, 3DGP predicts the camera location in a canonical coordinate system per class and sample-specific camera rotation and intrinsics. This assumes that samples within a class share a canonical system, and we observe that learning this complex distribution can aggravate training instability. Further, the approach needs to be trained on heavily filtered training data. In contrast, WildFusion can generate high-quality and diverse samples even when trained on the entire ImageNet dataset without any filtering (see Sec. \ref{sec:expdiffusion}).

Concurrently to \ourmodel, VQ3D~\citep{sargent2023vq3d} proposes an autoencoder architecture, but uses sequence-like latent variables and trains an autoregressive transformer in the latent space. Instead, \ourmodel trains a diffusion model on latent feature maps. Another difference is that VQ3D applies two discriminators on the generated images, one that distinguishes between reconstruction and training image, and another one that discriminates between reconstruction and novel view. \ourmodel only applies a single discriminator to supervise the novel views and instead has an additional discriminator on the depth. At time of submission code for experimental comparisons was not available.

\boldparagraph{Novel View Synthesis.} Our autoencoder is related to methods that synthesize novel views given a single input image: 
LoLNeRF~\citep{Rebain2021ARXIV} trains an autodecoder with per-pixel reconstruction loss and mask supervision.
In contrast, we add an adversarial objective to supervise novel views.
\cite{Mi2022ARXIV} proposes a similar approach but is not investigated in the context of generative modeling. Another line of recent works leverage GAN inversion to generate novel views from single input images~\citep{Li2022ARXIV,Cai2022CVPR,Lin2023WACV,Xie2022ARXIV,Yin2022ARXIV,Lan2022ARXIV,Pavollo2022ARXIV} but rely on pretrained 3D-aware GANs and thereby inherit their aforementioned limitations. 
Several recent works~\citep{Watson2022ARXIV,chan2023genvs,liu2023zero1to3} tackled novel view synthesis with view-conditioned 2D diffusion models, but are trained with explicit 3D or multiview supervision.
Unlike these approaches, we use our autoencoder to also train a 3D-aware generative model.

\section{Implementation Details}
\label{sec:supp_implementation}

\subsection{Camera System}
\label{sec:camsystem}
In the following, we describe the camera system we use to learn a 3D representation in view space. An overview is shown in~\figref{fig:camsystem}.
The input view $\mathbf{P}_0$ is defined by the camera intrinsics $\mathbf{K}_0$ and extrinsics $[\mathbf{R}_0, \mathbf{T}_0]$, where $\mathbf{R}_0$ and $\mathbf{T}_0$ denote the rotation and translation of the camera in the world coordinate system. 
We fix $\mathbf{R}_0$ and $\mathbf{T}_0$, such that the camera is located at a fixed radius and looks at the origin of the world coordinate center. For the intrinsics $\mathbf{K}_0$, we choose a small, fixed field of view since most images in the datasets are cropped and perspective distortion is generally small. For the experiments on all datasets, we set 
\begin{equation}
    \bR_0 = \begin{pmatrix}
1 & 0 & 0\\
0 & -1 & 0\\
0 & 0 & -1\\
\end{pmatrix} \quad
    \bT_0 = \begin{pmatrix}
0\\
0\\
2.7\\
\end{pmatrix} \quad
\bK_0 = \begin{pmatrix}
5.4 & 0 & 0.5\\
0 & 5.4 & 0.5\\
0 & 0 & 1.0\\ 
\end{pmatrix}\,.
\end{equation}
During training, we sample novel views around $\mathbf{P}_0$ by drawing offsets for azimuth and polar angles uniformly from $[-35^\circ, 35^\circ]$ and $[-15^\circ, 15^\circ]$, respectively.

As we consider unbounded scenes, we sample points along the camera rays linearly in disparity (inverse depth). Thereby, we effectively sample more points close to the camera and use fewer samples at large depths where fewer details need to be modeled. In practice, we set the near and far planes to $t_n=2.25$ and $t_f=5.0$, respectively. 
During rendering, the sampled points are mapped to 3D coordinates in $[-1, 1]$ and subsequently projected to 2D points on the triplanes (see~\figref{fig:camsystem}). Reflecting the disparity sampling, we choose a non-linear mapping function for the 3D coordinates that places sampling points with a large depth closer together. This assigns a larger area on the triplanes to points close to the camera while points far away are projected onto a smaller area. Specifically, we use a contraction function inspired by~\cite{barron2022mipnerf360} that maps points within a sphere of radius $r_s$ to a sphere of radius $r_{in} < 1$  and all points outside of the sphere to a sphere of radius $1$. 
Let $\bx\in\mathbb{R}^3$ denote a sampled point; then, the contracted coordinate $\bx_c$ is calculated as 
\begin{equation}\label{eq:contraction_function}
    \bx_c = \begin{cases}
      \bx\frac{r_{in}}{r}, & \text{if}\ ||\bx|| \leq r \\
      \left((1-r_{in})\left(1-\frac{1}{||\bx||-r+1}\right) + r_{in}\right) \frac{\bx}{||\bx||} , & \text{otherwise}
    \end{cases}
\end{equation}
We set $r_s=1.3$ and $r_{in}=0.8$ for all experiments.
\camsystem

\subsection{Network Architecture, Objectives, and Training} \label{supp:some_details}

\boldparagraph{First Stage: 3D-aware Autoencoder.}
The encoder network is a feature pyramid network (FPN)~\citep{Lin2017CVPRb}. In practice, we use the setup from variational autoencoders (VAEs)~\citep{Kingma2014ICLR,Rezende2014ICML} and predict means $\boldsymbol{\mu}$ and variances $\boldsymbol{\sigma}^2$ of a normal distribution from which we sample the latent representation $\mathbf{Z}$ 
\begin{align}
    [\boldsymbol{\mu},  \boldsymbol{\sigma}^2] = FPN(\mathbf{I}), \qquad \mathbf{Z} \sim \mathcal{N}(\boldsymbol{\mu},  \boldsymbol{\sigma}^2),
    \label{eq:ae_sampling}
\end{align}
where $\mathbf{Z}\in\mathbb{R}^{c\times h\times w}$ (formally, we assume a diagonal covariance matrix and predict means and variances for all latent dimensions independently).
We regularize the latent space by minimizing a low-weighted Kullback-Leibler divergence loss $\mathcal{L}_{KL}$ between $q_{\mathcal{E}}(\mathbf{Z}|\mathbf{I}_P)=
\mathcal{N}(\boldsymbol{\mu}, \boldsymbol{\sigma}^2)$ and a standard normal distribution $\mathcal{N}(\mathbf{0},\mathbf{I})$. 

The decoder consists of transformer blocks at resolution of the feature map $\mathbf{Z}$ followed by a CNN that increases the resolution to $128^2$ pixels. The CNN applies grouped convolutions with three groups to prevent undesired spatial correlation between the triplanes, see~\cite{wang2022rodin}. At resolution $128^2$pixels, we then add two further transformer blocks with efficient self-attention~\citep{XieNEURIPS2021} to facilitate learning the correct spatial correlations for the triplanes. The triplanes have a resolution of $128^2$pixels.
We use the superresolution modules and dual discriminator from EG3D~\citep{Chan2022CVPR}. The main task of the discriminator is to provide supervision for novel views, but it can also be used to improve the details in the reconstructed views~\citep{Esser2021CVPR}. We hence use $95\%$ novel views and $5\%$ input views when training the discriminator.
For the main models in the paper, the encoder has $\sim32M$ parameters. We use $8$ transformer blocks in the decoder accumulating to $\sim26M$ parameters for the full decoder. The discriminator has $\sim29M$ parameters. 
For computational efficiency, ablation studies (Table 3 of the main paper) are performed with downscaled models and at an image resolution of $128^2$ pixels instead of $256^2$pixels. The downscaled models have a reduced channel dimension; specifically, the triplane resolution is reduced to $64^2$ pixels and the decoder uses $4$ instead of $8$ transformer blocks. The resulting models count $1.6M$, $2.5M$, and $1.8M$ parameters for encoder, decoder and discriminator, respectively. 

The autoencoder uses Adam~\citep{Kingma2015ICLR} with a learning rate of $1.4\times10^{-4}$. However, for the superresolution modules and dual discriminator, we found it important to use the original learning rates from EG3D, which are $2\times10^{-3}$ and $1.9\times10^{-3}$, respectively. 
We train all autoencoders with a batch size of $32$ on $8$ NVIDIA A100-PCIE-40GB GPUs until the discriminator has seen around $5.5M$ training images. Training our autoencoder in this setting takes around 2.5 days.

\boldparagraph{Implementation and Training.}
\losses
Our autoencoder is trained with a reconstruction loss on the input view and an adversarial objective to supervise novel views~\citep{Mi2022ARXIV, Cai2022CVPR}. Similar to~\cite{rombach2021highresolution}, we add a small Kullback-Leibler (KL) divergence regularization term $\mathcal{L}_{KL}$ on the latent space $\mathbf{Z}$, as discussed above.
The reconstruction loss $\mathcal{L}_{rec}$ consists of a pixel-wise loss $\mathcal{L}_{px}= |\mathbf{\hat{I}}-\mathbf{I}|$, a perceptual loss $\mathcal{L}_{VGG}$~\citep{Zhang2018CVPRb}, and depth losses $\mathcal{L}_{depth}^{2D}$, $\mathcal{L}_{depth}^{3D}$. 
The full training objective is as follows
\begin{align}
    \mathcal{L}_{rec} &= \lambda_{px} \mathcal{L}_{px} + \lambda_{VGG} \mathcal{L}_{VGG} + \lambda_{depth}^{2D} \mathcal{L}_{depth}^{2D} + \lambda_{depth}^{3D} \mathcal{L}_{depth}^{3D} \\
    \begin{split}
    \min\limits_{\theta,\psi}\max\limits_{\phi} &\, [V(\bI, \bP_{nv}, \lambda; \theta, \psi, \phi) + V_{depth}(\bI, \bP_{nv}, \lambda_d; \theta, \psi, \chi) \\ 
    &\,+\, \cL_{rec}(\bI_P, \bP; \theta,\psi) + \lambda_{KL} \cL_{KL}(\bI_P; \theta)]
    \end{split}
\end{align}
where $\lambda_{\{\}}$ weigh the individual loss terms ($\lambda$ without subscript denotes the R1 regularization coefficient for the regular discriminator and $\lambda_d$ is the R1 regularization coefficient for the depth discriminator). The values for the weights are summarized in \tabref{tab:losses}. Note that $V$ and $V_{depth}$ denote the adversarial objectives of the regular and depth discriminator, respectively (see Eq. (5) in main paper).

Our code base builds on the official PyTorch implementation of StyleGAN~\citep{Karras2019CVPR} available at \url{https://github.com/NVlabs/stylegan3}, EG3D~\citep{Chan2022CVPR} available at \url{https://github.com/NVlabs/eg3d} and LDM~\citep{rombach2021highresolution} available at \url{https://github.com/CompVis/latent-diffusion}.
Similar to StyleGAN, we use a minibatch standard deviation layer at the end of the discriminator~\citep{Karras2018ICLR} and apply an exponential moving average of the autoencoder weights. 
Unlike ~\cite{Karras2019CVPR}, we do not train with path regularization or style-mixing.
To reduce computational cost and overall memory usage  R1-regularization~\citep{Mescheder2018ICML} is applied only once every 16 minibatches (also see R1-regularization coefficients $\lambda$ and $\lambda_d$ in~\tabref{tab:losses}). 

\boldparagraph{Second Stage: Latent Diffusion Model.}
\tblldmarchitecture
We provide detailed model and training hyperparameter choices in Table~\ref{tab:hyperparameters_ldm}. We follow the naming convention from LDM~\cite{rombach2021highresolution} and train the models for $\sim$200 epochs for each dataset. Our LDMs are trained on $4$ NVIDIA A100-PCIE-40GB GPUs for $8$ hours on SDIP elephant and for $1$ day on SDIP horse, dog. On ImageNet, we train a class-conditional model for $\sim$$5$ days, doubling the batch size from $128$ to $256$. Otherwise, we use the same hyperparameters and model size due to computational constraints. 

\boldparagraph{Guidance.} Class-conditioning is implemented through cross attention with learned embeddings, following ~\cite{rombach2021highresolution}. We also drop the class conditioning 10\% of the time to enable sampling with classifier-free guidance~\citep{ho2021classifierfree}. We use a guidance scale of $s=2$ in all our quantitative and qualitative results, unless indicated otherwise. The guidance scale $s$ is defined according to the equation
\begin{equation}
    \tilde{\boldsymbol{\epsilon}}^s_\omega (\mathbf{x}_\tau, \mathbf{c}) = \boldsymbol{\epsilon}_\omega (\mathbf{x}_\tau) + s \left(\boldsymbol{\epsilon}_\omega (\mathbf{x}_\tau, \mathbf{c})-\boldsymbol{\epsilon}_\omega (\mathbf{x}_\tau)\right),
\end{equation}
using the noise $\boldsymbol{\epsilon}$ prediction formulation (we can easily obtain the noise prediction $\boldsymbol{\epsilon}$ from the $\mathbf{v}$ prediction, which we use for training; see \cite{salimans2022progressive}). In the above equation, $\boldsymbol{\epsilon}_\omega (\mathbf{x}_\tau)$ denotes the unconditional score function, $\boldsymbol{\epsilon}_\omega (\mathbf{x}_\tau, \mathbf{c})$ the conditional score function when conditioning on class $\mathbf{c}$, and $\tilde{\boldsymbol{\epsilon}}^s_\omega (\mathbf{x}_\tau, \mathbf{c})$ is the resulting guided score function with guidance scale $s$. 

\boldparagraph{Compute Limitations and Further Scaling.} Note that due to their noisy training objective, diffusion models have been shown to scale well with more compute and larger batch sizes~\citep{rombach2021highresolution,karras2022edm,kingma2023understanding}. State-of-the-art models on regular, non-3D-aware image synthesis usually use significantly larger batch sizes ($>1000$) than we do. This suggests that our models in particular on the highly diverse ImageNet dataset could probably improved a lot with more computational resources.

\subsection{Monocular Depth}
\label{sec:monodepth}
In our pipeline, we leverage geometric cues from a pretrained monocular depth estimation network~\citep{Bhat2023ARXIV} to supervise the predicted depth $\hat{\bD}$ from the autoencoder. Note that the predicted depth is obtained using volume rendering, similarly to Eq.~(2) in the main paper
\begin{equation}
    \hat{\bD}(\br) = \frac{1}{\sum_{j=1}^N w_r^j} \sum_{i=1}^N w_r^i d_r^i
\end{equation}
where $d_r^i$ denotes the depth of sampling point $i$ along camera ray $r$ and $w_r^i$ is its corresponding rendering weight as defined in Eq.~(2) of the main paper.
The monocular depth used for supervision, however, is only defined up to scale. Let $\mathbf{D}$ denote the depth predicted by the monocular depth estimator. We first downsample it to match the resolution of the predicted, \ie, rendered depth, which refer to as $\mathbf{D}^{low}$.
Next, a scale $s$ and a shift $t$ are computed for each image by solving a least-squares criterion~\citep{Eigen2014NIPS,Ranftl2020PAMI}
\begin{equation}
    (s, t) = \arg\min_{s, t} \sum_{r\in\mathcal{R}} \left(s\hat{\bD}(\br) + t - \bD^{low}(\br)\right)\,.
\end{equation}

Defining $\bh=(s,t)^T$ and $\bd_r=(\hat{\bD}(\br), 1)^T$, the closed-form solution is given by
\begin{equation}
\bh = \left(\sum_r\bd_r\bd_r^T\right)^{-1}\left(\sum_r\bd_r\bD^{low}(\br)\right)\,.
\end{equation}

\section{Baselines}\label{sec:supp_baselines}
\boldparagraph{EG3D*.}
EG3D~\citep{Chan2022CVPR} relies on estimated camera poses which are not available for the datasets we consider in this work. Hence, we adapt it to work in view space and remove the need for camera poses. Both the generator and discriminator are originally conditioned on the camera poses. For our version, EG3D*, we remove the camera pose conditioning and model objects in view space by sampling novel views around the input view $\bP_0$ as described in ~\secref{sec:camsystem}. 
For fair comparison, we additionally train a variant of EG3D* that leverages monocular depth. Specifically, we equip EG3D* with the depth discriminator from our pipeline $D_\chi^{depth}$ which compares the rescaled rendered depth with the predictions from a pretrained monocular depth estimatior~\citep{Bhat2023ARXIV}, see~\secref{sec:monodepth} for more details on the rescaling of the depth.

We follow the training procedure from EG3D~\citep{Chan2022CVPR} and ensure that the models train stably by tuning the regularization strength $\lambda$ in the adversarial objective. We use $\lambda=1$ and $\lambda=2.5$ for both variants on SDIP Elephants and Horses, respectively. For SDIP Dogs, we found it best to use $\lambda=2.5$ for EG3D* and $\lambda=1.0$ EG3D*+ $D_{\text{depth}}$. For the depth discriminator, we set $\lambda_d=10\lambda$ for all experiments.
The models are trained until the discriminator has seen $10M$ images as we find that FID improvements are marginal afterwards. For evaluation, we select the best models in terms of FID.

Note that~\citep{skorokhodov20233dgp} includes a detailed study on training EG3D without camera poses. As 3DGP clearly outperforms EG3D in this setting, we did not train the original EG3D in canonical space but instead directly compare to 3DGP.

For inversion, we use code kindly provided by the authors of EG3D~\citep{Chan2022CVPR}. The inversion is performed using PTI~\citep{RoichTOG2023} and consists of two optimization stages. In the first stage, the style code $w$ is optimized to best match the input image. In the second stage, the network parameters are finetuned to further improve reconstruction quality. We observed that the inversion occasionally diverges. For the divergent cases, we reduce the learning rate in the optimization from $10^{-3}$ to $10^{-6}$ finding that this stabilizes the optimization.

\boldparagraph{POF3D, 3DGP.}
For POF3D~\citep{Shi2023CVPR}, we use the unpublished code that was kindly provided by the authors to train their model. We follow their training procedure and hyperparameter choices.
For 3DGP~\citep{skorokhodov20233dgp}, we trained the models using the publicly available code~\url{https://github.com/snap-research/3dgp}, which was released shortly before the submission deadline. 
We found that the training diverges on SDIP Elephants but, as suggested by the authors, were able to restart the model from the last stable checkpoint which then converged. For SDIP Horses training diverges after around $2.5M$ images, even when restarting from the last stable checkpoint, so we report results on the last stable checkpoint. 
Both POF3D and 3DGP are trained until the discriminator has seen $10M$ images as we observed no or only marginal improvements on FID with longer training.

\boldparagraph{StyleNeRF.}
We train StyleNeRF~\citep{Gu2021ARXIV} using the official implementation of the authors~\url{https://github.com/facebookresearch/StyleNeRF}. On SDIP datasets, we train until the disciminator has seen $20$M images, on Imagenet we stop training after $35$M images. In both cases, we only observed marginal changes in FID with longer training.

\boldparagraph{SceneScape*.}
As an additional baseline, we analyzed a combination of a 2D generative model and a 2D inpainting model. We base our implementation on the publicly available code of SceneScape~\citep{SceneScape} (\url{https://github.com/RafailFridman/SceneScape.git}).
Specifically, we generate images using the ImageNet checkpoint from LDM (\url{https://github.com/CompVis/latent-diffusion.git}). Next, we predict the corresponding depth using ZoeDepth, i.e. using the same pretrained depth estimator as in our approach, and warp the image to a novel view. Lastly, an inpainting model fills the wholes that result from warping. We use an inpainting variant of Stable Diffusion (\url{https://huggingface.co/docs/diffusers/using-diffusers/inpaint#stable-diffusion-inpainting}) and provide the warped image, its mask, and the text prompt “a photo of a <class name>” as input.

\section{Additional Results}\label{sec:supp_results}
\boldparagraph{Autoencoder for Compression and Novel-View Synthesis}
\aeresultssupp

\figref{fig:aeresultssupp} shows further examples of \ourmodel and baselines for novel view synthesis on images unseen during training (using GAN-inversion to embed the given images in latent space for EG3D* and EG3D* + $D^{depth}$). We see that our model generally correctly reconstructs the input object and is able to synthesize a high-quality novel view. Moreover, although there are small artifacts, \ourmodel also produces plausible geometry. In comparison, the baselines cannot accurately reconstruct the correct object and the geometry is often flat or incorrect.  %

\figref{fig:aeresultssupp_dog}, \figref{fig:aeresultssupp_ele}, \figref{fig:aeresultssupp_horse} and \figref{fig:aeresultssupp_imagenet} show further novel view synthesis results from \ourmodel's 3D-aware autoencoder. The results demonstrate our model's ability to correctly generate novel views, given the encoded input view. All viewpoint changes result in high-quality, realistic outputs.

\boldparagraph{3D-Aware Image Synthesis with Latent Diffusion Models}

\inpainting
We first compare \ourmodel against the additional baseline SceneScape*.
For quantitative analysis, we sample novel views similar to our evaluation by sampling yaw and pitch angles from Gaussian distributions with $\sigma = 0.3$ and $0.15$ radians, using the depth at the center of the image to define the rotation center. 
With this approach, we get an FID of $12.3$ on ImageNet, compared to $35.4$ for \ourmodel. However, as discussed in the main paper, FID only measures image quality and does not consider all important aspects of 3D-aware image synthesis, e.g. 3D consistency. In fact, we observe that the inpainting often adds texture-like artifacts or more instances of the given ImageNet class. We provide some examples in \figref{fig:inpainting}.

To enforce consistency between novel views, we run the full pipeline of SceneScape to fuse the mesh for multiple generated images. For this setting, we sample 9 camera poses by combining yaw angles of $[-0.3, 0., 0.3]$ and pitch angles of [-0.15, 0., 0.15] and iteratively update the mesh by finetuning the depth estimator. We show the final meshes in \figref{fig:inpainting} (bottom two rows). For all samples we evaluated, we observe severe intersections in the mesh and generally inferior geometry to our approach. We remark that SceneScape's test time optimization takes multiple minutes per sample and a large-scale quantitative evaluation was out of the scope of this work. 
Our rotating camera movements around a single object are much more challenging, e.g. due to larger occlusions,  than the backward motion shown in SceneScape. We hypothesize that this causes SceneScape to struggle more in our setting. 

\imagenetcompappendix
We also include more samples from \ourmodel and 3DGP on ImageNet in ~\figref{fig:diversity}. While samples from 3DGP look very similar within a class, \ourmodel generates diverse samples.
\figref{fig:ldmresultssupp_dog}, \figref{fig:ldmresultssupp_ele}, \figref{fig:ldmresultssupp_horse} and \figref{fig:ldmresultssupp_imagenet} show further \ourmodel results for 3D-aware image synthesis leveraging our latent diffusion model that synthesizes 3D-aware latent space encodings. We can see that all novel samples are high quality and the camera angle changes result in realistic view point changes of the scenes. Note that our model only trains on unposed, single view images and does not need to learn a complex pose distribution because it models objects in view space.

More generated results can be found in the supplementary video, including baseline comparisons and extracted geometries.

\boldparagraph{Ablation Studies.}

\ablation
~\figref{fig:ablation} visualizes the impact of different configurations on the geometry. For the base config, the model learns a planar geometry which reflects in a low NFS, \cf Tab. 3 in the main paper. Adding a discriminator and the ViT backbone improve geometry significantly but the geometry remains noisy. 
Incorporating geometry cues in the form of monocular depth and a depth discriminator $D_\chi^{depth}$ helps to remove artifacts from the geometry, resulting in a lower NFS. 
Modeling unbounded scenes with contracted coordinates does not significantly change geometry but improves nvFID, \cf Tab. 3 in the main paper. Further supervision on the rendered depth ($\mathcal{L}_{depth}^{2D}$) does not improve results but as it does not hurt performance either, we kept it in our pipeline. Lastly, both adding depth as an input to the encoder and directly supervising the rendering weights with $\mathcal{L}_{depth}^{3D}$
 result in slight improvements in NFS and qualitatively improve geometry. Note how the geometry is less planar when supervising the rendering weights with $\mathcal{L}_{depth}^{3D}$. We remark that the model used in this ablation is much smaller than our main models and has limited expressivity.
 
 \cfgmetrics
 We further ablate inference with different classifier-free guidance scales~\citep{ho2021classifierfree} in ~\tabref{tab:cfgmetrics}. For better compatibility with previous works, we also evaluated FID on 50K generated images for $s=3$, which drops from $28.5$ on 10K images to $25.3$ on 50K images. For computational efficiency, we report FID on 10K images on all other instances throughout the paper.
 
\boldparagraph{Limitations.}
Modeling instances in view space alleviates the need for posed images and learned camera distributions. However, it is a very challenging task. This can be seen, for instance, in ~\figref{fig:aeresultssupp}. It becomes difficult to produce sharp 3D geometry for both baseline models and \ourmodel, although \ourmodel produces 3D-aware novel-views with high quality and still the most realistic geometry.

Furthermore, as \ourmodel is trained with fixed azimuth and polar angle ranges, it is currently not possible to perform novel view synthesis across the full 360$^\circ$.
Increasing the ranges would be an interesting direction for future work. The challenge would lie in producing realistic content when largely unobserved regions become visible after large view point changes. Note, however, that to the best of our knowledge currently there exist no methods that can produce 360$^\circ$ views when trained on the kind of datasets we are training on, which only show a single view per instance, often from a similar front direction.

We observed that the synthesized samples occasionally exhibited plane-like geometry artifacts. 
Our adversarial loss in the autoencoder, in principle, should avoid this, as it enforces realistic renderings from different viewpoints.
We hypothesize that this is due to the autoencoder favoring in rare cases the simple solution of copying the image across triplanes to reduce the reconstruction loss.

Moreover, \ourmodel relies on fixed camera intrinsics (see \secref{sec:camsystem}), which we need to pick ourselves. However, we found that our choice worked well for all three datasets without further tuning. Hence, this is a minor limitation, in particular, compared to methods that work in canonical coordinate systems and need to estimate a pose for each instance (as discussed, this is not possible for many complex, non-aligned datasets). In future work, the camera intrinsics could potentially be learned.
\aeresultsoursdogs
\aeresultsoursele
\aeresultsourshorse
\aeresultsoursimagenet
\ldmresultsoursdog
\ldmresultsoursele
\ldmresultsourshorse
\ldmresultsoursimagenet

\end{document}